\documentclass[twoside,11pt]{article}

\usepackage[]{jmlr2e}

\usepackage{lastpage}
\jmlrheading{23}{2022}{1-\pageref{LastPage}}{8/21; Revised
3/22}{8/22}{21-0992}{Matthias Feurer and Katharina Eggensperger and Stefan Falkner and Marius Lindauer and Frank Hutter}

\ShortHeadings{Auto-sklearn 2.0}{Feurer, Eggensperger, Falkner, Lindauer and Hutter}
\firstpageno{1}

\usepackage{amsmath}

\usepackage{booktabs}
\usepackage[utf8]{inputenc}
\usepackage{amssymb}
\usepackage{multirow}
\usepackage[dvinames,dvipsnames]{xcolor}
\usepackage{wrapfig}
\usepackage[ruled]{algorithm2e}
\usepackage{algorithmic}
\usepackage{hyperref}

\usepackage{scalerel,stackengine}
\stackMath
\newcommand\reallywidehat[1]{%
\savestack{\tmpbox}{\stretchto{%
  \scaleto{%
    \scalerel*[\widthof{\ensuremath{#1}}]{\kern-.6pt\bigwedge\kern-.6pt}%
    {\rule[-\textheight/2]{1ex}{\textheight}}
  }{\textheight}%
}{0.5ex}}%
\stackon[1pt]{#1}{\tmpbox}%
}
\usepackage{bm}
\usepackage{dsfont}

\DeclareMathOperator*{\argmin}{argmin}

\newcommand{\automl}[0]{AutoML}
\newcommand{\autosklearn}[0]{\textit{Auto-sklearn}}
\newcommand{\asklone}[0]{\textit{Auto-sklearn 1.0}}
\newcommand{\askltwo}[0]{\textit{Auto-sklearn 2.0}}
\newcommand{\poshautosklearn}[0]{\textit{PoSH \autosklearn{}}}

\newcommand{\gerror}[0]{GE}
\newcommand{\nautodata}[0]{39}
\newcommand{\ndatasets}[0]{208}
\newcommand{\horizon}[1]{optimization budget#1}

\newcommand{\autoweka}[0]{\textit{Auto-WEKA}}
\newcommand{\hto}[0]{\textit{H2O}}
\newcommand{\tpot}[0]{\textit{TPOT}}
\newcommand{\configspace}[0]{configuration space}

\newcommand{\nfolds}[0]{K}
\newcommand{\folditer}[0]{k}
\newcommand{\dataiter}[0]{d}

\newcommand{\Lossfunction}[0]{\mathcal{L}}
\newcommand{\Dataset}[0]{\mathcal{D}}
\newcommand{\TrainingSet}[0]{\mathcal{D}_{\text{train}}}
\newcommand{\MetaSet}[0]{\mathbf{D}_{\text{meta}}}
\newcommand{\AutoMLSet}[0]{\mathbf{D}_{\text{test}}}
\newcommand{\TestSet}[0]{\mathcal{D}_{\text{test}}}

\newcommand{\vlambda}[0]{{\bm{\lambda}}}
\newcommand{\conf}[0]{\vlambda}
\newcommand{\confs}[0]{{\bm{\Lambda}}}

\newcommand{\Portfolio}[0]{\mathcal{P}}
\newcommand{\MSS}[0]{S}
\newcommand{\Model}[0]{\mathcal{M}}

\newcommand{\Candidates}[0]{\mathcal{C}}
\DeclareMathOperator*{\penaltyfunction}{penalty}
\newcommand{\policy}{\pi}
\newcommand{\policies}{\Pi}
\newcommand{\selector}{policy selector}
\newcommand{\modelselector}{model-based \selector{}}

\newcommand{\note}[1]{
	\noindent~\\
	\vspace{0.25cm}
	\fcolorbox{orange}{orange}{\parbox{0.95\columnwidth}{#1}}
	\vspace{0.25cm}
}
\renewcommand{\note}[1]{}
\newcommand{\CRCnote}[1]{}

\newcommand{\updates}[1]{{\textbf{\color{blue}{#1}}}}
\renewcommand{\updates}[1]{#1}

\begin{document}
\title{Auto-Sklearn 2.0: Hands-free AutoML via Meta-Learning}
\author{\name Matthias Feurer$^1$ \email{feurerm@cs.uni-freiburg.de} \\
        \name Katharina Eggensperger$^1$ \email{eggenspk@cs.uni-freiburg.de} \\
        \name Stefan Falkner$^2$ \email{Stefan.Falkner@de.bosch.com} \\
        \name Marius Lindauer$^3$ \email{lindauer@tnt.uni-hannover.de} \\
        \name Frank Hutter$^{1,2}$ \email{fh@cs.uni-freiburg.de} \\
        \addr $^1$Department of Computer Science,  Albert-Ludwigs-Universität Freiburg\\
        \addr $^2$Bosch Center for Artificial Intelligence, Renningen, Germany\\
        \addr $^3$Institute of Information Processing, Leibniz University Hannover 
}

\editor{Marc Schoenauer}

\maketitle

\begin{abstract}%
Automated Machine Learning (\automl{}) supports practitioners and researchers with the tedious task of designing machine learning pipelines and has recently achieved substantial success.
In this paper, we introduce new \automl{} approaches motivated by our winning submission to the second ChaLearn \automl{} challenge. We develop \poshautosklearn{}, which enables \automl{} systems to work well on large datasets under rigid time limits by using a new, simple and meta-feature-free meta-learning technique and by employing a successful bandit strategy for budget allocation.
However, \poshautosklearn{} introduces even more ways of running \automl{} and might make it harder for users to set it up correctly.
Therefore, we also go one step further and study the design space of \automl{} itself, proposing a solution towards truly hands-free \automl{}.
Together, these changes give rise to the next generation of our \automl{} system, \askltwo{}.
We verify the improvements by these additions in an extensive experimental study on $\nautodata$ \automl{} benchmark datasets. We conclude the paper by comparing to other popular \automl{} frameworks and \asklone{}, reducing the relative error by up to a factor of $4.5$, and yielding a performance in 10 minutes that is substantially better than what \asklone{} achieves within an hour.
\end{abstract}


\section{Introduction}\label{sec:introduction}

The recent substantial progress in machine learning (ML) has led to a growing demand for hands-free ML systems that can support developers and ML novices in efficiently creating new ML applications. Since different datasets require different ML pipelines, this demand has given rise to the area of automated machine learning (\automl{}; \citealp{hutter-book19a}). Popular \automl{} systems, such as \autoweka{}~\citep{thornton-kdd13a}, \emph{hyperopt-sklearn}~\citep{komer-automl14a}, \autosklearn~\citep{feurer-nips15a}, \tpot{}~\citep{olson-gecco16a} and \emph{Auto-Keras}~\citep{jin-sigkdd19a} perform a combined optimization across different preprocessors, classifiers or regressors and their hyperparameter settings, thereby reducing the effort for users substantially.

To assess the current state of AutoML and, more importantly, to foster progress in \automl{}, ChaLearn conducted a series of \automl{} challenges~\citep{guyon-automl19a}, which evaluated \automl{} systems in a systematic way under rigid time and memory constraints. 
Concretely, in these challenges, the \automl{} systems were required to deliver predictions in less than $20$ minutes. 
On the one hand, this would allow to efficiently integrate AutoML into the rapid prototype-driven workflow of many data scientists and, on the other hand, help to democratize ML by requiring less compute resources.

We won both the first and second AutoML challenge with modified versions of \autosklearn{}.
In this work, we describe in detail how we improved \autosklearn{} from the first version \citep{feurer-nips15a} to construct \poshautosklearn{}, which won the second competition and then describe how we improved \poshautosklearn{} further to yield our current approach for \askltwo{}. 

Particularly, while 
\automl{} \updates{relieves} the user from making low-level design decisions (e.g. which model to use), \automl{} itself opens a myriad of high-level design decisions, e.g. which model selection strategy to use~\citep{guyon-jmlr10a,guyon-ijcnn15a,raschka-arxiv2018a} or how to allocate the given time budget~\citep{jamieson-aistats16a}. Whereas our submissions to the \automl{} challenges were mostly hand-designed, in this work, we go one step further by automating \automl{} itself to fully unfold the potential of AutoML in practice.\footnote{The work presented in this paper is in part based on two earlier workshop papers introducing some of the presented ideas in preliminary form~\citep{feurer-automl18a,feurer-automl18b}.}

\begin{figure*}[t]
    \centering
    \includegraphics[width=0.99\textwidth]{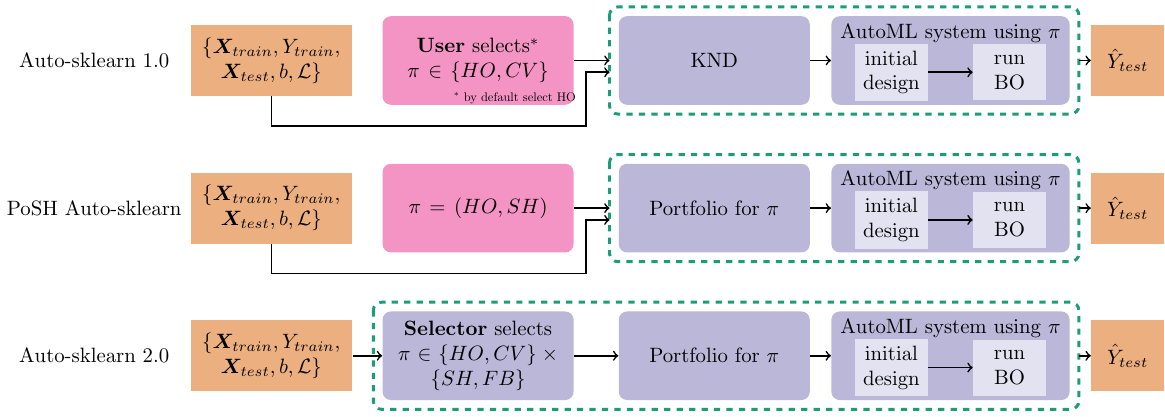}
\caption{Schematic overview of \asklone{}, \poshautosklearn{}, and \askltwo{}. Orange rectangular boxes refer to input and output data, while rounded purple boxes denote parts of the \automl{} system (surrounded by a green dashed line). The pink, rounded box refers to a human in the loop required for manual design decisions. The newer \automl{} system\updates{s} simplify the usage of \autosklearn{} and reduce the required user input.
\updates{We describe \poshautosklearn{} in Section~\ref{sec:partI} and give a schematic overview in Figure~\ref{fig:schema_posh}. Similarly, we describe \askltwo{} in Section~\ref{sec:partII} and provide a schematic overview in Figure~\ref{fig:schema_main}.
}}
    \label{fig:schema_askl_versions}
\end{figure*}

After detailing the \automl{} problem  we consider in Section~\ref{sec:background}, we present two main parts making the following contributions:
\begin{description}
    \item[Part I: Portfolio Successive Halving in \poshautosklearn{}.] In this part (see Section~\ref{sec:partI}), we introduce budget allocation strategies as a complementary design choice to model selection strategies (holdout (HO) and cross-validation (CV)) for \automl{} systems. \updates{We suggest using the budget allocation strategy successive halving (SH) as an alternative to always using the full budget (FB) to evaluate a configuration} to allocate more resources to promising ML pipelines. Furthermore, we introduce both the practical approach as well as the theory behind building better portfolios for the meta-learning component of \autosklearn{}. We show that this combination substantially improves performance, yielding stronger results in 10 minutes than \asklone{} achieved in 60 minutes.
    \item[Part II: Automating \automl{} in \askltwo{}.] In this part (see Section~\ref{sec:partII}), we propose a meta-learning technique based on algorithm selection \updates{to automatically select the best setting of the \automl{} system itself for a given dataset}. We dub the resulting system \askltwo{} and depict the evolution from \asklone{} via \poshautosklearn{} to \askltwo{} in Figure~\ref{fig:schema_askl_versions}. 
\end{description}
In Section~\ref{sec:partIII}, we additionally use the \automl{} benchmark~\citep{gijsbers-automl19a} to evaluate \askltwo{} against other popular \automl{} systems and show improved performance under rigid time constraints.
Section~\ref{sec:related} then puts our work into the context of related works, and Section~\ref{sec:discussion}
concludes the paper with open questions, limitations and future work.

\section{Problem Statement}\label{sec:background}

\automl{} is a widely used term, so, here we first define the problem we consider in this work. 
Let $P(\Dataset)$ be a distribution of datasets from which we can sample an individual dataset's distribution $P_d = P_d(\mathbf{x},y)$.
The \automl{} problem we consider is to generate a trained pipeline $\Model_{\conf} : \mathbf{x} \mapsto y$, hyperparameterized by $\conf \in \confs$ that automatically produces predictions for samples from the distribution $P_d$ minimizing the expected generalization error:\footnote{Our notation follows \cite{vapnik-neurips1991a}.}
\begin{equation}
    \gerror(\Model_\conf) = \mathds{E}_{(\mathbf{x}, y) \sim P_d} \left[\Lossfunction(\Model_\conf( \mathbf{x}), y)\right].
\end{equation}
Since a dataset can only be observed through a set of $n$ independent observations $\Dataset_d =  \{(\mathbf{x}_1,y_1),\dots,(\mathbf{x}_{n},y_{n})\} \sim P_d$, we can only empirically approximate the generalization error on sample data:
\begin{equation}
    \reallywidehat{\gerror}(\Model_\conf, \Dataset_d) = \frac{1}{n}\sum_{i = 1}^{n}
    \Lossfunction(\Model_\conf(\mathbf{x}_i),y_i). \label{eq:automl}
\end{equation}

\noindent
In practice we have access to two disjoint, finite samples which we from now on denote $\TrainingSet$ and $\TestSet$ ($\mathcal{D}_{\dataiter,\text{train}}$ and $\mathcal{D}_{\dataiter,\text{test}}$ in case we reference a specific dataset $P_\dataiter$). For searching the best ML pipeline, we only have access to $\TrainingSet$, however, in the end performance is estimated once on $\TestSet$. 
\automl{} systems use this to automatically search for the best $\Model_{\conf^*}$:
\begin{equation}
    \Model_{\conf^*} \in \argmin_{\conf \in \confs} \reallywidehat{\gerror}(\Model_\conf, \TrainingSet),
\end{equation}

\noindent
and estimate GE, e.g., by a $\nfolds$-fold cross-validation:
\begin{equation}
    \reallywidehat{\gerror}_{\text{CV}}(\Model_\conf,\TrainingSet) = \frac{1}{\nfolds} \sum_{\folditer=1}^\nfolds \reallywidehat{\gerror}(\Model_\conf^{\TrainingSet^{(\text{train},\folditer)}},{\TrainingSet^{(\text{val},\folditer)}}),
    \label{eq:ge_cv}
\end{equation}

\noindent
where $\Model_\conf^{\TrainingSet^{(\text{train},\folditer)}}$ denotes that $\Model_\conf$ was trained on the training split of $\folditer$-th fold $\TrainingSet^{(\text{train},\folditer)} \subset \TrainingSet$, and it is then evaluated on the validation split of the $\folditer$-th fold $\TrainingSet^{(\text{val},\folditer)} = \TrainingSet \setminus \TrainingSet^{(\text{train},\folditer)}$.\footnote{Alternatively, one could use holdout to estimate $\gerror$ with  $\reallywidehat{\gerror}_{\text{HO}}(\Model_\conf,\TrainingSet) = \reallywidehat{\gerror}(\Model_\conf^{\TrainingSet^{\text{train}}},\TrainingSet^{\text{val}})$.
}
Assuming that, via $\conf$, an \automl{} system can select both, an algorithm and its hyperparameter settings, this definition using $\reallywidehat{\gerror}_{\text{CV}}$ is equivalent to the definition of the CASH (\emph{C}ombined \emph{A}lgorithm \emph{S}election and \emph{H}yperparameter optimization) problem~\citep{thornton-kdd13a,feurer-nips15a}.
However, it is unlikely that, whatever optimization algorithm we use, the \automl{} system finds the exact optimum location $\conf^*$. Instead, the \automl{} system will return the best ML pipeline it has trained during the search process, which we denote by $\Model_{\hat{\conf}^{*}}$, and the hyperparameter settings it was trained with by $\hat{\conf}^{*}$.

\subsection{Time-bounded \automl{}}

In practice, users are not only interested in obtaining an optimal pipeline $\Model_{\conf^*}$ eventually, but have constraints on how much time and compute resources they are willing to invest. We denote the time it takes to evaluate $\reallywidehat{\gerror}(\Model_{\conf}, \TrainingSet)$ as $t_{\conf}$ and the overall \horizon{} by $T$. 
Our goal is to find
\begin{equation}
    \Model_{\conf^*} \in \argmin_{\conf \in \confs} \reallywidehat{\gerror}(\Model_\conf, \TrainingSet) \text{ s.t.} \left( \sum t_{\conf_i} \right) < T
\label{eq:minproblem}
\end{equation}
where the sum is over all evaluated pipelines $\conf_i$, explicitly honouring the \horizon~$T$. As before, the \automl{} system will return the best model it has found within the optimization budget, $\Model_{\hat{\conf}^{*}}$.

\subsection{Generalization of \automl{}}

Ultimately, an \automl{} system $\mathcal{A}: \Dataset \mapsto \Model_{\hat{\conf}^{*}}^\Dataset$ should not only perform well on a single dataset but on the entire distribution over datasets $P(\Dataset)$. Therefore, the meta-problem of \automl{} can be formalized as minimizing the generalization error over this distribution of datasets:
\begin{equation}
    \gerror(\mathcal{A}) = \mathds{E}_{\Dataset_d \sim P(\Dataset)} \left[ \reallywidehat{\gerror}(\mathcal{A}(\Dataset_d),\Dataset_d) \right],
\end{equation}
which in turn can again only be approximated by a finite set of meta-train datasets $\MetaSet$ (each with a finite set of observations):
\begin{equation}
\label{eq:general_automl_empirical_ge}
    \reallywidehat{\gerror}(\mathcal{A},\MetaSet) = \frac{1}{\mid\MetaSet\mid} \sum_{\dataiter=1}^{|\MetaSet|} \reallywidehat{\gerror}(\mathcal{A}(\Dataset_\dataiter),\Dataset_\dataiter).
\end{equation}

Having set up the problem statement, we can use this to further formalize our goals. Instead of using a single, fixed \automl{} system $\mathcal{A}$, we will introduce optimization policies~$\pi$, a combination of hyperparameters of the \automl{} system and specific components to be used in a run, which can be used to configure an \automl{} system for specific use cases. We then denote such a configured \automl{} system as $\mathcal{A}_\pi$. 

We will first construct $\pi$ manually in Section~\ref{sec:partI}, introduce a novel system for designing $\pi$ from data in Section~\ref{sec:partII} and then extend this to a (learned) mapping $\Xi : \Dataset \rightarrow \pi$ which automatically suggests an optimization policy for a new dataset using algorithm selection. 
This problem setup can also be used to introduce generalizations of the algorithm selection problem such as algorithm configuration~\citep{birattari-gecco02a,hutter-jair09a,kleinberg-ijcai17a}, per-instance algorithm configuration~\citep{xu-aaai10a,malitsky-cp12a} and dynamic algorithm configuration~\citep{biedenkapp-ecai20a} on a meta-level; but we leave these for future work. In addition, instead of selecting between multiple policies of a single \automl{} system, the presented method can be applied to choose between different \automl{} systems without adjustments. However, instead of maximizing performance by invoking many \automl{} systems, thereby increasing the complexity, our goal is to improve single \automl{} systems to make them easier to use by decreasing complexity for the user.

\section{Part I: Portfolio Successive Halving in \poshautosklearn{}}
\label{sec:partI}

In this section we introduce our winning solution for the second AutoML competition~\citep{guyon-automl19a}, \poshautosklearn{}, short for \textbf{Po}rtfolio \textbf{S}uccessive \textbf{H}alving. We first describe our use of portfolios to warmstart an AutoML system and then motivate using the successive halving bandit strategy. Next, we describe practical considerations for building \poshautosklearn{}\updates{, give a schematic overview and recap additional handcrafted techniques we used in the competition}. We end this first part of our main contributions  with an experimental evaluation demonstrating the performance of \poshautosklearn{}.

\subsection{Portfolio Building}
\label{sec:portfolio} 
Finding the optimal solution to the time-bounded optimization problem from Equation~\eqref{eq:minproblem} requires searching a large space of possible ML pipelines as efficiently as possible. BO is a strong approach for this, but its vanilla version starts from scratch for every new problem. A better solution is to warmstart BO with ML pipelines that are expected to work well, as done in the k-nearest dataset (KND) approach of \asklone{}~(\citealp{reif-ml12a,feurer-aaai15a,feurer-nips15a}; see also the related work in Section~\ref{ssec:components}).
However, we found this solution to introduce new problems: 
\begin{enumerate}
\setlength\itemsep{0.1em}
    \item It is time-consuming since it requires to compute meta-features describing the characteristics of datasets.
    \item It adds complexity to the system as the computation of the meta-features must also be done with a time and memory limit. 
    \item Many meta-features are not defined with respect to categorical features and missing values, making them hard to apply for most datasets. 
    \item It is not immediately clear which meta-features work best for which problem. 
    \item In the KND approach,
    there is no mechanism to guarantee that we do not execute redundant ML pipelines. 
\end{enumerate}
We indeed suffered from these issues in the first \automl{} challenge, failing on one track due to running over time for the meta-feature generation, although we had already removed landmarking meta-features due to their potentially high runtime.
Therefore, here we propose a meta-feature-free approach that does not warmstart with a set of configurations specific to a new dataset, but which uses a static \emph{portfolio} -- a set of complementary configurations that covers as many diverse datasets as possible and minimizes the risk of failure when facing a new task.

So, instead of evaluating configurations chosen \emph{online} by the KND method, we construct a portfolio, consisting of high-performing and complementary ML pipelines to perform well on as many datasets as possible, \emph{offline}. Then, for a dataset at hand, all pipelines in this portfolio are simply evaluated one after the other. If time is left afterwards, we continue with pipelines suggested by BO warmstarted with the evaluated portfolio pipelines. 
We introduce portfolio-based warmstarting to avoid computing meta-features for a new dataset. However, the portfolios also work inherently differently. 
While the KND method is aimed at using only well-performing configurations, a portfolio is built such that there is a diverse set of configurations, starting with ones that perform well on average and then moving to more specialized ones.
Thus, it can be seen as an optimized initial design for the BO method.

In the following, we describe our offline procedure for constructing such a portfolio \updates{and give theoretical underpinning by a performance bound}.

\subsubsection{Approach} \label{ssec:portfolio:approach}
We first describe how we construct a portfolio given a finite set of candidate pipelines $\Candidates = \{\conf_1, ..., \conf_l\}$. Additionally, we assume that there exists a set of datasets $\MetaSet = \{\Dataset_1, \dots, \Dataset_{\vert\MetaSet\vert}$\} and we wish to build a portfolio $\Portfolio$ consisting of a subset of the pipelines in $\Candidates$ that performs well on $\MetaSet$. 

We outline the process to build such a portfolio in Algorithm~\ref{alg:build_greedy_portfolio}. 
First, we initialize our portfolio $\Portfolio$ to the empty set (Line 2). Then, we repeat the following procedure until $|\Portfolio|$ reaches a pre-defined limit: From a set of candidate ML pipelines $\Candidates$, we greedily add a candidate $\conf^+ \in \Candidates$ to $\Portfolio$ that reduces the estimated generalization error over all meta-datasets most (Line 4), and then remove $\conf^+$ from $\Candidates$ (Line 5). 

The estimated generalization error of a portfolio $\Portfolio$ on a single dataset $\Dataset$ is the performance of the best pipeline $\conf \in \Portfolio$ on $\Dataset$ according to the model selection and budget allocation strategy. This can be described via a function $\MSS(\cdot,\cdot,\cdot)$, which takes as input a function to compute the estimated generalization error (e.g., as defined in Equation~\ref{eq:ge_cv}), a set of machine learning pipelines to train, and a dataset. It then returns the pipeline with the lowest estimated generalization error as
\begin{equation}
\label{eq:S}
    \Model_{\conf^*}^{\Dataset} = S(\reallywidehat{GE},\Portfolio, \Dataset) \in \argmin_{\Model_{\conf}^{\Dataset} \in \Portfolio} \reallywidehat{GE}(\Model_{\conf}^{\Dataset}, \Dataset).
\end{equation}
In case the result of $\argmin$ is not unique, we return the model that was evaluated first. The estimated generalization error of $\Portfolio$ across all meta-datasets $\MetaSet = \{\Dataset_1, \dots, \Dataset_{\vert\MetaSet\vert}\}$ is then 

\begin{equation}
 \label{eq:portfolio_set_function}
 \reallywidehat{\gerror}_\MSS(\Portfolio,\MetaSet) = \sum_{\dataiter = 1}^{\vert\MetaSet\vert} \reallywidehat{\gerror}\left(\MSS\left(\reallywidehat{GE},\Portfolio, \Dataset_\dataiter\right), 
 \Dataset_\dataiter^{\text{val}}\right),
\end{equation}

Here, we give the equation for using holdout, and in Appendix~\ref{app:pseudo-code} we provide the exact notation for cross-validation and successive halving.

\begin{algorithm}[tbp]
\begin{small}
\caption{Greedy Portfolio Building}
  \label{alg:build_greedy_portfolio}
\begin{algorithmic}[1]
    \STATE {\bfseries Input:} Set of candidate ML pipelines $\Candidates$,  $\MetaSet = \{\Dataset_1, \dots, \Dataset_{\vert\MetaSet\vert}\}$, maximal portfolio \\size $p$, model selection strategy $\MSS$
    \STATE $\Portfolio = \emptyset$
    \WHILE{$|\Portfolio| < p$}
    \STATE $\conf^+ = \argmin_{\conf \in \Candidates} \reallywidehat{GE}_\MSS(\Portfolio \cup \{\conf\}, \MetaSet)$ \\\tcp*[h]{Ties are broken favoring the model trained first.}
    \STATE $\Portfolio = \Portfolio \cup \conf^+,\text{ } \Candidates = \Candidates \setminus \{\conf^+\}$
    \ENDWHILE
    \RETURN Portfolio $\Portfolio$
\end{algorithmic}
\end{small}
\end{algorithm}

\begin{figure}
    \centering
    \includegraphics[width=0.7\textwidth]{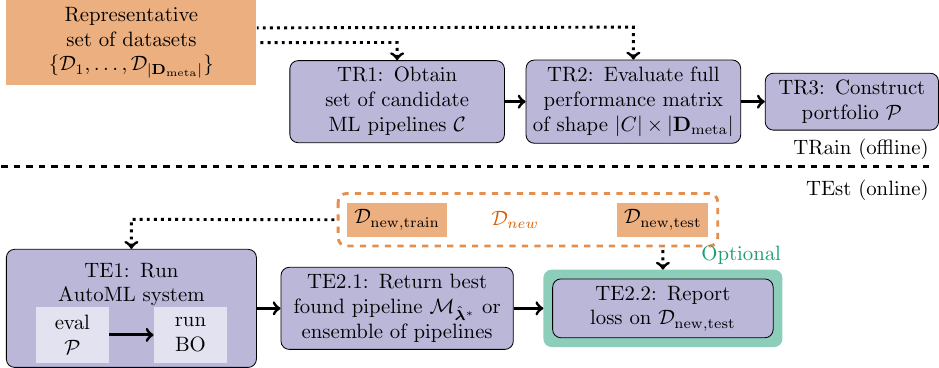}
    \caption{\updates{Schematic Overview of \poshautosklearn{} with the offline portfolio building phase (TR1-TR3) above and the test phase (TE1-TE2) below the dashed line. Rounded, purple boxes refer to computational steps while rectangular, orange boxes depict the input data to the \automl{} system.}}
    \label{fig:schema_posh}
\end{figure}

We now detail how to construct the set of candidate pipelines $\Candidates$ and describe how finding the candidate pipelines and constructing the portfolio fit in the larger picture.
We give a schematic overview of this process in Figure~\ref{fig:schema_posh}.
It consists of a training (TR1--TR3) and a testing stage (TE1--TE2).

Having collected datasets $\MetaSet$ (we describe in Section~\ref{sssec:datasets} how we did this for our experiments), we obtain the candidate ML pipelines (TR1) by running \autosklearn{} without meta-learning and without ensembling on each dataset. We limit ourselves to a finite set of portfolio candidates $\Candidates$, and pick one candidate per dataset. Then, we build a performance matrix of size $|\Candidates| \times \vert\MetaSet\vert$ by evaluating each of these candidate pipelines on each dataset (TR2, we refer to Section~\ref{sssec:meta-data-generation} for a detailed description of the meta-data generation). 
Finally, we then use this matrix to build a portfolio using Algorithm~\ref{alg:build_greedy_portfolio} for the combination of model selection strategy holdout and budget allocation strategy SH in training step TR3.

For a new dataset $\Dataset_{\text{new}} \in \AutoMLSet{}$, we apply the \automl{} system using SH, holdout and the portfolio to $\Dataset_{\text{new}}$ (TE1). Finally, we return the best found pipeline $\Model_{\hat{\conf}^{*}}$, or an ensemble of the evaluated pipelines, based on the training set of $\Dataset_{\text{new}}$ (TE2.1). Optionally, we can then compute the loss of $\Model_{\hat{\conf}^{*}}$ on the test set of $\Dataset_{\text{new}}$ (TE2.2); we emphasize that this would be the only time we ever access the test set of $\Dataset_{\text{new}}$.

To build a portfolio across datasets, we need to take into account that the generalization errors for different datasets live on different scales~\citep{bardenet-icml13a}. Thus, before taking averages, for each dataset, we transform the generalization errors to the distance to the best observed performance scaled between zero and one, a metric named \emph{distance to minimum}; which when averaged across all datasets is known as \emph{average distance to the minimum}~(ADTM)~\citep{wistuba-dsaa15a,wistuba-ml18a}. We compute the statistics for zero-one scaling individually for each combination of model selection and budget allocation (i.e., we use the lowest observed test loss and the largest observed test loss for each meta-dataset).

For each meta-dataset $\Dataset_\dataiter \in \MetaSet$ we have access to both $\Dataset_{d,\text{train}}$ and $\Dataset_{d,\text{test}}$. In the case of holdout, we split the training set $\Dataset_{\dataiter,\text{train}}$ into two smaller disjoint sets $\Dataset^{\text{train}}_{\dataiter,\text{train}}$ and $\Dataset^{\text{val}}_{\dataiter,\text{train}}$. We usually train models using $\Dataset^{\text{train}}_{\dataiter,\text{train}}$ and use $\Dataset^{\text{val}}_{\dataiter,\text{train}}$ to choose a ML pipeline $\Model_{\conf}$ from the portfolio by means of the model selection strategy $S$ (instead of holdout we can of course also use cross-validation to compute the validation loss). 
However, if we instead choose the ML pipeline on the test set $\Dataset_{\dataiter,\text{test}}$, Equation~\ref{eq:portfolio_set_function} becomes a monotone and submodular set function, which results in favorable guarantees for the greedy algorithm that we detail in Section \ref{subsec:submodularity}.
We follow this approach for the portfolio construction in the offline phase; we emphasize that for a new dataset $\Dataset_{\text{new}}$, we of course do not require access to the test set $\Dataset_{\text{new},\text{test}}$.

\subsubsection{Theoretical Properties of the Greedy Algorithm}\label{subsec:submodularity}

Besides the already mentioned practical advantages of the proposed greedy algorithm,
this algorithm also enjoys a bounded worst-case error.

\newtheorem{mydef}{Definition}
\newtheorem{prop}{Proposition}

\begin{prop}
Minimizing the test loss of a portfolio $\Portfolio$ on a set of datasets $\Dataset_1, \dots, \Dataset_{\vert\MetaSet\vert}$, when choosing an ML pipeline from $\Portfolio$ for $\Dataset_\dataiter$ using holdout or cross-validation based on its performance on $\Dataset_{\dataiter,\text{test}}$, is equivalent to the \emph{sensor placement problem} for minimizing detection time~\citep{krause_jwrpm2008a}.
\end{prop}
We detail this equivalence in Appendix~\ref{app:proof}. Thereby, we can apply existing results for the sensor placement problem to our problem. Using the test set of the meta-datasets $\MetaSet$ to construct a portfolio is perfectly fine as long as we do not use new datasets $\Dataset_{\text{new}} \in \AutoMLSet$ which we use for testing the approach.
\begin{corollary}
\label{cor1}
The penalty function for all meta-datasets is submodular.
\end{corollary}
We can directly apply the proof from \cite{krause_jwrpm2008a} that the so-called penalty function (i.e., maximum estimated generalization error minus the observed estimated generalization error) is submodular and monotone to our problem setup. Since linear combinations of submodular functions are also submodular~\citep{krause-trac14a}, the penalty function is also submodular.
\begin{corollary}
The problem of finding an optimal portfolio $\Portfolio^*$ is NP-hard~\citep{nemhauser-1978,krause_jwrpm2008a}.
\end{corollary}
\begin{corollary}
Let $R$ denote the expected penalty reduction of a portfolio across all datasets, compared to the empty portfolio (which yields the worst possible score for each dataset).
The greedy algorithm returns a portfolio $\Portfolio$ such that $R(\Portfolio^*) \ge R(\Portfolio) \ge (1 - \frac{1}{e})R(\Portfolio^*)$. 
\end{corollary}
This means that the greedy algorithm closes at least 63\% of the gap between the worst ADTM score (1.0) and the score the best possible portfolio $\Portfolio^*$ of size $|\Portfolio|$ would achieve \citep{nemhauser-1978,krause-trac14a}. A generalization of this result given by \updates{\citet[Theorem 1.5]{krause-trac14a}} also tightens this bound to close 99\% of the gap between the worst ADTM score and the score the optimal portfolio $\Portfolio^*$ of size $|\Portfolio^*|$ would achieve, by extending the portfolio constructed by the greedy algorithm to size $5\cdot|\Portfolio|$. Please note that a portfolio of size $5 \cdot |\Portfolio|$ could be better than the optimal portfolio of size $|\Portfolio|$.
This means that we can find a close-to-optimal portfolio on the meta-train datasets $\MetaSet$ at the very least. Under the assumption that we apply the portfolio to datasets from the same distribution of datasets, we have a strong set of default ML pipelines.

We could also apply other strategies for the sensor set placement in our setting, such as mixed integer programming strategies, which can solve it optimally; however, these do not scale to portfolio sizes of a dozen ML pipelines~\citep{krause_jwrpm2008a,pfisterer-arxiv2018a}.

The same proposition (with the same proof) and corollaries apply if we select an ML pipeline based on an intermediate step in a learning curve or use cross-validation instead of holdout.
We discuss using the validation set and other model selection and budget allocation strategies in Appendix~\ref{sec:choose_on_valid} and Appendix~\ref{app:choosing-for-sh}.

\subsection{Budget Allocation using Successive Halving}
\label{ssec:resource_allocation} 

\updates{A key issue we identified during the last \automl{} challenge was that training expensive configurations on the complete training set, combined with a low time budget, does not scale well to large datasets.
At the same time, we noticed that our (then manual) strategy to run predefined pipelines on subsets of the data already yielded predictions good enough for ensemble building. 
This questions the common choice of assigning the same amount of resources to all pipeline evaluations, i.e. time, compute and data.} 

\updates{For this reason we introduce the principle of \emph{budget allocation strategies} to \automl{}, that describe how the resources are allocated to the pipeline evaluations.
This is an orthogonal design decision to the \emph{model selection strategy}, which approximates the generalization error of a single ML pipeline, and which is typically tackled by holdout or K-fold cross-validation (see Section~\ref{ssec:components}).}

\updates{
As a principled alternative to always using the full budget, we used the successive halving bandit strategy~(SH;~\citealp{karnin-icml13a,jamieson-aistats16a}), which assigns more budget to promising machine learning pipelines and can easily be combined with iterative algorithms.
}

\subsubsection{Approach}\label{ssec:shapproach}
\automl{} systems evaluate each pipeline under the same resource limitations and on the same budget (e.g., number of iterations using iterative algorithms). To increase efficiency for cases with tight resource limitations, we suggest allocating more resources to promising pipelines by using SH~\citep{karnin-icml13a,jamieson-aistats16a} to prune poor-performing pipelines aggressively. 

Given a minimal and maximal budget per ML pipeline, SH starts by training a fixed number of ML pipelines for the smallest budget. Then, it iteratively selects $\frac{1}{\eta}$ of the pipelines with the lowest generalization error, multiplies their budget by $\eta$, and re-evaluates. This process is continued until only a single ML pipeline is left or the maximal budget is spent\updates{, and replaces the standard holdout procedure in which every ML pipeline is trained for the full budget.} 

While SH itself chooses new pipelines $\Model_\conf$ to evaluate at random, \updates{we aim to extend on our work on \asklone{} and continue to use BO}. To do so, we follow work combining SH with BO~\citep{falkner-icml18a}.\footnote{\cite{falkner-icml18a} proposed using Hyperband~\citep{li-jmlr18a} together with BO; however, we use only SH as we expect it to work better in the extreme of having very little time, as it more aggressively reduces the budget per ML pipeline.} Specifically, we use BO to iteratively suggest new ML pipelines $\Model_\conf$, which we evaluate on the lowest budget until a fixed number of pipelines has been evaluated. Then, we run SH as described above. We are using \autosklearn{}'s standard random forest-based BO method SMAC and, according to the methodology of \cite{falkner-icml18a} build the model for BO on the highest available budget for which we have sufficient datapoints. While the original model had a mathematical requirement for $n+1$ finished pipelines, where $n$ is the number of hyperparameters to be optimized, the random forest model can guide the optimization with fewer datapoints, and we define sufficient as $\frac{n}{2}$. 
\updates{The portfolios we have introduced in Section~\ref{sec:portfolio} integrate seamlessly into this scheme: as long as not all members of the portfolio have been evaluated, we suggest them instead of asking BO for a new suggestion.}

SH potentially provides large speedups, but it could also too aggressively cut away good configurations that need a higher budget to perform best. Thus, we expect SH to work best for large datasets, for which there is not enough time to train many ML pipelines for the full budget (FB), but for which training an ML pipeline on a small budget already yields a good indication of the generalization error.

We note that SH can be used in combination with both, holdout or cross-validation, and thus indeed adds another hyper-hyperparameter to the \automl{} system, namely whether to use SH or FB. However, it also adds more flexibility to tackle a broader range of problems.

\subsection{Practical Considerations and Challenge Results}
\label{ssec:practical}

In order to make best use of the successive halving algorithm we had to do certain adjustments to obtain high performance.

First, we restricted the search space to contain only iterative algorithms and no more feature preprocessing. This simplifies the usage of SH as we only have to deal with a single type of fidelity, the number of iterations, while we would otherwise have to also consider dataset subsets as an alternative. This leaves us with extremely randomized trees~\citep{geurts-ml06a}, random forests~\citep{breimann-mlj01a}, histogram-based gradient boosting~\citep{friedman_as2001a,ke-neurips2017a}, a linear model fitted with a passive aggressive algorithm~\citep{crammer-jmlr2006a} or stochastic gradient descent and a multi-layer perceptron. The exact \configspace{} can be found in Table~\ref{tab:configspace} of the appendix.

Second, because of using only iterative algorithms, we are able to store partially fitted models to disk to prevent having no predictions in case of time- and memouts. That is, after $2, 4, 8, \dots$ iterations, we make predictions for the validation set and dump the model for later usage. 
We provide further details, such as the restricted search space, in Appendix~\ref{app:practical-considerations}.

For our submission to the second AutoML challenge, we implemented the following safeguards and tricks~\citep{feurer-automl18a}, which we do not use in this paper since we instead focus on automatically designing a robust \automl{} system:
\begin{itemize}
    \item For the submission, we also employed support vector machines using subsets of the dataset as lower fidelities. Since none of the five final ensembles in the competition contained support vector machines, we did not consider them anymore for this paper, simplifying our methodology.
    \item We developed an additional library pruning method for ensemble selection. However, in preliminary experiments, we found that this, in the best case, provides an insignificant boost for the area under curve and not balanced accuracy, which we use in this work and thus did not follow up on that any further.
    \item To increase robustness against arbitrarily large datasets, we reduced all datasets to have at most $500$ features using univariate feature selection. Similarly, we also reduced all datasets to have at most $45\ 000$ \updates{datapoints using stratified subsampling}. We do not think these are good strategies in general and only implemented them because we had no information about the dimensionality of the datasets used in the challenge, and to prevent running out of time and memory. Retrospectively, only one out of five datasets triggered this feature selection step. Now, we have instead a fallback strategy that is defined by data, see Section~\ref{ssec:partII:approach}.
    \item In case the datasets had less than $1000$ datapoints, we would have reverted from holdout to cross-validation. However, this fallback was not triggered due to the datasets being larger in the competition.
    \item We manually added a linear regression fitted with stochastic gradient descent with its hyperparameters optimized for fast runtime as the first entry in the portfolio to maximize the chances of fitting a model within the given time. We had implemented this strategy because we did not know the time limit of the competition. However, as for the paper at hand and future applications of \autosklearn{}, we expect to know the \horizon{} we are optimizing the portfolio for, we no longer require such a safeguard.
\end{itemize}

Our submission, \poshautosklearn{}, was the overall winner of the second \automl{} challenge. We give the results of the competition in Table~\ref{tab:competition_results} and refer to \cite{feurer-automl18a} and \cite{guyon-automl19a} for further details, especially for information on our competitors.

\begin{table}[tbp]
    \centering
    \small
    \begin{tabular}{lcccccc}
        \toprule
        Name & Rank & Dataset \#1 & Dataset \#2 & Dataset \#3 & Dataset \#4 & Dataset \#5 \\
        \midrule
        \poshautosklearn{} & $\mathbf{2.8}$ & $0.5533 (3)$ & $0.2839 (4)$ & $\mathbf{0.3932 (1)}$ & $\mathbf{0.2635 (1)}$ & $0.6766 (5)$ \\
        narnars0 & $3.8$ & $0.5418 (5)$ & $0.2894 (2)$ & $0.3665 (2)$ & $0.2005 (9)$ & $\mathbf{0.6922 (1)}$ \\
        Malik & $5.4$ & $0.5085 (7)$ & $0.2297 (7)$ & $0.2670 (6)$ & $0.2413 (5)$ & $0.6853 (2)$ \\
        wlWangl & $5.4$ & $\mathbf{0.5655 (2)}$ & $\mathbf{0.4851 (1)}$ & $0.2829 (5)$ & $-0.0886 (16)$ & $0.6840 (3)$ \\
        thanhdng & $5.4$ & $0.5131 (6)$ & $0.2256 (8)$ & $0.2605 (7)$ & $0.2603 (2)$ & $0.6777 (4)$ \\
        \bottomrule
    \end{tabular}
    \caption{Results for the second \automl{} challenge~\citep{guyon-automl19a}. \emph{Name} is the team name, \emph{Rank} the final rank of the submission, followed by the individual results on the five datasets used in the competition. All performances are the normalized area under the ROC curve~\citep{guyon-ijcnn15a} with the per-dataset rank in brackets. \updates{In case a rank is missing, for example, rank 1 for dataset 1, this rank was achieved by a contestant who did not place within the top 5.}
    \label{tab:competition_results}}
\end{table}

\subsection{Experimental Setup}
\label{ssec:exp_setup}

So far, \automl{} systems were designed without any \horizon{} or with a single, fixed \horizon{} $T$ in mind (see Equation~\ref{eq:minproblem}).\footnote{The OBOE \automl{} system~\citep{yang-kdd2019a} is a potential exception that takes the \horizon{} into consideration, but the experiments by~\cite{yang-kdd2019a} were only conducted for a single \horizon{}, not demonstrating that the system adapts itself to multiple \horizon{s}.}
Our system takes the \horizon{} into account when constructing the portfolio. 
We will study two \horizon{s}: a short, 10 minute \horizon{} and a long, 60 minute \horizon{} as in the original \autosklearn{} paper.
To have a single metric for binary classification, multiclass classification and unbalanced datasets, we report the balanced error rate ($1 - \text{balanced accuracy}$), following the 1${}^\text{st}$ \automl{} challenge~\citep{guyon-automl19a}. As different datasets can live on different scales, we apply a linear transformation to obtain comparable values. Concretely, we obtain the minimal and maximal error obtained by executing \autosklearn{} with portfolios and using ensembles for each combination of model selection and budget allocation strategies per dataset. Then, we rescale by subtracting the minimal error and dividing by the difference between the maximal and minimal error (ADTM, as introduced in Section~\ref{ssec:portfolio:approach}).\footnote{We would like to highlight that this is slightly different than in Section~\ref{ssec:portfolio:approach} where we did not have access to the ensemble performance and also only normalized per model selection strategy.} With this transformation, we obtain a normalized error which can be interpreted as the regret of our method.

We also limit the time and memory for each ML pipeline evaluation. For the time limit, we allow for at most $1/10$ of the \horizon{}, while for the memory, we allow the pipeline 4GB before forcefully terminating the execution.

\subsubsection{Datasets}
\label{sssec:datasets}
We require two disjoint sets of datasets for our setup: (i)~$\MetaSet$, on which we build portfolios and the \modelselector{} that we will introduce in Section~\ref{sec:partII}, and (ii)~$\AutoMLSet$, on which we evaluate our method. The distribution of both sets ideally spans a wide variety of problem domains and dataset characteristics.
For $\AutoMLSet$, we rely on $\nautodata$ datasets selected for the  \automl{} benchmark proposed by~\cite{gijsbers-automl19a}, which consists of datasets for comparing classifiers~\citep{bischl-neurips21a} and datasets from the \automl{} challenges~\citep{guyon-automl19a}.

We collected the meta datasets $\MetaSet$ based on OpenML~\citep{vanschoren-sigkdd14a} using the OpenML-Python API~\citep{feurer-jmlr21a}. To obtain a representative set, we considered all datasets on OpenML with more than $500$ and less than $1\,000\,000$ samples with at least two attributes.
Next, we dropped all datasets that are sparse, contain time attributes or string type attributes as $\AutoMLSet{}$ does not contain any such datasets.
Then, we dropped synthetic datasets and subsampled clusters of highly similar datasets.
Finally, we manually checked for overlap with $\AutoMLSet$ and ended up with a total of $\ndatasets$ training datasets and used them to train our method.

\begin{figure}[t]
    \centering
    \includegraphics[width=0.49\columnwidth]{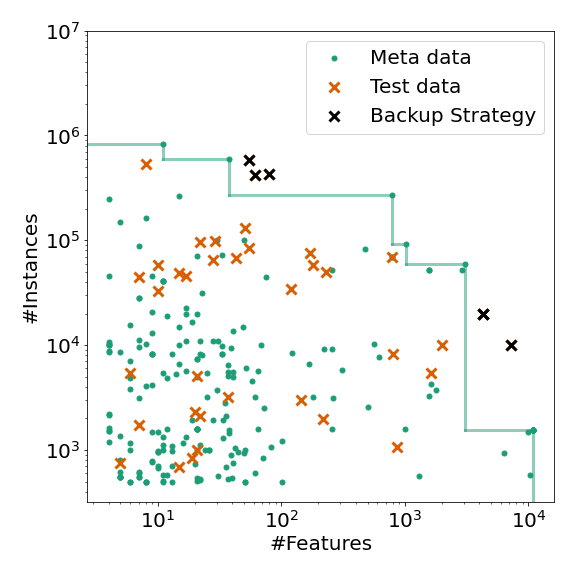}
    \caption{Distribution of meta and test datasets. We visualize each dataset w.r.t. its meta-features and highlight the datasets outside our meta distribution using black crosses.}
    \label{fig:datadist}
\end{figure}

We show the distribution of the datasets in Figure~\ref{fig:datadist}. Green points refer to $\MetaSet$ and orange crosses to $\AutoMLSet$. We can see that $\MetaSet$ spans the underlying distribution of $\AutoMLSet$ quite well, but several datasets are outside the $\MetaSet$ distribution indicated by the green lines, marked with a black cross. We give the full list of datasets for $\MetaSet$ and $\AutoMLSet$ in Appendix~\ref{app:datasets}.

For all datasets, we use a single holdout test set of $33.33\%$, which is defined by the corresponding OpenML task. The remaining $66.66\%$ are the training data of our \automl{} systems, which handle further splits for model selection themselves based on the chosen model selection strategy.

\subsubsection{Meta-data Generation}
\label{sssec:meta-data-generation}

For each \horizon{} we created four performance matrices of size $|\MetaSet| \times |\Candidates|$, see Section \ref{ssec:portfolio:approach} for details on performance matrices.
Each matrix refers to one way of assessing the generalization error of a model: holdout, 3-fold CV, 5-fold CV or 10-fold CV.
To obtain each matrix, we did the following.
For each dataset $\Dataset$ in $\MetaSet$, we used combined algorithm selection and hyperparameter optimization to find a customized ML pipeline. In practice, we ran \autosklearn{} without meta-learning and without ensemble building three times and picked the best resulting ML pipeline on the test split of $\Dataset$. To ensure that \autosklearn{} finds a good configuration, we ran it for ten times the \horizon{} given by the user (see Equation~\ref{eq:minproblem}). 
Then, we ran the cross-product of all candidate ML pipelines and datasets to obtain the performance matrix. We also stored intermediate results for the iterative algorithms so that we could build custom portfolios for SH, too.

\subsubsection{Other Experimental Details}
\label{sssec:expdetails}

We always report results averaged across $10$ repetitions to account for randomness and report the mean and standard deviation over these repetitions. 
To check whether performance differences are significant, where possible, we ran the Wilcoxon signed-rank test as a statistical hypothesis test with $\alpha=0.05$~\citep{demsar-2006a}.
In addition, we plot the average rank as follows. \updates{For each dataset, we draw one run per method (out of 10 repetitions) and rank these draws according to performance, using the average rank in case of ties. We then average over all 39 dataset and repeat this sampling $500$ times to and then plot the median and the 10th and 90th percentile of these samples. In the case of only three methods to compare, we can enumerate all $1000$ combinations of the seeds and do so. We use the exact method for Figure~\ref{fig:RQ1:plots} and the sampling method for Figure~\ref{fig:app:irr} in the appendix.}

We conducted all experiments using ensemble selection, and we constructed ensembles of size $50$ with replacement. We give results without ensemble selection in the Appendix~\ref{app:performance_wo_posthoc_ensembles}.

All experiments were conducted on a compute cluster with machines equipped with 2~\emph{Intel Xeon Gold 6242 CPUs} with 2.8GHz (32 cores) and 192 GB RAM, running Ubuntu 20.04.01. We provide scripts for reproducing all our experimental results at \url{https://github.com/automl/ASKL2.0_experiments} and provide a clean integration of our methods into the official \autosklearn{} repository.

\subsection{Experimental Results}
\label{ssec:results_posh}

In this subsection, we now validate the improvements for \poshautosklearn{}. First, we will compare using a portfolio to the previous KND approach and no warmstarting and second, we will compare \poshautosklearn{} to the previous \asklone{}.

\begin{table}[tbp]
    \centering
\begin{tabular}{l|rrr|rrr}
\toprule
{} & \multicolumn{3}{c}{10 minutes} & \multicolumn{3}{c}{60 minutes} \\
{}          &    BO & KND & Port &  BO & KND & Port \\
\midrule
FB; holdout     &  $5.98$ &  $5.29$ &       $\mathbf{3.70}$ &  $3.84$ &  $3.98$ &       $\mathbf{3.08}$ \\
SH; holdout &  $5.15$ &  $\underline{4.82}$ &       $\mathbf{4.11}$ &  $3.77$ &  $\underline{3.55}$ &       $\mathbf{3.19}$ \\
FB; 3CV         &  $8.52$ &  $\underline{7.76}$ &       $\mathbf{6.90}$ &  $6.42$ &  $6.31$ &       $\mathbf{4.96}$ \\
SH; 3CV     &  $7.82$ &  $7.67$ &       $\mathbf{6.16}$ &  $6.08$ &  $5.91$ &       $\mathbf{5.17}$ \\
FB; 5CV         &  $9.48$ &  $9.45$ &       $\mathbf{7.93}$ &  $6.64$ &  $6.47$ &       $\mathbf{5.05}$ \\
SH; 5CV     &  $9.48$ &  $\underline{8.85}$ &       $\mathbf{7.05}$ &  $6.19$ &  $5.83$ &       $\mathbf{5.40}$ \\
FB; 10CV        & $16.10$ & $\underline{15.11}$ &      $\mathbf{12.42}$ & $10.82$ & $\underline{10.44}$ &       $\mathbf{9.68}$ \\
SH; 10CV    & $16.14$ & $\underline{15.10}$ &      $\mathbf{12.61}$ & $10.54$ & $10.33$ &       $\mathbf{9.23}$ \\
\bottomrule
\end{tabular}
\caption{Averaged normalized balanced error rate. We report the aggregated performance across $10$ repetitions and $\nautodata$ datasets of our \automl{} system using only Bayesian optimization (\emph{BO}), or BO warmstarted with k-nearest-datasets (\emph{KND}) or a greedy portfolio (\emph{Port}). Per line, we boldface the best mean value (per model selection and budget allocation strategy and \horizon{}, and underline results that are not statistically different according to a Wilcoxon-signed-rank Test ($\alpha=0.05$)).
    }
    \label{tab:portfolio:results}
\end{table}

\subsubsection{Portfolio vs. KND}

Here, we study the performance of the learned portfolio and compare it against \asklone{}'s default meta-learning strategy using $25$ configurations. Additionally, we also study how pure BO would perform. We give results in Table \ref{tab:portfolio:results}. 

For the new \automl{}-hyperparameter $|\Portfolio|$, we chose $32$ to allow two full iterations of SH with our hyperparameter setting of SH.
Unsurprisingly, warmstarting, in general, improves the performance on all \horizon{}s and most model selection strategies, often by a large margin. The portfolios always improve over BO, while KND does so in all but one case. When comparing the portfolios to KND, we find that the raw results are always favorable and that for half of the settings, the differences are also significant.

\subsubsection{\poshautosklearn{} vs \asklone{}} 

We can also look at the performance of \poshautosklearn{} compared to \asklone{}. 

First, we compare the performance of \poshautosklearn{} to \asklone{} using the full search space, and we provide those numbers in Table~\ref{tab:posh-vs-1}. For both time horizons, there is a strong reduction in the loss (10min: $16.21 \rightarrow 4.11$ and 60min: $7.17 \rightarrow 3.19$), indicating that the proposed \poshautosklearn{} is indeed an improvement over the existing solution and is able to fit better machine learning models in the given time limit.

\begin{table}[t]
    \centering
    \begin{tabular}{
    @{\hskip 0cm}l@{\hskip 0.1cm}l
    c@{\hskip 0.2cm}c
    c@{\hskip 0.2cm}c}
\toprule
{} & {} & \multicolumn{2}{c}{10MIN} & \multicolumn{2}{c}{60MIN} \\
{} & {} & $\varnothing$ & std & $\varnothing$ & std \\
\midrule
(1) & PoSH-Auto-sklearn                         &   $\mathbf{4.11}$ & $0.09$ & $\mathbf{3.19}$ & $0.12$ \\
(2) & Auto-sklearn (1.0)                        & $16.21$ & $0.27$ & $\underline{7.17}$ & $0.30$ \\
\bottomrule
\end{tabular}
    \caption{Final performance of \poshautosklearn{} and \asklone{}. We report the normalized balanced error rate averaged across $10$ repetitions on $\nautodata$ datasets.  We boldface the best mean value (per \horizon{}) and underline results that are not statistically different according to a Wilcoxon signed-rank test ($\alpha=0.05$).
     }
    \label{tab:posh-vs-1}
\end{table}

Second, we compare the performance of \poshautosklearn{}~(SH; holdout and Port) to \asklone{}~(FB; holdout and KND) using only the reduced search space based on the results in Table~\ref{tab:portfolio:results}.
Again, there is a strong reduction in the loss for both time horizons (10min: $5.29 \rightarrow 4.11$ and 60min: $3.98 \rightarrow 3.19$), confirming abovementioned findings. Combined with the portfolio, the average results are inconclusive about whether our use of successive halving was the right choice or whether plain holdout would have been better. We also provide the raw numbers in Appendix~\ref{app:raw_results}, but they are inconclusive, too.

\section{Part II: Automating Design Decisions in \automl{}}
\label{sec:partII} 

The goal of \automl{} is to yield state-of-the-art performance without requiring the user to make low-level decisions, e.g., which model and hyperparameter configurations to apply. Using a portfolio and SH, \poshautosklearn{} is already an improvement over \asklone{} in terms of efficiency and scalability. However, high-level design decisions, such as choosing between cross-validation and holdout or whether to use SH or not, remain. Thus, \poshautosklearn{}, and \automl{} systems in general, suffer from a similar problem as they are trying to solve, as users have set their arguments on a per-dataset basis manually. 

\begin{figure}[t]
    \centering  
    \begin{tabular}{cc}
        \includegraphics[width=0.45\columnwidth]{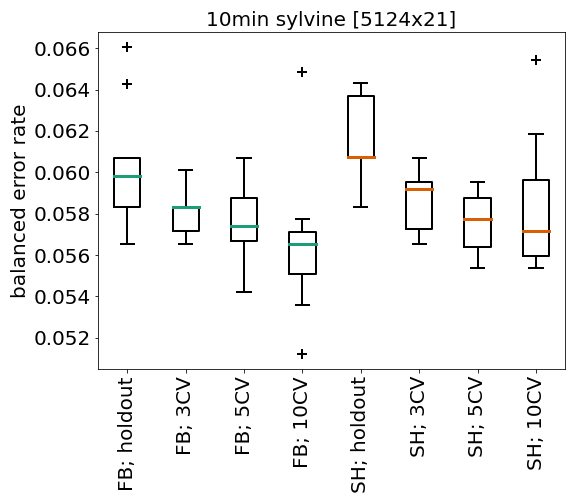} &  
        \includegraphics[width=0.45\columnwidth]{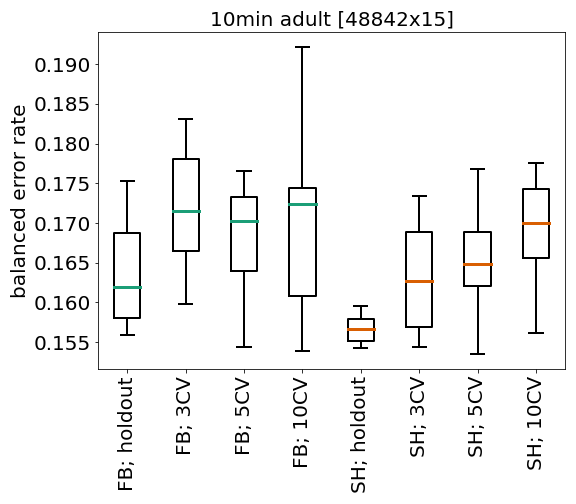} \\
        \includegraphics[width=0.45\columnwidth]{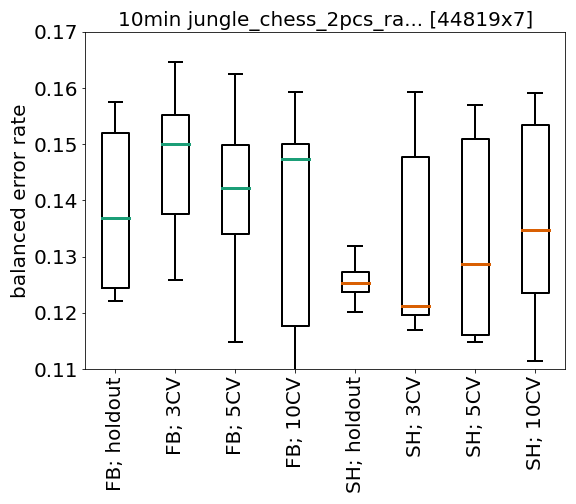} &
        \includegraphics[width=0.45\columnwidth]{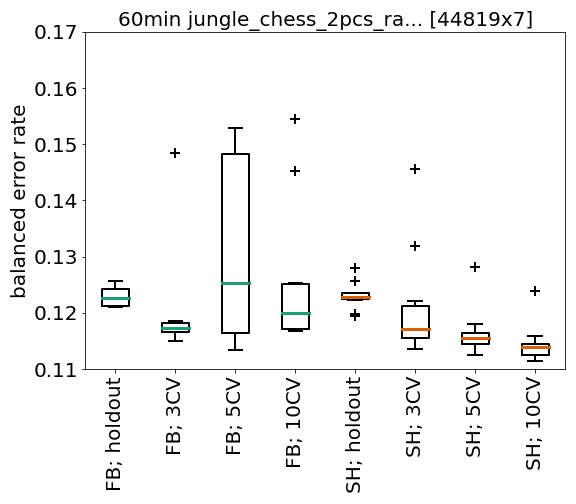} \\
    \end{tabular}
    \caption{Final balanced error rate of BO using different model selection strategies averaged across $10$ repetitions. Top row: Results for a \horizon{{}} of $10$ minutes on two different datasets. Bottom row: Results for a \horizon{} of $10$ and $60$ minutes on the same dataset.}    \label{fig:boxplots}
\end{figure}

To highlight this dilemma, in Figure~\ref{fig:boxplots} we show exemplary results comparing the balanced error rates of the best ML pipeline found by searching our \configspace{} with BO using holdout, 3CV, 5CV and 10CV with SH and FB on different \horizon{s} and datasets. 
The top row shows results obtained using the same \horizon{} of $10$ minutes on two different datasets. While \emph{FB; 10CV} is best on dataset \emph{sylvine} (top left) the same strategy on median performs amongst the worst strategies on dataset \emph{adult} (top right). Also, on \emph{sylvine}, SH performs overall slightly worse in contrast to \emph{adult}, where SH performs better on average. The bottom rows show how the given time-limit impacts the performance on the dataset \emph{jungle\_chess\_2pcs\_raw\_endgame\_complete}. Using a quite restrictive \horizon{} of $10$~minutes (bottom left), \emph{SH; 3CV}, which aggressively cuts ML pipelines on lower budgets, performs best on average. With a higher \horizon{} (bottom right), the overall results improve and more strategies become competitive.

Therefore, we propose to extend \automl{} systems with a \selector{} to automatically choose an optimization policy given a dataset (see Figure~\ref{fig:schema_askl_versions} in Section~\ref{sec:introduction} for a schematic overview). In this second part, we discuss the resulting approach, which led to \askltwo{} as the first implementation of it.

\subsection{Automated Policy Selection}
Specifically, we consider the case, where an \automl{} system can be run with different optimization policies $\policy \in \policies$ and study how to further automate \automl{} using algorithm selection on this meta-meta level. In practice, we extend the formulation introduced in Equation~\ref{eq:general_automl_empirical_ge} to not use an \automl{} system $\mathcal{A}_\pi$ with a fixed policy $\pi$, but to contain a \selector{} $\Xi : \Dataset \rightarrow \pi$:
\begin{equation}
    \reallywidehat{\gerror}(\mathcal{A},\Xi,\MetaSet) = \frac{1}{\mid\MetaSet\mid} \sum_{\dataiter = 1}^{|\MetaSet|} \reallywidehat{\gerror}(\mathcal{A}_{\Xi(\Dataset_\dataiter)}(\Dataset_\dataiter),\Dataset_\dataiter).
\end{equation}
In the remainder of this section, we describe how to construct such a \selector{}.

\subsubsection{Approach}
\label{ssec:partII:approach}

\begin{figure*}[t]
    \centering
    \includegraphics[width=0.99\textwidth]{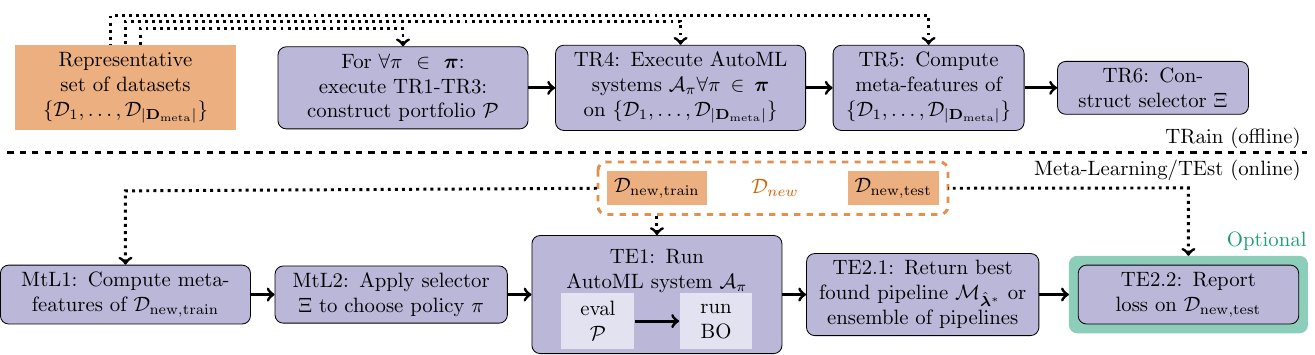}
\caption{\updates{Schematic overview of the proposed \askltwo{} system with the training phase (TR1--TR6) above and the test phase (MtL1--MtL2\&TE1--TE2) below the dashed line. Rounded, purple boxes refer to computational steps, while rectangular, orange boxes depict the input data to the \automl{} system.}}
    \label{fig:schema_main}
\end{figure*}

\automl{} systems themselves are often heavily hyperparameterized. 
In our case, we deem the model selection strategy and budget allocation strategy (see Sections~\ref{ssec:resource_allocation} and~\ref{ssec:components}) as important choices the user has to make when using an \automl{} system to obtain high performance. These decisions depend on both the given dataset and the available resources.
As there is also an interaction between the two strategies and the optimal portfolio $\Portfolio$, we consider here that the optimization policy $\policy$ is parameterized by a combination of (i) model selection strategy, (ii) budget allocation strategy and (iii) a portfolio constructed for the choice of the two strategies. In our case, these are eight different policies ($\{$3-fold CV, 5-fold CV, 10-fold CV, holdout$\}\times\{$SH, FB$\}$).

We introduce a new layer on top of \automl{} systems that automatically selects a policy~$\policy$ for a new dataset. 
We show an overview of this system in Figure~\ref{fig:schema_main} which consists of a training (TR1--TR6) and a testing stage (MtL1--2 and TE1--TE2).
\updates{In brief, in training steps TR1--TR3, we perform the same steps that we have already outlined in Figure~\ref{fig:schema_posh}. However, we now do so for each combination of model selection and budget allocation strategy.} Our policies are  combinations of a portfolio, a model selection strategy and a budget allocation strategy. We then execute the full \automl{} system for each such policy in step TR4 to obtain a realistic performance estimate. In step TR5, we compute meta-features and use them together with the performance estimate from TR4 in step TR6 to train a \modelselector{} $\Xi$, which we will use in the online test phase.

In order to not overestimate the performance of $\pi$ on a dataset $\Dataset_\dataiter$, dataset $\Dataset_\dataiter$ must not be part of the meta-data for constructing the portfolio. To overcome this issue, we perform an inner 5-fold cross-validation and build each $\pi$ on four fifths of the \emph{meta}-datasets $\MetaSet$ and evaluate it on the left-out fifth of \emph{meta}-datasets $\MetaSet$.  For the final \automl{} system we then use a portfolio built on all \emph{meta}-datasets $\MetaSet$.

For a new dataset $\Dataset_{\text{new}} \in \AutoMLSet{}$, we first compute meta-features describing $\Dataset_{\text{new}}$ (MtL1) and use the \modelselector{} from step TR6 to automatically select an appropriate policy for $\Dataset_{\text{new}}$ based on the meta-features (MtL2). This will relieve users from making this decision on their own.
Given an optimization policy $\pi$, we then apply the \automl{} system $\mathcal{A}_\pi$ to $\Dataset_{\text{new}}$ (TE1). Finally, we return the best found pipeline $\Model_{\hat{\conf}^{*}}$ based on the training set of $\Dataset_{\text{new}}$ (TE2.1). Optionally, we can then compute the loss of $\Model_{\hat{\conf}^{*}}$ on the test set of $\Dataset_{\text{new}}$ (TE2.2); we emphasize that this would be the only time we ever access the test set of $\Dataset_{\text{new}}$. Steps TE1--TE2 are the same as in Figure~\ref{fig:schema_posh}, and the only difference at evaluation time is that we use algorithm selection to decide which policy $\pi$ to use at test time instead of relying on a hand-picked one.

In the following, we describe two ways to construct a policy selector and introduce an additional backup strategy to make it robust towards failures.

\paragraph{Constructing the single best policy}

A straightforward way to construct a selector relies on the assumption that the \emph{meta}-datasets $\MetaSet$ are homogeneous and that a new dataset is similar to these. In such a case, we can use per-set algorithm selection~\citep{kerschke-evocomp2019a}, which aims to find the single algorithm that performs best on average on a set of problem instances. In our context, it aims to find the combination of model selection and budget allocation that is best on average for the given set of \emph{meta}-datasets $\MetaSet$. This single best policy is then the automated replacement for our manual selection of SH and holdout in \poshautosklearn{}. While this seems to be a trivial baseline, it actually requires the same amount of compute power as the more elaborate strategy we introduce next.

\paragraph{Constructing the per-dataset Policy Selector}

Instead of using a fixed, learned policy, we now propose to adapt the policy to the dataset at hand by using per-instance algorithm selection, which means we select the appropriate algorithm for each dataset by taking its properties into account.
To construct the meta selection model (TR6), we follow the \selector{} design of HydraMIP~\citep{xu-rcra11a}: for each pair of \automl{} policies, we fit a random forest to predict whether policy $\pi_A$ outperforms policy $\pi_B$ given the current dataset's meta-features.
Since the misclassification loss depends on the difference between the losses of the two policies (i.e. the ADTM when choosing the wrong policy), we weight each meta-observation by their loss difference.
To make errors comparable across different datasets~\citep{bardenet-icml13a},
we scale the individual error values for each dataset.
At test time (TE2), we query all pairwise models for the given meta-features and use voting for $\Xi$ to choose a policy $\pi$. We will refer to this strategy as the \textit{Policy Selector}.

To improve the performance of the \modelselector{}, we applied BO to optimize the \modelselector{}'s hyperparameters to minimize the cross-validation error~\citep{lindauer-jair15a}. We optimized in total seven hyperparameters, five of which are related to the random forest, one is how to combine the pairwise models to get a prediction, and the last one is the strategy of how to scale error values to compute weights for comparing datasets, i.e. using the raw observations, scale with $[min, max]$ / $[min, 1]$ across a pair or all policies or use the difference in ranks as the weight (see Table~\ref{tab:selector:space}). Hyperparameters are shared between all pairwise models to avoid factorial growth of the number of hyperparameters with the number of new model selection strategies. We allow a tree depth of 0\updates{, i.e., a tree with all data in a single leaf}, which is equivalent to the single best strategy described above.
\begin{table}[]
    \centering
    \begin{tabular}{lcc}
\toprule
hyperparameter & type & values \\
\midrule
Min. number of samples to create a further split & int & $[3, 20]$ \\
Min. number of samples to create a new leaf & int & $[2, 20]$ \\
Max. depth of a tree & int & $[0, 20]$ \\
Max. number of features to be used for a split & int & $[1, 2]$ \\
Bootstrapping in the random forest & cat & $\{yes, no\}$ \\
Soft or hard voting when combining models & cat & $\{soft, hard\}$ \\
Error value scaling to compute dataset weights & cat & see text \\ 
\bottomrule
    \end{tabular}
    \caption{\configspace{} of the \modelselector{}.}
    \label{tab:selector:space}
\end{table}

\paragraph{Meta-Features.}
To train our \modelselector{} and to select a policy, as well to use the backup strategy, we use meta-features~\citep{brazdil-08a,vanschoren-automl2019a} describing all meta-train datasets (TR5) and new datasets (TE1). To avoid the problems discussed in Section~\ref{sec:portfolio} we only use very simple and robust meta-features, which can be reliably computed in linear time for every dataset: 1) the number of datapoints and 2) the number of features. In fact, these are already stored as meta-data for the data structure holding the dataset. \updates{Using only these two meta-features for the selector can be regarded as learning the manually-designed fallbacks that we discussed in Section~\ref{ssec:practical}.} In our experiments, we will show that even with only these trivial and cheap meta-features, we can substantially improve over a static policy.

\paragraph{Backup strategy.}
Since there is no guarantee that our \modelselector{} will extrapolate well to datasets outside of the meta-datasets, we implement a fallback measure to avoid failures. Such failures can be harmful if a new dataset is, e.g., much larger than any dataset in the meta-dataset, and the \modelselector{} proposes to use a policy that would time out without any solution. 
More specifically, if there is no dataset in the meta-datasets that has higher or equal values for each meta-feature (i.e. dominates the dataset meta-features),
our system falls back to use \emph{holdout} with SH, which is the most aggressive and cheapest policy we consider. We visualize this in Figure~\ref{fig:datadist} where we mark datasets outside our meta distribution using black crosses.

\subsection{Experimental Results}

\begin{table}[b]
    \centering
    \begin{tabular}{
    @{\hskip 0cm}l@{\hskip 0.1cm}l
    c@{\hskip 0.2cm}c
    c@{\hskip 0.2cm}c}
\toprule
{} & {} & \multicolumn{2}{c}{10MIN} & \multicolumn{2}{c}{60MIN} \\
{} & {} & $\varnothing$ & std & $\varnothing$ & std \\
\midrule
(1) & Auto-sklearn (2.0)                        &   $\mathbf{3.58}$ & $0.23$ & $\mathbf{2.47}$ & $0.18$ \\
(2) & PoSH-Auto-sklearn                         &   $4.11$ & $0.09$ & $3.19$ & $0.12$ \\
(3) & Auto-sklearn (1.0)                        & $16.21$ & $0.27$ & $7.17$ & $0.30$ \\
\bottomrule
\end{tabular}
    \caption{Average normalized balanced error (ADTM, lower is better) of \askltwo{}, \poshautosklearn{} and \asklone{} averaged across $10$ repetitions on $\nautodata$ datasets.  We boldface the best mean value (per \horizon{}) and underline results that are not statistically different according to a Wilcoxon-signed-rank Test ($\alpha=0.05$).
     }
    \label{tab:RQ1}
\end{table}

\begin{figure}[t]
    \centering
    \includegraphics[width=\columnwidth]{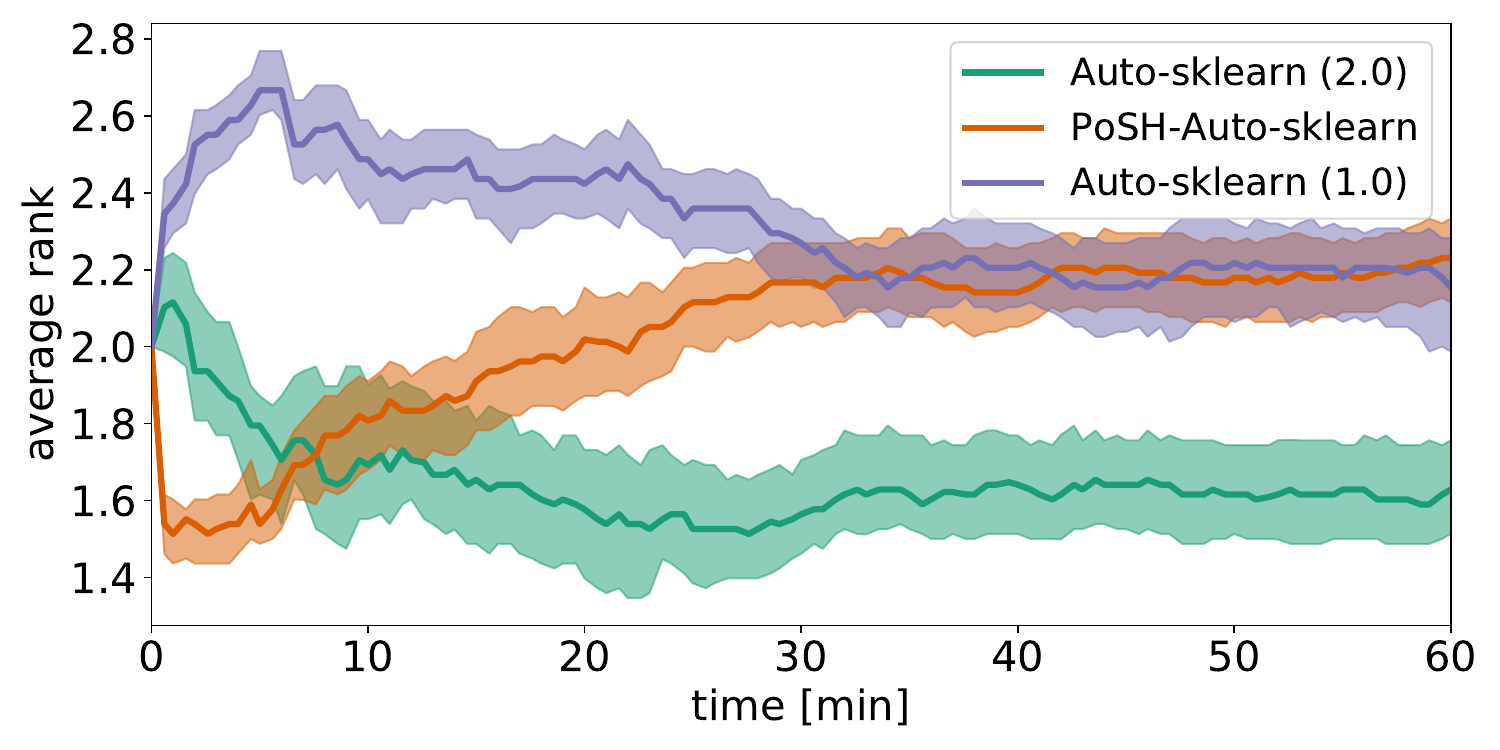}
    \caption{Performance over time. We report the \updates{median} ranks (lower is better) \updates{and the 10th and 90th percentiles} over time \updates{for \askltwo{} and the previous \automl{} systems}. \updates{Concretely, we compute the mean rank for for all $\nautodata{}$ for all $1000$ combinations of the $10$ seeds of the $3$ \automl{} systems, and compute the median and percentiles of these $1000$ average ranks.}
    }
    \label{fig:RQ1:plots}
\end{figure}

To study the performance of the \selector{}, we compare it to \poshautosklearn{} as described in Section~\ref{sec:partI} and \asklone{}. From now on we refer to \poshautosklearn{} + \selector{} as \askltwo{}. As before, we study two horizons, $10$ minutes and $60$ minutes, and use versions of \poshautosklearn{} and \askltwo{} that were constructed with these time horizons in mind. Similarly, we use the same $\ndatasets$ datasets for constructing our \automl{} systems and the same $\nautodata$ for evaluating them.

Looking at Table~\ref{tab:RQ1}, we see that \askltwo{} achieves the lowest error, being significantly better for both  \horizon{s}.
Most notably, \askltwo{} reduces the relative error compared to \asklone{} by $78\%$ (10MIN) and $65\%$, respectively, which means a reduction by a factor of $4.5$ and three.

It turns out that these results are skewed by several large datasets (task IDs \href{https://www.openml.org/t/189873}{189873} and \href{https://www.openml.org/t/75193}{75193}
for both horizons;  \href{https://www.openml.org/t/189866}{189866}, \href{https://www.openml.org/t/189874}{189874}, \href{https://www.openml.org/t/168796}{168796} and \href{https://www.openml.org/t/168797}{168797} only for the ten minutes horizon) 
on which the KND initialization of \asklone{} only suggests ML pipelines that time out or hit the memory limit and thus exhaust the optimization budget for the full \configspace{}. 
Our new \automl{} system does not suffer from this problem as it a) selects SH to avoid spending too much time on unpromising ML pipelines and b) can return predictions and results even if an ML pipeline was not evaluated for the full budget or converged early; and even after removing the datasets in question from the average, the performance of \asklone{} is substantially worse than that \askltwo{}.

When looking at the intermediate system, i.e. \poshautosklearn{}, we find that it outperforms \asklone{} in terms of the normalized balanced error rate, but that the additional step of selecting the model selection and budget allocation strategy gives \askltwo{} an edge. When not considering the large datasets \asklone{} failed on, their performance becomes very similar.

Figure~\ref{fig:RQ1:plots} provides another view on the results, presenting average ranks (where failures obtain less weight compared to the averaged performance).  
\askltwo{} is still able to deliver best results, \poshautosklearn{} should be preferred to \asklone{} for the first 30 minutes and then converges to roughly the same ranking.

\subsection{Ablation}
\label{ssec:exp:ablation} 

Now, we study the contribution of each of our improvements in an ablation study. We iteratively disable one component and compare the performance to the entire system using the $\nautodata$ datasets from the \automl{} benchmark as done in the previous experimental sections. These components are (1)~using a per-dataset \modelselector{} to choose a policy, (2)~using only a subset of the available policies, and (3)~warmstarting BO with a portfolio. 

\subsubsection{Do we need per-dataset selection?}

We first examine how much performance we gain by having a \modelselector{} to decide between different \automl{} strategies based on meta-features and how to construct this \modelselector{}, or whether it is sufficient to select a single strategy based on meta-training datasets. 
We compare the performance of the entire system using a \modelselector{} to using a single, static strategy (single best) and both, the \modelselector{} and the single best, without the fallback mechanism for out-of-distribution datasets and give all results in Table~\ref{tab:RQ2:selector}. We also provide two additional baselines: a random baseline, which randomly assigns a policy to a run and an oracle baseline, which marks the lowest possible error that can be achieved by any of the policies.\footnote{We would like to note that the oracle performance can be unequal to zero because we normalize the results using the single best test loss found for a single model to normalize the results. When evaluating the best policy on a dataset, this most likely results in selecting a model on the validation set that is not the single best model on the test set we use to normalize data.}

First, we compare the performance of the \modelselector{} with the single best. We can observe that for 10 minutes, there is a slight improvement in terms of performance, while the performance for 60 minutes is almost equal. While there is no significant difference to the single best for 10 minutes, there is for 60 minutes. These numbers can be compared with Table~\ref{tab:portfolio:results} to see how we fare against picking a single policy by hand. We find that our proposed algorithm selection compares favorably, especially for the longer time horizon.

Second, to study how much resources we need to spend on generating training data for our \modelselector{}, we consider three approaches: (P) only using the portfolio performance which we pre-computed and stored in the performance matrices as described in Section~\ref{ssec:portfolio:approach}, (P+BO) actually running \autosklearn{} using the portfolio and BO for $10$ and $60$ minutes, respectively, and (P+BO+E) additionally also constructing ensembles, which yields the most realistic meta-data. Running BO on all $\ndatasets$ datasets (P+BO) is by far more expensive than the table lookups (P); building an ensemble (P+BO+E) adds only several seconds to minutes on top compared to (P+BO).

For both \horizon{s} using P+BO yields the best results using the \modelselector{} closely followed by P+BO+ENS, see Table~\ref{tab:RQ2:selector}. The cheapest method, P, yields the worst results showing that it is worth investing resources into computing good meta-data. 
Surprisingly, looking at the single best, performance gets worse when using seemingly better meta-data. We investigated the reason why P+BO performs slightly better than P+BO+ENS. When using a \modelselector{}, this can be explained by a single dataset for both time horizons for which the policy chosen by the \modelselector{} is worse than the single best policy. When looking at the single best, there is no single dataset which stands out. To summarize, investing additional resources to compute realistic meta-data results in improved performance, but so far, it appears that having the effect of BO in the meta-data is sufficient, while the ensemble seems to lead to lower meta-data quality.

\begin{table}[t]
    \centering
    \begin{tabular}{@{\hskip 0cm}l
    c@{\hskip 0.15cm}c@{\hskip 0.15cm}c|
    c@{\hskip 0.15cm}c@{\hskip 0.15cm}c
    @{\hskip 0cm}}
    \toprule
    {} & \multicolumn{3}{c}{10 Min} & \multicolumn{3}{c}{60 Min} \\

    \textit{\scriptsize{trained on}} & \scriptsize{P} & \scriptsize{P+BO} & \scriptsize{P+BO+E} &  \scriptsize{P} & \scriptsize{P+BO} & \scriptsize{P+BO+E} \\
    \midrule
    \modelselector{} & $\underline{3.58}$ & $\textbf{\underline{3.56}}$ & $\underline{3.58}$ 
    & $2.53$ & $\textbf{\underline{2.32}}$ & $\underline{2.47}$ \\
    \modelselector{} w/o fallback & $\underline{5.43}$ & $\underline{5.68}$ & $\underline{4.79}$ 
    & $4.98$ & $5.36$ & $\underline{5.43}$ \\
    single best & $3.88$ & $\underline{3.67}$ & $\underline{3.69}$ 
    & $2.49$ & $\underline{2.38}$ & $2.44$ \\
    single best w/o fallback & $5.18$ & $\underline{6.38}$ & $\underline{6.40}$ & $5.10$ & $5.01$ & $5.07$ \\
    \midrule
    oracle & \multicolumn{3}{c}{$2.33$} &  \multicolumn{3}{c}{$1.22$} \\
    random & \multicolumn{3}{c}{$8.32$} &  \multicolumn{3}{c}{$6.18$} \\
    \bottomrule
    \end{tabular}
    \caption{Average normalized balanced error (ADTM, lower is better) for $10$ and $60$ minutes. We report the performance for the \textit{\modelselector{}} policy and the \textit{single best} when trained on different data obtained on $\MetaSet{}$ (P = Portfolio, BO = Bayesian Optimization, E = Ensemble) as well as the \textit{\modelselector{} without the fallback}. The second part of the table shows the results of always choosing the best policy on the test set (\textit{oracle}) and results for choosing a random policy (\textit{random}) as baselines.
    We boldface the best mean value (per \horizon{}) and underline results that are not statistically different according to a Wilcoxon-signed-rank Test ($\alpha=0.05$).
    }
    \label{tab:RQ2:selector}
\end{table}

Finally, we also take a closer look at the impact of the fallback mechanism to verify that our improvements are not solely due to this component. We observe that the performance drops for all policy selection strategies \updates{that do not use the fallback mechanism}. For the shorter 10 minutes setting, we find that the \modelselector{} still outperforms the single best, while for the longer 60 minutes setting, the single best leads to better performance. The rather stark performance degradation compared to the regular \modelselector{} can mainly be explained by a few, huge datasets, to which the \modelselector{} cannot extrapolate (and which the single best does not account for).
Based on these observations, we suggest research into an adaptive fallback strategy that can change the model selection strategy during the execution of the \automl{} system so that a \selector{} can be used on out-of-distribution datasets. We conclude that using a \modelselector{} is beneficial, and using a fallback strategy to cope with out-of-distribution datasets can substantially improve performance.

\subsubsection{Do we need different model selection strategies?}
%
\begin{table}[t]
    \centering
\begin{tabular}{ll|cc|cc|cc}
\toprule
\multicolumn{2}{c}{} & \multicolumn{2}{c}{Selector} & \multicolumn{2}{c}{Random} & \multicolumn{2}{c}{Oracle} \\
{} & {} & $\varnothing$ & std & $\varnothing$ & std & $\varnothing$ & std \\
\midrule
\multirow{5}{*}{10 Min} & All &  3.58 &    0.23 &  7.46 &    2.02 & \textbf{2.33} & 0.06 \\
& Only Holdout          &  4.03 &    0.14 &  \textbf{3.78} &    0.23 & 3.23 & 0.10 \\
& Only CV      &  6.11 &    0.11 &  8.66 &    0.70 & 5.28 & 0.06 \\
& Only FB & \textbf{3.50} & 0.20 & 7.64 & 2.00 & 2.59 & 0.09 \\
& Only SH & 3.63 & 0.19 & 6.95 & 1.98 & 2.75 & 0.07 \\
\midrule
\multirow{5}{*}{60 Min} & All &  2.47 &    0.18 &  5.64 &    1.95 & \textbf{1.22} & 0.08\\
& Only Holdout          &  3.18 &    0.15 &  \textbf{3.13} &    0.12 & 2.62 & 0.07\\
& Only CV      &  5.09 &    0.19 &  6.85 &    0.86 & 3.94 & 0.10 \\
& Only FB & \textbf{2.39} & 0.18 & 5.46 & 1.52 & 1.51 & 0.06 \\
& Only SH & 2.44 & 0.24 & 5.13 & 1.72 & 1.68 & 0.12 \\
\bottomrule
\end{tabular}
    \caption{Average Normalized balanced error (ADTM, lower is better) for the full system and when not considering all model selection strategies.
    }
    \label{tab:RQ2:mss}
\end{table}
Next, we study whether we need the different model selection strategies. For this, we 
build \modelselector{}s on different subsets of the available eight combinations of model selection strategies and budget allocations: $\{$3-fold CV, 5-fold CV, 10-fold CV, holdout$\}\times\{$SH, FB$\}$. \emph{Only Holdout} consists of holdout with SH or FB (2 combinations), \emph{Only CV} comprises 3-fold CV, 5-fold CV and 10-fold CV, all of them with SH or FB (6 combinations), \emph{FB} contains both holdout and cross-validation and assigns each pipeline evaluation the same budget (4 combinations) and \emph{Only SH} uses SH to assign budgets (4 combinations).

In Table~\ref{tab:RQ2:mss}, we compare the performance of selecting a policy at random (random), the performance of selecting the best policy on the test set and thus giving a lower bound on the ADTM (oracle) and our \modelselector{}. The oracle indicates the best possible performance with each of these subsets of model selection strategies. It turns out that both \emph{Only Holdout} and \emph{Only CV} have a much worse oracle performance than \emph{All}, with the oracle performance of \emph{Only CV} being even worse than the performance of the \modelselector{} for \emph{All}. 
Looking at \emph{Full budget} (FB), it turns out that this subset would be slightly preferable in terms of performance with a \selector{}. However, the oracle performance is worse than that of \emph{All} which shows that there is some complementarity between the different policies which cannot yet be exploited by the \selector{}.
For \emph{Only Holdout}, surprisingly, the random \selector{} performs slightly better than the \modelselector{}. We attribute this to the fact that holdout with both SH and FB performs similarly and that the choice between these two cannot yet be learned, possibly also indicated by the close performance of the random selector.

These results show that a large variety of available model selection strategies to choose from increases best possible performances. However, they also show that a \modelselector{} cannot yet necessarily leverage this potential. This questions the usefulness of choosing from all model selection strategies, similar to a recent finding which proves that increasing the number of different policies a \selector{} can choose from leads to reduced generalization~\citep{balcan-arxiv2020a}. However, we believe this points to the research question of whether we can learn on the meta-datasets which model selection and budget allocation strategies to include in the set of strategies to choose from. Also, with an ever-growing availability of meta-datasets and continued research on robust \selector{}s, we expect this flexibility to eventually yield improved performance.

\subsubsection{Do we still need to warm-start Bayesian optimization?}

\begin{table}[t]
    \centering
    \begin{tabular}{ll|cc|cc}
    \toprule
    {} & {} & \multicolumn{2}{c}{10min} & \multicolumn{2}{c}{60min} \\
    {} & {} & $\varnothing$ & std & $\varnothing$ & std \\
    \midrule
    \multirow{2}{*}{With Portfolio} & Policy selector    &  \textbf{3.58} &    0.23 &  \underline{2.47} &    0.18 \\
    & Single best     &  \underline{3.69} &    0.14 &  \textbf{2.44} &    0.12 \\
    \midrule
    \multirow{2}{*}{Without Portfolio} & Policy selector &  5.63 &    0.89 &  3.42 &    0.32 \\
    & Single best  &  5.37 &    0.58 &  3.61 &    0.61 \\
    \bottomrule
    \end{tabular}
    \caption{Average normalized balanced error (ADTM, lower is better) after $10$ and after $60$ minutes with portfolios (top) and without (bottom). The row "with portfolio" and "policy selector" constitutes the full \automl{} system including portfolios, BO and ensembles) and the row "without portfolios" and "\selector{}" only removes the portfolios (both from the meta-data for \modelselector{} construction and at runtime).
    We boldface the best mean value (per \horizon{}) and underline results that are not statistically different according to a Wilcoxon-signed-rank Test ($\alpha=0.05$).
    }
    \label{tab:RQ2:port}
\end{table}

Last, we analyze the impact of the portfolio. Given the other improvements, we now discuss whether we still need to add the additional complexity and invest resources to warm-start BO (and can therefore save the time to build the performance matrices to construct the portfolios). For this study, we completely remove the portfolio from our \automl{} system, meaning that we directly start with BO and construct ensembles -- both for creating the data we train our \selector{} on and for reporting performance. We report the results in Table~\ref{tab:RQ2:port}.

Comparing the performance of an \automl{} system with a \modelselector{} with and without portfolios (Row 1 and 3), there is a clear drop in performance when disabling the portfolios.
Comparing Rows 2 and 4 also demonstrates that a portfolio is necessary when using the single best policy.
This ablation highlights the importance of initializing the search procedure of \automl{} systems with well-performing pipelines.

\section{Comparison to other \automl{} systems}
\label{sec:partIII} 
Having established that \askltwo{} does indeed improve over \asklone{}, we now compare our system to other well established \automl{} systems. For this, we use the publicly available \automl{} benchmark suite which defines a fixed benchmarking environment for \automl{} systems~\citep{gijsbers-automl19a} comparisons. We use the original implementation of the benchmark and compare \asklone{} and \askltwo{} to the provided implementations of \autoweka{}~\citep{thornton-kdd13a}, \tpot{}~\citep{olson-gecco16a,olson-evo16a}, \hto{} \automl{}~\citep{ledell-automl20a} and a random forest baseline with hyperparameter tuning on $39$ datasets as implemented by the benchmark suite. These $\nautodata$ datasets are the same datasets as in $\AutoMLSet$ and we provide details in Table~\ref{tab:setcharactertest} in the appendix. 

\subsection{Integration and setup}

To avoid hardware-dependent performance differences, we (re-)ran all \automl{} systems on our local hardware (see Section~\ref{sssec:expdetails}). We used the pre-defined \emph{1h8c} setting, which divides each dataset into ten folds and gives each framework one hour on eight CPU cores to produce a final model. We furthermore assigned each run $32$GB of RAM, which a SLURM cluster manager controlls. In addition, we conducted five repeats to account for randomness. The benchmark comes with \emph{Docker} containers~\citep{merkel-linux14a}. However, \emph{Docker} requires superuser access on the execution nodes, which is not available on our compute cluster. Therefore, we extended the \automl{} benchmark with support for \emph{Singularity} images~\citep{kurtzer-plos17a}, and used them to isolate the framework installations from each other. For reproducibility, we give the exact versions we used in Table~\ref{tab:automl:versions} in the Appendix.

The default resource allocation of the \automl{} benchmark is a highly parallel setting with eight cores. We chose the most straightforward way of making use of these resources for \autosklearn{} and evaluated eight ML pipelines in parallel, assigning each $total\_memory/num\_cores$ RAM, which are $4$GB. This allows us to evaluate configurations obtained from the portfolio or KND in parallel but also requires a parallel strategy for running BO afterwards.
We extended the Bayesian optimization package SMAC3~\citep{lindauer-jmlr22a} to allow for asynchronous parallel optimization.
In preliminary experiments, we found that the inherent randomness of the random forest used by SMAC combined with the interleaved random search of SMAC is sufficient to obtain results that perform a lot better than the previous parallelism implemented in \autosklearn{} via SMAC~\citep{ramage-thesis2015a}. 
Whenever a pipeline finishes training, \autosklearn{} checks whether there is an instance of the ensemble construction running, and if not, it uses one of the eight slots to conduct ensemble building and otherwise continues to fit a new pipeline. We implemented this version of parallel \autosklearn{} using \emph{Dask}~\citep{dask}.

\subsection{Results}

\begin{table}[]
    \centering
    \footnotesize
    \begin{tabular}{lrrrrrr}
\toprule
{} &  AS 2.0 &  AS 1.0 &  AW &              TPOT &               H2O &           TunedRF \\
\midrule
adult             &          $0.0692$ &          $0.0701$ &  $0.0920$ &          $0.0750$ & $\mathbf{0.0690}$ &          $0.0902$ \\
airlines          &          $0.2724$ &          $0.2726$ &  $0.3241$ &          $0.2758$ & $\mathbf{0.2682}$ &                 - \\
albert            &          $0.2413$ & $\mathbf{0.2381}$ &         - &          $0.2681$ &          $0.2530$ &          $0.2616$ \\
amazon            &          $0.1233$ &          $0.1412$ &  $0.1836$ &          $0.1345$ & $\mathbf{0.1218}$ &          $0.1377$ \\
apsfailure        &          $0.0085$ & $\mathbf{0.0081}$ &  $0.0365$ &          $0.0099$ &          $0.0081$ &          $0.0087$ \\
australian        & $\mathbf{0.0594}$ &          $0.0702$ &  $0.0709$ &          $0.0670$ &          $0.0607$ &          $0.0610$ \\
bank-marketing    & $\mathbf{0.0607}$ &          $0.0616$ &  $0.1441$ &          $0.0664$ &          $0.0610$ &          $0.0692$ \\
blood-transfusion & $\mathbf{0.2428}$ &          $0.2474$ &  $0.2619$ &          $0.2761$ &          $0.2430$ &          $0.3122$ \\
car               & $\mathbf{0.0012}$ &          $0.0046$ &  $0.1910$ &          $2.7843$ &          $0.0032$ &          $0.0421$ \\
christine         &          $0.1821$ & $\mathbf{0.1703}$ &  $0.2026$ &          $0.1821$ &          $0.1763$ &          $0.1908$ \\
cnae-9            & $\mathbf{0.1424}$ &          $0.1779$ &  $0.7045$ &          $0.1483$ &          $0.1807$ &          $0.3119$ \\
connect-4         &          $0.3387$ &          $0.3535$ &  $1.7083$ &          $0.3856$ & $\mathbf{0.3127}$ &          $0.4777$ \\
covertype         & $\mathbf{0.1103}$ &          $0.1435$ &  $3.3515$ &          $0.5332$ &          $0.1281$ &                 - \\
credit-g          &          $0.2031$ &          $0.2159$ &  $0.2505$ &          $0.2144$ &          $0.2078$ & $\mathbf{0.1985}$ \\
dilbert           &          $0.0399$ & $\mathbf{0.0332}$ &  $2.0791$ &          $0.1153$ &          $0.0359$ &          $0.3283$ \\
dionis            & $\mathbf{0.5620}$ &          $0.7171$ &         - &                 - &          $4.7758$ &                 - \\
fabert            &          $0.7386$ &          $0.7466$ &  $5.4784$ &          $0.8431$ & $\mathbf{0.7274}$ &          $0.8060$ \\
fashion-mnist     & $\mathbf{0.2511}$ &          $0.2524$ &  $0.9505$ &          $0.4314$ &          $0.2762$ &          $0.3613$ \\
guillermo         &          $0.0945$ & $\mathbf{0.0871}$ &  $0.1251$ &          $0.1680$ &          $0.0911$ &          $0.0973$ \\
helena            & $\mathbf{2.4974}$ &          $2.5432$ & $14.3523$ &          $2.8738$ &          $2.7578$ &                 - \\
higgs             & $\mathbf{0.1824}$ &          $0.1846$ &  $0.3379$ &          $0.1969$ &          $0.1846$ &          $0.1966$ \\
jannis            &          $0.6709$ & $\mathbf{0.6637}$ &  $2.9576$ &          $0.7244$ &          $0.6695$ &          $0.7288$ \\
jasmine           &          $0.1141$ &          $0.1196$ &  $0.1356$ &          $0.1123$ &          $0.1141$ & $\mathbf{0.1118}$ \\
jungle\_chess      &          $0.2104$ &          $0.1956$ &  $1.6969$ &          $0.9557$ & $\mathbf{0.1479}$ &          $0.4020$ \\
kc1               &          $0.1611$ &          $0.1594$ &  $0.1780$ & $\mathbf{0.1530}$ &          $0.1745$ &          $0.1590$ \\
kddcup09          & $\mathbf{0.1580}$ &          $0.1632$ &         - &          $0.1696$ &          $0.1636$ &          $0.2058$ \\
kr-vs-kp          & $\mathbf{0.0001}$ &          $0.0003$ &  $0.0217$ &          $0.0003$ &          $0.0002$ &          $0.0004$ \\
mfeat-factors     & $\mathbf{0.0726}$ &          $0.0901$ &  $0.5678$ &          $0.1049$ &          $0.1009$ &          $0.2091$ \\
miniboone         & $\mathbf{0.0121}$ &          $0.0128$ &  $0.0352$ &          $0.0177$ &          $0.0129$ &          $0.0183$ \\
nomao             & $\mathbf{0.0035}$ &          $0.0039$ &  $0.0157$ &          $0.0047$ &          $0.0036$ &          $0.0049$ \\
numerai28.6       &          $0.4696$ &          $0.4705$ &  $0.4729$ &          $0.4741$ & $\mathbf{0.4695}$ &          $0.4792$ \\
phoneme           & $\mathbf{0.0299}$ &          $0.0366$ &  $0.0416$ &          $0.0307$ &          $0.0325$ &          $0.0347$ \\
riccardo          &          $0.0002$ & $\mathbf{0.0002}$ &  $0.0020$ &          $0.0021$ &          $0.0003$ &          $0.0002$ \\
robert            &          $1.4302$ & $\mathbf{1.3800}$ &         - &          $1.8600$ &          $1.4927$ &          $1.6877$ \\
segment           & $\mathbf{0.1482}$ &          $0.1749$ &  $1.2497$ &          $0.1660$ &          $0.1580$ &          $0.1718$ \\
shuttle           & $\mathbf{0.0002}$ &          $0.0004$ &  $0.0100$ &          $0.0008$ &          $0.0004$ &          $0.0006$ \\
sylvine           &          $0.0105$ &          $0.0091$ &  $0.0290$ & $\mathbf{0.0075}$ &          $0.0106$ &          $0.0159$ \\
vehicle           &          $0.3341$ &          $0.3754$ &  $2.0662$ &          $0.4402$ & $\mathbf{0.3067}$ &          $0.4839$ \\
volkert           & $\mathbf{0.7477}$ &          $0.7862$ &  $3.4235$ &          $0.9852$ &          $0.8121$ &          $0.9792$ \\
\midrule
Rank  & $\mathbf{1.79}$ & $2.64$ & $5.72$ & $4.08$ & $2.38$ & $4.38$ \\
\midrule
Best performance & $\mathbf{19}$ & $8$ & $0$ & $2$ & $8$ & $2$ \\
Wins/Losses/Ties of AS 2.0 & - & $28/11/0$ & $39/0/0$ & $35/4/0$ & $26/13/0$ & $36/3/0$ \\
P-values (AS 2.0 vs.\ other methods), & - & $.009$ & $<.000$ & $<.000$ & $.053$ & $<.000$ \\
based on a Binomial sign test & \\
\bottomrule
\end{tabular}
\caption{Results of the \automl{} benchmark averaged across five repetitions. We report log loss for multiclass datasets and $1 - AUC$ for binary classification datasets (lower is better). AS is short for \autosklearn{} and AW for \autoweka{}. \autosklearn{} has the best overall rank, the best performance in most datasets and, based on the P-values of a Binomial sign test, we gain further confidence in its strong performance.
    \label{tab:automlbench}}
\end{table}

We give results for the \automl{} benchmark in Table~\ref{tab:automlbench}. For each dataset, we give the average performance of the \automl{} systems across all ten folds and five repetitions and boldface the one with the lowest error (we cannot give any information about whether differences are significant as we cannot compute significances on cross-validation folds as described by \citealp{bengio-jmlr2004a}).

We report the log loss for multiclass datasets and $1 - AUC$ for binary datasets (lower is better). In addition, we provide the average rank as an aggregate measure (computed by averaging all folds and repetitions per dataset and then computing the rank). Furthermore, we count how often each framework is the winner on a dataset (champion), and give the losses, wins and ties against \askltwo{}. We then use these to perform a binomial sign test~\citep{demsar-2006a} to compare the individual algorithms against \askltwo{}.

The results in Table~\ref{tab:automlbench} show that none of the \automl{} systems is best on all datasets, and even the TunedRF performs best on a few datasets. However, we can also observe that the proposed \askltwo{} has the lowest average rank. It is followed by \hto{} \automl{} and \asklone{} which perform roughly en par wrt the ranking scores and the number of times they are the winner on a dataset. According to both aggregate metrics, the TunedRF, \autoweka{} and \tpot{} cannot keep up and lead to substantially worse results. 
Finally, both versions of \autosklearn{} appear to be quite robust as they reliably provide results on all datasets, including the largest ones where several of the other methods fail.

\section{Related Work} \label{sec:related}

We now present related work on our individual contributions (portfolios, model selection strategies, and algorithm selection) as well as on related \automl{} systems.

\subsection{Related Work on Portfolios}

Portfolios were introduced for hard combinatorial optimization problems, where the runtime between different algorithms varies drastically and allocating time shares to multiple algorithms instead of allocating all available time to a single one reduces the average cost for solving a problem~\citep{huberman-science97a,gomes-aij01a}, and had applications in different sub-fields of AI~\citep{smithmiles-acmcs08a,kotthoff-aim14a,kerschke-evocomp2019a}. 
Algorithm portfolios were introduced to ML 
by the name of \emph{algorithm ranking}
to reduce the required time to perform model selection compared to running all algorithms under consideration~\citep{brazdil_ecml2000a,soares-pkdd2000}, ignoring redundant ones~\citep{brazdil_reducing_2001}.
ML portfolios can be superior to hyperparameter optimization with Bayesian optimization~\citep{wistuba-icdm2015a}, Bayesian optimization with a model which takes past performance data into account~\citep{wistuba-dsaa15a} or can be applied when there is simply no time to perform full hyperparameter optimization~\citep{feurer-automl18a}.
Furthermore, such a portfolio-based model-free optimization is both easier to implement than regular Bayesian optimization and meta-feature based solutions,
and the portfolio can be shared easily across researchers and practitioners without the necessity of sharing meta-data~\citep{wistuba-dsaa15a,wistuba-icdm2015a,pfisterer-arxiv2018a} or additional hyperparameter optimization software.
Here, our goal is to have strong hyperparameter settings when there is no time to optimize with a typical black-box algorithm.

The efficient creation of algorithm portfolios is an active area of research with the Greedy Algorithm being a popular choice~\citep{xu-aaai10a,xu-rcra11a,seipp-aaai15a,wistuba-icdm2015a,lindauer-aij17a,feurer-automl18a,feurer-automl18b} due to its simplicity. 
\cite{wistuba-icdm2015a} first proposed the use of the Greedy Algorithm for pipelines of ML portfolios, minimizing the average rank on meta-datasets for a single ML algorithm. Later, they extended their work to update the members of a portfolio in a round-robin fashion, this time using the average normalized misclassification error as a loss function and relying on a Gaussian process model~\citep{wistuba-dsaa15a}. The loss function of the first method does not optimize the metric of interest, while the second method requires a model and does not guarantee that well-performing algorithms are executed early on, which could be harmful under time constraints.

Research into the Greedy Algorithm continued after our submission to the second \automl{} challenge and the publication of the employed methods~\citep{feurer-automl18a}.  \cite{pfisterer-arxiv2018a} suggested using a set of default values to simplify hyperparameter optimization. They argued that constructing an optimal portfolio of hyperparameter settings is a generalization of the \emph{Maximum  coverage  problem} and propose two solutions based on \emph{Mixed Integer Programming} and the \emph{Greedy Algorithm} which we also use as the base of our algorithm.
The greedy algorithm recently also drew interest in deep learning research, where it was applied in its basic form for tuning the hyperparameters of the popular ADAM algorithm~\citep{metz-arxiv20a}.

Extending these portfolio strategies, which are learned offline, there are online portfolios that can select from a fixed set of machine learning pipelines, taking previous evaluations into account~\citep{leite-mldmpr12a,wistuba-dsaa15a,wistuba-icdm2015a,fusi-neurips2018a,yang-kdd2019a,yang-kdd2020a}. However, such methods cannot be directly combined with all budget allocation strategies as they require the definition of a special model for extrapolating learning curves~\citep{klein-iclr17a,falkner-icml18a} and also introduce additional complexity into \automl{} systems.

There exists other work on building portfolios without prior discretization (which we do for our work and was done for most work mentioned above), which directly optimizes the hyperparameters of ML pipelines to add next to the portfolio in a greedy fashion~\citep{xu-aaai10a,xu-rcra11a,seipp-aaai15a}, to jointly optimize all configurations of the portfolio with global optimization~\citep{winkelmolen-arxiv2020a}, and also to build parallel portfolios~\citep{lindauer-aij17a}. We consider these to be orthogonal to using portfolios in the first place and plan to study improved optimization strategies in future work.

\subsection{Related Work on Successive Halving}\label{ssec:shbackground}

Large datasets, expensive ML pipelines and tight resource limitations demand sophisticated methods to speed up pipeline selection. One line of research, multi-fidelity optimization methods, tackle this problem by using cheaper approximations of the objective of interest. Practical examples are evaluating a pipeline only on a subset of the dataset or for iterative algorithms limiting the number of iterations. There exists a large body of research on optimization methods leveraging lower fidelities, for example working with a fixed set of auxiliary tasks~\citep{forrester-prs07a,swersky-nips13a,poloczek-nips17a,moss-ecml20a}, solutions for specific model classes~\citep{swersky-arxiv14a,domhan-ijcai15a,chandrashekaran-ecml17a} and selecting a fidelity value from a continuous range~\citep{klein-aistats17a,kandasamy-icml17a,wu-uai20a,takeno-icml20a}. Here, we focus on a methodologically simple bandit strategy, SH~\citep{karnin-icml13a,jamieson-aistats16a}, which successively reduces the number of candidates and at the same time increases the allocated resources per run till only one candidate remains. Our use of SH in the 2nd \automl{} challenge also inspired work on combining a genetic algorithm with SH~\citep{parmentier-ictai2019a}. Another way of quickly discarding unpromising pipelines is the intensify procedure which was used by \autoweka{}~\citep{thornton-kdd13a} to speed up cross-validation. Instead of evaluating all folds at once, it evaluates the folds in an iterative fashion. After each evaluation, the average performance on the evaluated folds is compared to the performance of the so-far best pipeline on these folds. The evaluation is only continued if the performance is equal or better. While this allows evaluating many configurations in a short time, it cannot be combined with post-hoc ensembling and reduces the cost of a pipeline to, at most, the cost of holdout, which might already be too high.

\subsection{Related Work on Algorithm Selection}

Automatically choosing a model selection strategy to assess the performance of an ML pipeline for hyperparameter optimization has not previously been tackled, and only \cite{guyon-ijcnn15a} acknowledge the lack of such an approach. However, treating the choice of model selection strategy as an algorithm selection problem allows us to apply methods from the field of algorithm selection~\citep{smithmiles-acmcs08a,kotthoff-aim14a,kerschke-evocomp2019a} and we can in future work reuse existing techniques besides the pairwise classification we employ in this paper~\citep{xu-rcra11a}, such as the AutoAI system AutoFolio~\citep{lindauer-jair15a}.

\subsection{Background on \automl{} Systems and Their Components}
\label{ssec:backautoml} 

\automl{} systems have recently gained traction in the research community, and there exists a multitude of approaches, often accompanied by open-source software. In the following, we provide background on the main components of \automl{} frameworks before describing several prominent instantiations in more depth.

\subsubsection{Components of \automl{} systems}
\label{ssec:components}

\automl{} systems require a flexible pipeline \configspace{} and are driven by an efficient method to search this space. Furthermore, they rely on model selection and budget allocation strategies when evaluating different pipelines. Additionally, to speed up the search procedure, information gained on other datasets can be used to kick-start or guide the search procedure (i.e. meta-learning). Finally, one can also combine the models trained during the search phase in a post-hoc ensembling step.

\paragraph{Configuration Space and Search Mechanism}

While there are \configspace{} formulations that allow the application of multiple search mechanisms, not all formulations of a configuration space and a search mechanism can be mixed and matched, and we, therefore, describe the different formulations and the applicable search mechanisms in turn.

The most common description of the search space is the CASH formulation. There is a fixed amount of hyperparameters, each with a range of legal values or categorical choices, and some of them can be conditional, meaning that they are only active if other hyperparameters fulfill certain conditions. One such example is the choice of a classification algorithm and its hyperparameters. The hyperparameters of an SVM are only active if the categorical hyperparameter of the classification algorithm is set to SVM.

Standard black-box optimization algorithms can solve the CASH problem, and SMAC \citep{hutter-lion11a} and TPE~\citep{bergstra-nips11a} were proposed first for this task. Others proposed the use of evolutionary algorithms~\citep{burger-icpram2015correct} and random search~\citep{ledell-automl20a}. It is also known as the full model selection problem~\citep{escalante-jmlr09a}, and solutions in that strain of work proposed the use of particle swarm optimization~\citep{escalante-jmlr09a} and a combination of a genetic algorithm with particle swarm optimization~\citep{sun-mcs13a}. To improve performance one can prune the configuration space to reduce the size of the space the optimization algorithm has to search through~\citep{zhang-kdd16a}, split the \configspace{} into smaller, more manageable subspaces~\citep{alaa-icml18a,liu-aaai2020a}, or heavily employ expert knowledge~\citep{ledell-automl20a}.

Instead of a fixed configuration space, genetic programming can make use of a flexible and possibly infinite space of components to be connected~\citep{olson-evo16a,olson-gecco16a}. This approach can be formalized further by using grammar-based genetic programming~\citep{deSa-eurogp2017a}. Context-free grammars can also be searched by model-based reinforcement learning algorithms~\citep{drori-automl2019a}.

Formalizing the search problem as a search tree allows the application of a custom Monte-Carlo tree search~\citep{rakotoarison-ijcai2019a} and hierarchical task networks with best-first search~\citep{mohr-ml18a}. With discrete spaces it is also possible to use combinations of meta-learning and matrix factorization~\citep{yang-kdd2019a,yang-kdd2020a,fusi-neurips2018a}. In the special case of using only neural networks in an \automl{} system it is possible to stick with standard black-box optimization~\citep{mendoza-automl16a,mendoza-automlbook19a,zimmer-tpami21a}, but one can also employ recent advances in neural architecture search~\citep{elsken-automlbook2019a}.

\paragraph{Meta-Learning.}
When there is knowledge about previous runs of the \automl{} system on other datasets available, it is possible to employ meta-learning. One option is to define a dataset similarity measure, often by using hand-crafted meta-features which describe the datasets~\citep{brazdil-ecml1994a}, to use the best solutions on the closest seen datasets to warmstart the search  algorithm~\citep{feurer-nips15a}. While this way of meta-learning can be seen as an add-on to existing methods, other works use search strategies designed to take meta-learning into account, for example matrix factorization~\citep{yang-kdd2019a,yang-kdd2020a,fusi-neurips2018a} or reinforcement learning~\citep{drori-automl2019a,heffetz-kdd2020a}.

\paragraph{Model Selection.}
Given training data, the goal of an \automl{} system is to find the best performing ML pipeline. Doing so requires to best approximate the generalization error to 1) provide a reliable and precise signal for the optimization procedure\footnote{Different model selection strategies could be ignored from an optimization point of view, where the goal is to optimize performance given a loss function, as is often done in the research fields of meta-learning and hyperparameter optimization. However, for \automl{} systems, this is highly relevant as we are not interested in the optimization performance (of some subpart) of these systems, but the final estimated generalization performance when applied to new data.} and 2) select the model to be returned in the end. Typically, the generalization error is assessed via the \emph{train-validation-test} protocol~\citep{bishop-book95a,raschka-arxiv2018a}. This means that several models are trained on a \emph{training set}, the best one is selected via holdout (using a single split) or the K-fold cross-validation, 
and the generalization error is then reported on the test set. 
The \automl{} system then returns a single model in case of holdout and a combination of $K$ models in case of $K$-fold cross-validation~\citep{caruana-icdm06a}.
One could also use model selection strategies aiming to reduce the effect of overfitting to the validation set~\citep{dwork2015reusable,tsamardinos-ml2018a}, but while such model selection strategies are an important area of research, houldout or K-fold cross-validation remain the most prominent choices~\citep{henery-book1994a,kohavi-ml95a,hastie-esl01a,guyon-jmlr10a,bischl-evco12a,raschka-arxiv2018a}.

The influence of the model selection strategy on the performance is well known~\citep{kalousis-icml03a}, and researchers have studied their impact~\citep{kohavi-ijcai95a}. 
However, there is no single best strategy, since there is a tradeoff between approximation quality and time required to compute the validation loss.

\paragraph{Post-hoc Ensembling.} 

\automl{} systems evaluate dozens or hundreds of models during their optimization procedure. Thus, it is a natural next step to not only use a single model at the end but to ensemble multiple for improved performance and reduced overfitting.

This was first proposed to combine the solutions found by particle swarm optimization~\citep{escalante-ieee10a} and then by an evolutionary algorithm~\citep{burger-icpram2015correct}. While these works used heuristic methods to combine multiple models into a final ensemble, it is also possible to treat this as another optimization problem~\citep{feurer-nips15a} and solve it with ensemble selection~\citep{caruana-icml04a} or stacking~\citep{ledell-automl20a}.

Instead of using a single layer of machine learning models, Automatic Frankensteining~\citep{wistuba-siam17a} proposed two-layer stacking, applying \automl{} to the outputs of an \automl{} system instead of a single layer of ML algorithms followed by an ensembling mechanism. Auto-Stacker went one step further, directly optimizing for a two-layer \automl{} system~\citep{chen-gecco18a}.

\subsubsection{\automl{} systems}

To the best of our knowledge, the first \automl{} system which tunes both hyperparameters and chooses algorithms was an ensemble method~\citep{caruana-icml04a}. This system randomly produces $2\ 000$ classifiers from a wide range of ML algorithms and constructs a post-hoc ensemble. It was later robustified~\citep{caruana-icdm06a} and employed in a winning submission to the KDD challenge~\citep{niculescu-kddcup2009}.

The first \automl{} system to jointly optimize the whole pipeline was \emph{Particle Swarm Model Selection}~\citep{escalante-ijcnn07a,escalante-jmlr09a}.
It used a fixed-length representation of the pipeline and contained feature selection, feature processing, classification and post-processing implemented in the CLOP package\footnote{\url{http://clopinet.com/CLOP/}} and was developed for the IJCNN 2007 agnostic learning vs. prior knowledge challenge~\citep{guyon-ijcnn07a}. It placed 2nd among the solutions using the CLOP package provided by the organizers, only losing to a submission based on robust hyperparameter optimization and ensembling~\citep{reunanen-ijcnn07a}.
Later systems started employing model-based global optimization algorithms, such as \autoweka{}~\citep{thornton-kdd13a,kotthoff-automlbook2019a}, which is built around the WEKA software~\citep{hall-sigkdd09a} and SMAC~\citep{hutter-lion11a} and uses cross-validation with racing for model evaluation, and Hyperopt-sklearn~\citep{komer-automl14a}, which was the first tool to use the now-popular scikit-learn~\citep{scikit-learn} and paired it with the TPE algorithm from the hyperopt package~\citep{bergstra-nips11a,bergstra-icml13a} and holdout.

We extended the approach of parametrizing a popular machine learning library and optimizing its hyperparameters with a black-box optimization algorithm using meta-learning and post-hoc ensembles in \autosklearn{}~\citep{feurer-nips15a,feurer-automlbook2019a}. For classification, the space of possible ML pipelines currently spans $16$ classifiers, $14$ feature preprocessing methods and numerous data preprocessing methods, adding up to $122$ hyperparameters for the latest release. \autosklearn{} uses holdout as a default model selection strategy but allows for other strategies such as cross-validation. \autosklearn{} was the dominating solution of the first \automl{} challenge~\citep{guyon-automl19a}.

The \emph{tree-based pipeline optimization tool} (\tpot{}; \citealp{olson-evo16a,olson-automl2019a}) uses grammatical evolution 
to construct ML pipelines of arbitrary length.
Currently, it uses scikit-leearn~\citep{scikit-learn} and XGBoost~\citep{chen-kdd16a} for its ML building blocks and 5-fold cross-validation to evaluate individual solutions. TPOT-SH~\citep{parmentier-ictai2019a}, inspired by our submission to the second \automl{} challenge, uses successive halving to speed up \tpot{} on large datasets.

\updates{There are also multiple \automl{} systems that exploit stacking~\citep{wolpert-nn92a}. First, Automatic Frankensteining~\citep{wistuba-siam17a} introduces a two-stage optimization process to build a two-layer stacking model. Second, AutoStacker directly optimizes a two-layer stacking model with a genetic algorithm~\citep{chen-gecco18a}. Third, \emph{H2O \automl{}} package builds on a manually designed set of defaults and random search and combines them in a post-hoc stacking step, using building blocks from the H2O library~\citep{h2o_platform} and XGBoost~\citep{chen-kdd16a}, and employing cross-validation. Lastly, AutoGluon takes a radically different approach and completely drops hyperparameter optimization and invests all available time into building a robust stacking model~\citep{erickson-arxiv20a}.}

\updates{
Recently, there also have been works that aim to use dataset subsets to speed up the evaluation~\citep{parmentier-ictai2019a,wang-mlsys21a}.
}

Finally, there is also work on creating \automl{} systems that can leverage recent advancements in deep learning, using either black-box optimization~\citep{mendoza-automl16a,zimmer-tpami21a} or neural architecture search~\citep{jin-sigkdd19a}.

Of course, there are also many techniques related to \automl{} which are not used in one of the \automl{} systems discussed in this section, and we refer to \cite{hutter-book19a} for an overview of the field of Automated Machine Learning, to \cite{brazdil-08a} for an overview on meta-learning research which pre-dates the work on \automl{} and to \cite{escalante-ncs2021a} for a discussion on the history of \automl{}.

\section{Discussion and Conclusion}
\label{sec:discussion} 

In this paper, we introduced our winning entry to the 2nd ChaLearn \automl{} challenge, \poshautosklearn{}, and automated its internal settings further, resulting in the next generation of our \automl{} system: \askltwo{}. It provides a truly hands-free solution, which, given a new task and resource limitations, automatically chooses the best setup. Specifically, we introduce three improvements for faster and more efficient \automl{}: (i) to get strong results quickly, we propose to use portfolios, which can be built offline and thus reduce startup costs, (ii) to reduce time spent on poorly performing pipelines we propose to add successive halving as a budget allocation strategy to the \configspace{} of our \automl{} system and (iii) to close the design space we opened up for \automl{} we propose to automatically select the best configuration of our system.

We conducted a large-scale study based on $\ndatasets$ meta-datasets for constructing our \automl{} systems and $\nautodata$ datasets for evaluating them and obtained substantially improved performance compared to \asklone{}, reducing the ADTM by up to a factor of $4.5$ and achieving a lower loss after $10$ minutes than \asklone{} after $60$ minutes.
Our ablation study showed that using a \modelselector{} to choose the model selection strategy has the largest impact on performance and allows \askltwo{} to run robustly on new, unseen datasets. Furthermore, we showed that our method is highly competitive and outperforms other state-of-the-art \automl{} systems in the OpenML AutoML benchmark.

However, our system also introduces some shortcomings since it optimizes performance towards a given \horizon{}, performance metric and \configspace{}. Although all of these, along with the meta datasets, could be provided by a user to automatically build a customized version of \askltwo{}, it would be interesting whether we can learn how to transfer a specific \automl{} system to different \horizon{s} and metrics. 
\updates{Although we have observed strong empirical performance using SH, we do not have any performance guarantee when we combine SH with BO. Therefore, we deem developing approaches that increase successive halving's lower budget over time promising next steps.}
Also, there remain several hand-picked hyperparameters on the level of the \automl{} system, which we plan to automate in future work. These are, for example, automatically learning the portfolio size, learning more hyper-hyperparameters of the different budget allocation strategies (for instance, of SH) and proposing suitable \configspace{}s given a dataset and resources.
\updates{Besides these, our use of two meta-features for the selector opens up the research question of whether other meta-features could result in better performance. We expect that we can tackle many of these problems by performing an additional optimization loop on the training data.}
Finally, building the training data is currently quite expensive. Even though this has to be done only once, it will be interesting to see whether we can take shortcuts here, for example, by using a joint ranking model~\citep{tornede-ds20a} or non-linear collaborative filtering~\citep{fusi-neurips2018a}. 
\acks{The authors acknowledge support by the state of Baden-W\"{u}rttemberg through bwHPC and the German Research Foundation (DFG) through grant no INST 39/963-1 FUGG. This work has partly been supported by the European Research Council (ERC) under the European Union’s Horizon 2020 research and innovation programme under grant no.\ 716721. Robert Bosch GmbH is acknowledged for financial support. We furthermore thank all contributors to \autosklearn{} for their help in making it a useful \automl{} tool and also thank Francisco Rivera for providing a Singularity integration for the \automl{} benchmark.}

\newpage

\appendix

\section{Additional pseudo-code}
\label{app:pseudo-code} 

We give pseudo-code for computing the estimated generalization error of $\Portfolio$ across all meta-datasets $\MetaSet$ for $\nfolds$-folds cross-validation in Algorithm~\ref{alg:cv} and successive halving in Algorithm~\ref{alg:sh}.

\begin{algorithm}
   \caption{Estimating the generalization error of a portfolio with K-Fold Cross-Validation}
   \label{alg:cv}
\begin{algorithmic}[1]
   \STATE {\bfseries Input:} Ordered set of ML pipelines $\Portfolio$, datasets $\MetaSet$, number of folds $\nfolds$, 
   \STATE $L = 0$
   \FOR{$\dataiter \in (1, 2, \dots, |\MetaSet|)$} 
   \STATE $l_\dataiter = \infty$
   
   \FOR{$p \in \Portfolio$}
   \STATE $l = 0$

   \FOR{$\folditer \in (1, 2, \dots, \nfolds)$}
   \STATE $l = l + \reallywidehat{\gerror}(\Model_\conf^{\TrainingSet^{(\text{train},\folditer)}},{\TrainingSet^{(\text{val},\folditer)}})$ 
   \ENDFOR
   
   \STATE $l = l / \nfolds$

   \IF{$l < l_\dataiter$}
   \STATE $l_\dataiter = l$
   \ENDIF
   \ENDFOR
   \STATE $L = L + l_\dataiter$
   \ENDFOR
   \RETURN $L / |\MetaSet|$
\end{algorithmic}
\end{algorithm}

\begin{algorithm}
   \caption{Estimating the generalization error of a portfolio with Successive Halving}
   \label{alg:sh}
\begin{algorithmic}[1]
   \STATE {\bfseries Input:} Ordered set of ML pipelines $\Portfolio$, datasets $\MetaSet$, minimal budget $b_{min}$, maximal budget $b_{max}$, downsampling rate $\eta$
   \STATE $L = \infty$
   \STATE $R = b_{max} / b_{min}$
   \STATE $s_{max} = \lfloor \log_\eta(R) \rfloor$
   \STATE $B = (s_{max} + 1)R$
   \STATE $n = \lceil \frac{B}{R} \frac{\eta^{s_{max}}}{(s_{max} + 1)} \rceil$
   \STATE $r = R\eta^{-s_{max}}$
   
   \FOR{$\dataiter \in (1, 2, \dots, |\MetaSet|)$} 
   \STATE $l_\dataiter = \infty$
   \STATE $\Portfolio_\dataiter = \Portfolio$
   \WHILE{True}
   
   \STATE $\Portfolio' = \Portfolio.pop(r)$ \# Pop top $r$ machine learning pipelines
   \STATE $\mathbf{l} = [\text{ }]$
   
   \FOR{$i \in (0,\dots,s_{max})$}
   
   \STATE $n_i = \lfloor n \eta^{-i} \rfloor$
   \STATE $r_i = r \eta^{i}$
   
   \FOR{$p \in \Portfolio'$}
   
   \STATE $l = \reallywidehat{\gerror}(\Model_\conf^{\TrainingSet^{\text{train}}},\TrainingSet^{\text{val}})$
   \STATE $\mathbf{l} = \mathbf{l} \cup l$
   \IF{$l < l_\dataiter$}
   \STATE $l_\dataiter = l$
   \ENDIF
   
   \ENDFOR
   
   \STATE$\Portfolio' = top(\Portfolio', \mathbf{l}, \lfloor(n_i / eta)\rfloor)$, where $top(\Portfolio, \mathbf{l}, k)$ returns the top $k$ performing machine learning pipelines.
   
   \ENDFOR
   
   \IF{$|\Portfolio_\dataiter| == 0$}
      \STATE break
   \ENDIF
   
   \ENDWHILE
   \STATE $L = L + l_\dataiter$
   \ENDFOR
   \RETURN $L / |\MetaSet|$
\end{algorithmic}
\end{algorithm}

\section{Additional results and experiments}
\label{app:practical-considerations} 

In this section we will give additional results backing up our findings. Concretely, we will give further details on the reduced search space and provide further experimental evidence, we will provide the main results from the main paper without post-hoc ensembles, and we will give the raw numbers before averaging.

\subsection{Early Stopping and Retrieving Intermittent Results}
Estimating the generalization error of a pipeline $\Model_\conf$ practically requires to restrict the CPU-time per evaluation to prevent that one single, very long algorithm run stalls the optimization procedure~\citep{thornton-kdd13a,feurer-nips15a}. If an algorithm does not return a result within the assigned time limit, it is terminated and the worst possible generalization error is assigned. If the time limit is set too low, a majority of the algorithms do not return a result and thus provide very scarce information for the optimization procedure. A too high time limit, however, might as well not return any meaningful results since all time may be spent on long-running, under-performing pipelines.
Additionally, for iterative algorithms (e.g., gradient boosting and linear models trained with stochastic gradient descent), it is important to set the number of iterations such that the training converges and does not overfit, but most importantly finishes within this timelimit. Setting this number too high (training exceeds time limit and/or overfit) or too low (training has not yet converged although there is time left) has detrimental effects to the final performance of the \automl{} system. 
To mitigate this risk we implemented two measures for iterative algorithms. Firstly, we use the early stopping mechanisms implemented by scikit-learn. Specifically, training stops if the loss on the training or validation set (depending on the model and the configuration) increases or stalls, which prevents overfitting (i.e. early stopping). 
Secondly, we make use of intermittent results 
retrieval, e.g., saving the results at checkpoints spaced at geometrically increasing iteration numbers, thereby ensuring that every evaluation of an iterative algorithm returns a performance and thus yields information for the optimizer. With this, our \automl{} tool can robustly tackle large datasets without the necessity to finetune the number of iterations dependent on the time limit.

\begin{table}[t]
    \centering
    \begin{tabular}{ll|rrrr}
\toprule
{} & {} &      10 & STD 10 &     60 & STD 60 \\
\midrule
(1) & Auto-sklearn (1.0)                  & $16.21$ & $0.27$ & $\underline{7.17}$ & $0.30$ \\
(2) & Auto-sklearn (1.0) ISS              & $18.10$ & $0.13$ & $9.57$ & $0.22$ \\
(3) & Auto-sklearn (1.0) ISS + IRR        &  $5.29$ & $0.13$ & $3.98$ & $0.21$ \\
(4) & Auto-sklearn (1.0) ISS + IRR + Port &  $\mathbf{3.70}$ & $0.14$ & $\mathbf{3.08}$ & $0.13$ \\
\bottomrule
\end{tabular}
    \caption{Comparison of \asklone{} (1) with using only the iterative search space (2), using the iterative search space and iterative results retrieval (3) and also using a portfolio (4).}
    \label{tab:app:irr}
\end{table}

To study the effect of using the iterative results retrieval we compare standard \asklone{} with \autosklearn{} with the following changes applied one after the other: 1) move to a \configspace{} which consists only of iterative algorithms 2) enable intermittent results retrieval and 3) replace the KND by the portfolio. We give results in Table~\ref{tab:app:irr} and note that the KND uses meta-data gathered specifically for use with the reduced \configspace{}. Only restricting the \configspace{} leads to decreased performance which we attribute to the reduced hypothesis space. Intermittently writing results to disk reduces the amount of failures, and using a portfolio instead of the KND results in the best overall performance.

\begin{figure}
    \centering
    \includegraphics[width=\textwidth]{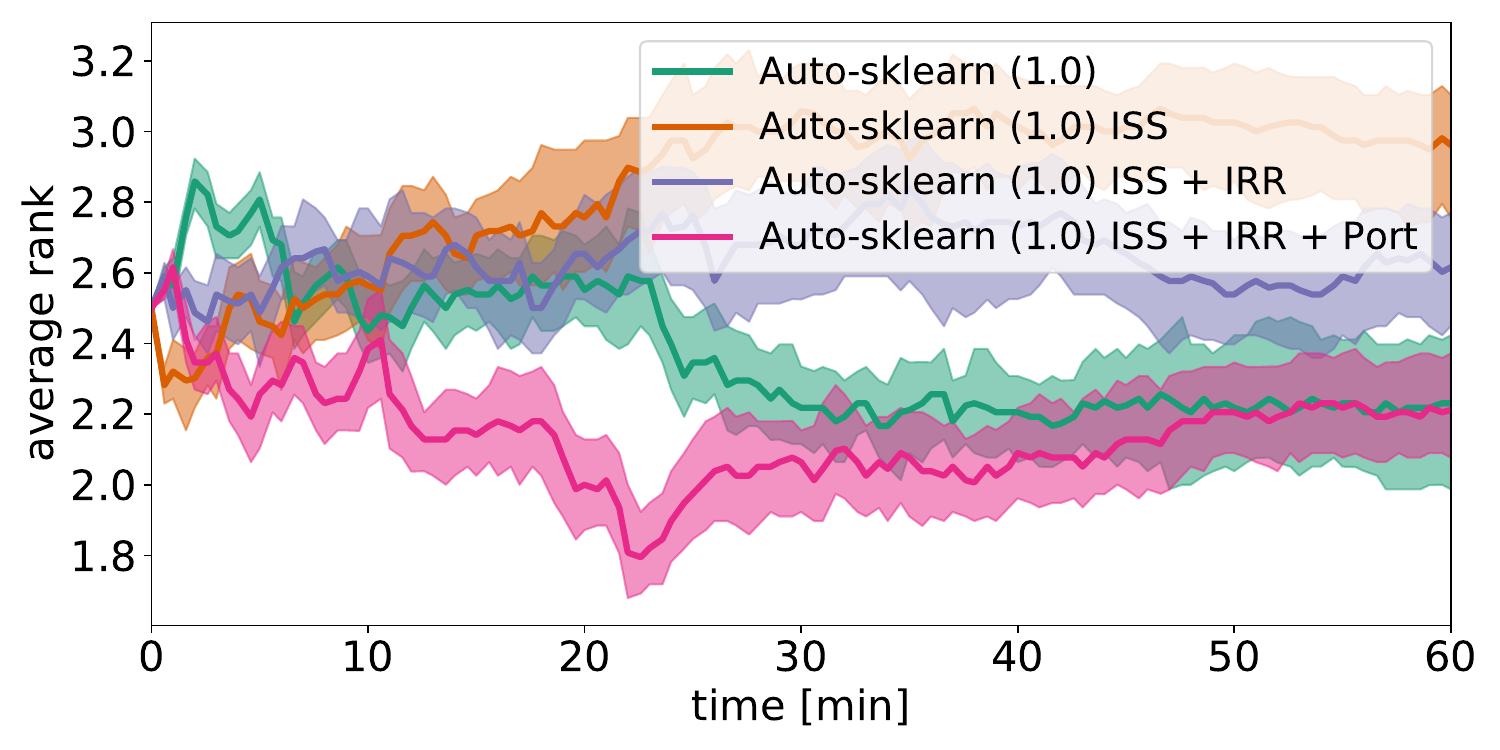}
    \caption{Ranking plot (lower is better) comparing \asklone{} (1) with using only the iterative search space (2), using the iterative search space and iterative results retrieval (3) and also using a portfolio (4). \updates{Compared to Figure~\ref{fig:RQ1:plots} we randomly sample $500$ combinations of the $10000$ combinations of the $10$ seeds of the $4$ \automl{} systems.}
    }
    \label{fig:app:irr}
\end{figure}

Once again, we also view the results through the eyes of a ranking plot in Figure~\ref{fig:app:irr}. These results demonstrate that the iterative search space combined with intermittent results retrieval and a portfolio is especially dominating in the short term, and it takes a total of 50 minutes for \asklone{} to catch up. We would like to note that the performance of \askltwo{} is even better as can be seen in Table~\ref{tab:RQ1}, but it would be interesting to see how a portfolio of the full \configspace{} would perform, which we note as a further research question.

\subsection{Performance Without Post-Hoc Ensembling}
\label{app:performance_wo_posthoc_ensembles}

We first give numbers comparing only Bayesian optimization, k-nearest datasets (KND) and a greedy portfolio. These results are similar to Table~\ref{tab:portfolio:results}, but do not show the results of post-hoc ensembling, but using the single best model. Overall, they are qualitatively very similar, but it can be observed that the ensemble improves the average normalized balanced error rate in every case.

\begin{table}[]
    \centering
    \begin{tabular}{l|rrr|rrr}
\toprule
{} & \multicolumn{3}{c}{10 minutes} & \multicolumn{3}{c}{60 minutes} \\
{}          &    BO & KND & Port &  BO & KND & Port \\
\midrule
holdout     &  $7.27$ &  $6.43$ &       $\mathbf{4.76}$ &  $4.58$ &  $4.99$ &       $\mathbf{4.02}$ \\
SH; holdout &  $6.61$ &  $\underline{6.70}$ &       $\mathbf{5.76}$ &  $4.70$ &  $4.63$ &       $\mathbf{3.97}$ \\
3CV         &  $9.58$ &  $\underline{8.95}$ &       $\mathbf{7.88}$ &  $7.10$ &  $\underline{7.12}$ &       $\mathbf{5.98}$ \\
SH; 3CV     &  $8.88$ &  $8.97$ &       $\mathbf{7.20}$ &  $6.81$ &  $\underline{6.47}$ &       $\mathbf{6.01}$ \\
5CV         & $\mathbf{10.48}$ & $\underline{15.24}$ &      $\underline{13.77}$ &  $7.34$ &  $7.47$ &       $\mathbf{5.66}$ \\
SH; 5CV     & $11.70$ & $13.29$ &       $\mathbf{8.06}$ &  $7.05$ &  $6.69$ &       $\mathbf{5.93}$ \\
10CV        & $23.20$ & $\underline{27.45}$ &      $\mathbf{18.73}$ & $17.59$ & $\underline{17.47}$ &      $\mathbf{16.17}$ \\
SH; 10CV    & $23.98$ & $27.70$ &      $\mathbf{18.84}$ & $\underline{16.94}$ & $\underline{16.98}$ &      $\mathbf{16.07}$ \\
\bottomrule
\end{tabular}
    \caption{Results from Table~\ref{tab:portfolio:results} without post-hoc ensembles.}
    \label{tab:portfolio:results_wo_ensemble}
\end{table}

Next, we compare \askltwo{} with \poshautosklearn{} and \asklone{}, but again only show the performance of the single best model and not of an ensemble as in the main paper. Again, the ensemble result in uniform performance improvements with \askltwo{} still leading in terms of performance.

\begin{table}[]
    \centering
    \begin{tabular}{lrrrr}
\toprule
{} & \multicolumn{2}{c}{10MIN} & \multicolumn{2}{c}{60MIN} \\
{} & $\varnothing$ & std & $\varnothing$ & std \\
\midrule
Auto-sklearn (2.0) &  $\mathbf{5.01}$ & $0.18$ & $\mathbf{3.18}$ & $0.31$ \\
PoSH-Auto-sklearn  &  $5.76$ & $0.12$ & $3.97$ & $0.22$ \\
Auto-sklearn (1.0) & $23.24$ & $0.29$ & $8.68$ & $0.21$ \\
\bottomrule
\end{tabular}
    \caption{Results from Table~\ref{tab:RQ1} without post-hoc ensembles.}
    \label{tab:my_label}
\end{table}

\subsection{Unaggregated results}
\label{app:raw_results}

To allow the readers to asses the performance of the individual methods on the individual datasets we present the balanced error rates before normalizing and averaging them. We give the raw results for portfolios from Table~\ref{tab:portfolio:results} in Tables~\ref{tab:raw_portfolios_10m} and \ref{tab:raw_portfolios_60m}. Additionally, we give the raw results for \askltwo{}, \poshautosklearn{} and \asklone{} in Tables~\ref{tab:raw_askl2_10m} and \ref{tab:raw_askl2_60m}.

\begin{table}[]
    \centering
    \scriptsize
\begin{tabular}{rl|rrrrrrrr}
\toprule
Task ID &                                    Name &           holdout &       SH; holdout &               3CV &           SH; 3CV &               5CV &           SH; 5CV &              10CV &          SH; 10CV \\
\midrule
167104 &                              Australian &          $0.1721$ &          $0.1569$ &          $0.1622$ &          $0.1617$ &          $0.1583$ &          $0.1602$ & $\mathbf{0.1556}$ &          $0.1559$ \\
167184 &        blood-transfusion &          $0.3641$ & $\mathbf{0.3610}$ &          $0.3725$ &          $0.3666$ &          $0.3689$ &          $0.3722$ &          $0.3674$ &          $0.3689$ \\
167168 &                                 vehicle &          $0.2211$ &          $0.2267$ &          $0.2017$ &          $0.2093$ &          $0.2172$ &          $0.2052$ &          $0.2310$ & $\mathbf{0.1870}$ \\
167161 &                                credit-g &          $0.2942$ & $\mathbf{0.2841}$ &          $0.2939$ &          $0.2955$ &          $0.2942$ &          $0.2911$ &          $0.2939$ &          $0.2934$ \\
167185 &                                  cnae-9 &          $0.0658$ &          $0.0680$ &          $0.0651$ &          $0.0616$ & $\mathbf{0.0550}$ &          $0.0629$ &          $0.0626$ &          $0.0553$ \\
189905 &                                     car &          $0.0049$ &          $0.0049$ &          $0.0097$ &          $0.0029$ &          $0.0047$ &          $0.0017$ &          $0.0023$ & $\mathbf{0.0009}$ \\
167152 &                           mfeat-factors &          $0.0152$ &          $0.0164$ &          $0.0141$ & $\mathbf{0.0107}$ &          $0.0150$ &          $0.0117$ &          $0.0153$ &          $0.0149$ \\
167181 &                                     kc1 &          $0.2735$ &          $0.2688$ &          $0.2720$ &          $0.2713$ &          $0.2547$ &          $0.2660$ & $\mathbf{0.2477}$ &          $0.2719$ \\
189906 &                                 segment &          $0.0666$ &          $0.0687$ &          $0.0681$ & $\mathbf{0.0620}$ &          $0.0664$ &          $0.0621$ &          $0.0643$ &          $0.0671$ \\
189862 &                                 jasmine &          $0.2044$ &          $0.2051$ & $\mathbf{0.1982}$ &          $0.1986$ &          $0.2010$ &          $0.2027$ &          $0.2043$ &          $0.2027$ \\
167149 &                                kr-vs-kp & $\mathbf{0.0067}$ &          $0.0077$ &          $0.0093$ &          $0.0085$ &          $0.0079$ &          $0.0078$ &          $0.0071$ &          $0.0080$ \\
189865 &                                 sylvine &          $0.0592$ &          $0.0594$ &          $0.0600$ &          $0.0608$ &          $0.0582$ &          $0.0582$ & $\mathbf{0.0560}$ &          $0.0578$ \\
167190 &                                 phoneme &          $0.1231$ &          $0.1245$ &          $0.1168$ &          $0.1160$ &          $0.1152$ &          $0.1136$ & $\mathbf{0.1129}$ &          $0.1144$ \\
189861 &                               christine &          $0.2670$ &          $0.2621$ &          $0.2608$ &          $0.2556$ & $\mathbf{0.2517}$ &          $0.2567$ &          $0.2587$ &          $0.2645$ \\
189872 &                                  fabert &          $0.3387$ &          $0.3399$ &          $0.3140$ &          $0.3120$ & $\mathbf{0.3096}$ &          $0.3204$ &          $0.3180$ &          $0.3172$ \\
189871 &                                 dilbert &          $0.0241$ &          $0.0248$ &          $0.0258$ &          $0.0220$ & $\mathbf{0.0191}$ &          $0.0211$ &          $0.0303$ &          $0.0647$ \\
168794 &                                  robert & $\mathbf{0.5489}$ &          $0.5861$ &          $0.5762$ &          $0.5583$ &          $0.5854$ &          $0.5873$ &          $0.6230$ &          $0.6230$ \\
168797 &                                riccardo &          $0.0035$ &          $0.0052$ &          $0.0067$ &          $0.0054$ & $\mathbf{0.0027}$ & $\mathbf{0.0027}$ &          $0.5000$ &          $0.5000$ \\
168796 &                               guillermo &          $0.2186$ & $\mathbf{0.2102}$ &          $0.2311$ &          $0.2228$ &          $0.2165$ &          $0.2837$ &          $0.5000$ &          $0.5000$ \\
75097  &                  Amazon & $\mathbf{0.2361}$ &          $0.2431$ &          $0.2526$ &          $0.2526$ &          $0.2379$ &          $0.2385$ &          $0.2448$ &          $0.2443$ \\
126026 &                                   nomao &          $0.0353$ &          $0.0381$ &          $0.0360$ &          $0.0345$ & $\mathbf{0.0312}$ &          $0.0331$ &          $0.0403$ &          $0.0401$ \\
189909 &  jungle\_chess &          $0.1212$ &          $0.1251$ &          $0.1232$ &          $0.1156$ &          $0.1280$ &          $0.1180$ & $\mathbf{0.1134}$ &          $0.1141$ \\
126029 &                          bank-marketing &          $0.1397$ &          $0.1436$ &          $0.1402$ &          $0.1407$ & $\mathbf{0.1352}$ &          $0.1435$ &          $0.1378$ &          $0.1362$ \\
126025 &                                   adult &          $0.1579$ &          $0.1575$ &          $0.1553$ & $\mathbf{0.1540}$ &          $0.1591$ &          $0.1562$ &          $0.1545$ &          $0.1585$ \\
75105  &                      KDDCup09 &          $0.2450$ &          $0.2495$ & $\mathbf{0.2449}$ &          $0.2512$ &          $0.2525$ &          $0.2487$ &          $0.2497$ &          $0.2456$ \\
168795 &                                 shuttle &          $0.0093$ &          $0.0086$ &          $0.0085$ & $\mathbf{0.0084}$ &          $0.0085$ &          $0.0087$ &          $0.0088$ &          $0.0084$ \\
168793 &                                 volkert &          $0.3735$ &          $0.3724$ &          $0.3939$ & $\mathbf{0.3703}$ &          $0.3720$ &          $0.3775$ &          $0.3867$ &          $0.3957$ \\
189874 &                                  helena &          $0.7483$ &          $0.7624$ & $\mathbf{0.7475}$ &          $0.7478$ &          $0.7483$ &          $0.7534$ &          $0.7476$ &          $0.7575$ \\
167201 &                               connect-4 &          $0.2629$ &          $0.2721$ &          $0.2653$ &          $0.2630$ & $\mathbf{0.2537}$ &          $0.2565$ &          $0.3003$ &          $0.2771$ \\
189908 &                           Fashion-MNIST & $\mathbf{0.1050}$ &          $0.1098$ &          $0.1217$ &          $0.1195$ &          $0.1181$ &          $0.1153$ &          $0.1437$ &          $0.1437$ \\
189860 &                              APSFailure &          $0.0384$ &          $0.0402$ &          $0.0355$ &          $0.0364$ &          $0.0355$ & $\mathbf{0.0354}$ &          $0.0410$ &          $0.0455$ \\
168792 &                                  jannis &          $0.3654$ &          $0.3685$ &          $0.3648$ &          $0.3655$ &          $0.3651$ & $\mathbf{0.3567}$ &          $0.3912$ &          $0.3881$ \\
167083 &                             numerai28.6 &          $0.4776$ &          $0.4765$ &          $0.4752$ &          $0.4749$ & $\mathbf{0.4747}$ &          $0.4775$ &          $0.4789$ &          $0.4788$ \\
167200 &                                   higgs &          $0.2736$ &          $0.2764$ &          $0.2730$ &          $0.2742$ & $\mathbf{0.2724}$ &          $0.2744$ &          $0.2832$ &          $0.2844$ \\
168798 &                               MiniBooNE & $\mathbf{0.0581}$ &          $0.0589$ &          $0.0691$ &          $0.0633$ &          $0.0585$ &          $0.0644$ &          $0.0691$ &          $0.0685$ \\
189873 &                                  dionis & $\mathbf{0.1172}$ &          $0.1205$ &          $1.0000$ &          $1.0000$ &          $1.0000$ &          $1.0000$ &          $1.0000$ &          $1.0000$ \\
189866 &                                  albert & $\mathbf{0.3135}$ &          $0.3171$ &          $0.3469$ &          $0.3277$ &          $0.5000$ &          $0.3354$ &          $0.3714$ &          $0.3703$ \\
75127  &                                airlines &          $0.3423$ &          $0.3424$ &          $0.3450$ & $\mathbf{0.3384}$ &          $0.3419$ &          $0.3429$ &          $0.3408$ &          $0.3456$ \\
75193  &                               covertype &          $0.0568$ &          $0.0564$ &          $0.0683$ &          $0.0600$ & $\mathbf{0.0548}$ &          $0.0556$ &          $0.2519$ &          $0.2527$ \\
\bottomrule
\end{tabular}
    \caption{Results from Table~\ref{tab:portfolio:results} for 10 minutes using portfolios. We boldface the lowest error.}
    \label{tab:raw_portfolios_10m}
\end{table}

\begin{table}[]
    \centering
    \scriptsize
\begin{tabular}{rl|rrrrrrrr}
\toprule
Task ID &                                    Name &           holdout &       SH; holdout &               3CV &           SH; 3CV &               5CV &           SH; 5CV &              10CV &          SH; 10CV \\
\midrule
167104 &                              Australian &          $0.1742$ &          $0.1674$ &          $0.1623$ &          $0.1626$ &          $0.1598$ &          $0.1608$ &          $0.1625$ & $\mathbf{0.1557}$ \\
167184 &        blood-transfusion &          $0.3648$ & $\mathbf{0.3618}$ &          $0.3631$ &          $0.3641$ &          $0.3689$ &          $0.3692$ &          $0.3684$ &          $0.3692$ \\
167168 &                                 vehicle &          $0.2125$ &          $0.2344$ &          $0.1702$ &          $0.1944$ & $\mathbf{0.1657}$ &          $0.1960$ &          $0.1959$ &          $0.2151$ \\
167161 &                                credit-g &          $0.2922$ & $\mathbf{0.2895}$ &          $0.3035$ &          $0.2957$ &          $0.3056$ &          $0.2978$ &          $0.3008$ &          $0.2931$ \\
167185 &                                  cnae-9 &          $0.0733$ &          $0.0761$ &          $0.0560$ &          $0.0616$ &          $0.0537$ &          $0.0536$ &          $0.0675$ & $\mathbf{0.0518}$ \\
189905 &                                     car &          $0.0036$ &          $0.0013$ &          $0.0037$ &          $0.0098$ & $\mathbf{0.0007}$ &          $0.0012$ &          $0.0008$ &          $0.0010$ \\
167152 &                           mfeat-factors &          $0.0169$ &          $0.0186$ &          $0.0130$ & $\mathbf{0.0117}$ &          $0.0139$ &          $0.0132$ &          $0.0151$ &          $0.0122$ \\
167181 &                                     kc1 &          $0.2728$ &          $0.2739$ &          $0.2680$ &          $0.2724$ &          $0.2678$ &          $0.2804$ & $\mathbf{0.2546}$ &          $0.2576$ \\
189906 &                                 segment &          $0.0708$ &          $0.0692$ &          $0.0647$ &          $0.0635$ & $\mathbf{0.0588}$ &          $0.0596$ &          $0.0621$ &          $0.0613$ \\
189862 &                                 jasmine &          $0.2048$ &          $0.2049$ &          $0.1995$ &          $0.1989$ &          $0.1995$ & $\mathbf{0.1976}$ &          $0.1995$ &          $0.1980$ \\
167149 &                                kr-vs-kp &          $0.0060$ &          $0.0080$ &          $0.0081$ &          $0.0068$ &          $0.0064$ &          $0.0068$ &          $0.0055$ & $\mathbf{0.0053}$ \\
189865 &                                 sylvine &          $0.0590$ &          $0.0591$ &          $0.0584$ &          $0.0587$ &          $0.0577$ &          $0.0578$ & $\mathbf{0.0573}$ &          $0.0573$ \\
167190 &                                 phoneme &          $0.1222$ &          $0.1237$ &          $0.1152$ &          $0.1155$ &          $0.1111$ &          $0.1130$ &          $0.1117$ & $\mathbf{0.1105}$ \\
189861 &                               christine &          $0.2673$ &          $0.2666$ &          $0.2575$ &          $0.2584$ & $\mathbf{0.2532}$ &          $0.2575$ &          $0.2549$ &          $0.2588$ \\
189872 &                                  fabert &          $0.3381$ &          $0.3319$ &          $0.3099$ &          $0.3097$ &          $0.3119$ &          $0.3080$ & $\mathbf{0.3027}$ &          $0.3071$ \\
189871 &                                 dilbert &          $0.0200$ &          $0.0200$ & $\mathbf{0.0132}$ &          $0.0146$ &          $0.0185$ &          $0.0209$ &          $0.0212$ &          $0.0149$ \\
168794 &                                  robert &          $0.5273$ &          $0.5199$ &          $0.5238$ & $\mathbf{0.5183}$ &          $0.5456$ &          $0.5605$ &          $0.5652$ &          $0.5407$ \\
168797 &                                riccardo &          $0.0029$ & $\mathbf{0.0016}$ &          $0.0018$ &          $0.0019$ &          $0.0025$ &          $0.0076$ &          $0.5000$ &          $0.5000$ \\
168796 &                               guillermo & $\mathbf{0.2012}$ &          $0.2025$ &          $0.2057$ &          $0.2081$ &          $0.2100$ &          $0.2039$ &          $0.5000$ &          $0.5000$ \\
75097  &                  Amazon &          $0.2376$ &          $0.2394$ &          $0.2338$ &          $0.2381$ &          $0.2431$ &          $0.2384$ & $\mathbf{0.2312}$ &          $0.2324$ \\
126026 &                                   nomao &          $0.0352$ &          $0.0353$ &          $0.0334$ &          $0.0331$ &          $0.0320$ &          $0.0327$ & $\mathbf{0.0313}$ &          $0.0319$ \\
189909 &  jungle\_chess &          $0.1214$ &          $0.1221$ &          $0.1154$ &          $0.1172$ &          $0.1171$ &          $0.1153$ & $\mathbf{0.1108}$ &          $0.1141$ \\
126029 &                          bank-marketing &          $0.1388$ &          $0.1398$ &          $0.1380$ &          $0.1392$ &          $0.1382$ &          $0.1382$ & $\mathbf{0.1370}$ &          $0.1380$ \\
126025 &                                   adult &          $0.1546$ &          $0.1541$ &          $0.1550$ &          $0.1540$ &          $0.1550$ &          $0.1550$ &          $0.1539$ & $\mathbf{0.1538}$ \\
75105  &                      KDDCup09 &          $0.2492$ & $\mathbf{0.2461}$ &          $0.2477$ &          $0.2532$ &          $0.2466$ &          $0.2488$ &          $0.2617$ &          $0.2485$ \\
168795 &                                 shuttle &          $0.0136$ &          $0.0107$ &          $0.0125$ &          $0.0093$ &          $0.0084$ & $\mathbf{0.0063}$ &          $0.0127$ &          $0.0087$ \\
168793 &                                 volkert &          $0.3600$ &          $0.3673$ & $\mathbf{0.3449}$ &          $0.3551$ &          $0.3496$ &          $0.3487$ &          $0.3581$ &          $0.3563$ \\
189874 &                                  helena &          $0.7449$ &          $0.7494$ & $\mathbf{0.7331}$ &          $0.7369$ &          $0.7407$ &          $0.7404$ &          $0.7562$ &          $0.7452$ \\
167201 &                               connect-4 &          $0.2539$ &          $0.2556$ &          $0.2382$ &          $0.2428$ &          $0.2370$ &          $0.2373$ &          $0.2416$ & $\mathbf{0.2369}$ \\
189908 &                           Fashion-MNIST &          $0.1010$ & $\mathbf{0.0971}$ &          $0.1046$ &          $0.1066$ &          $0.1102$ &          $0.1105$ &          $0.1191$ &          $0.1075$ \\
189860 &                              APSFailure &          $0.0362$ &          $0.0374$ &          $0.0345$ &          $0.0364$ &          $0.0372$ &          $0.0347$ &          $0.0343$ & $\mathbf{0.0334}$ \\
168792 &                                  jannis &          $0.3670$ &          $0.3638$ &          $0.3589$ &          $0.3576$ &          $0.3584$ &          $0.3565$ & $\mathbf{0.3473}$ &          $0.3572$ \\
167083 &                             numerai28.6 &          $0.4765$ &          $0.4763$ &          $0.4774$ &          $0.4770$ &          $0.4750$ &          $0.4767$ &          $0.4755$ & $\mathbf{0.4743}$ \\
167200 &                                   higgs &          $0.2712$ &          $0.2734$ &          $0.2718$ & $\mathbf{0.2680}$ &          $0.2696$ &          $0.2680$ &          $0.2701$ &          $0.2683$ \\
168798 &                               MiniBooNE &          $0.0576$ &          $0.0583$ &          $0.0560$ & $\mathbf{0.0536}$ &          $0.0571$ &          $0.0565$ &          $0.0560$ &          $0.0608$ \\
189873 &                                  dionis & $\mathbf{0.0961}$ &          $0.1068$ &          $1.0000$ &          $1.0000$ &          $1.0000$ &          $1.0000$ &          $1.0000$ &          $1.0000$ \\
189866 &                                  albert &          $0.3116$ &          $0.3168$ &          $0.3170$ &          $0.3172$ & $\mathbf{0.3094}$ &          $0.3199$ &          $0.3183$ &          $0.3186$ \\
75127  &                                airlines &          $0.3403$ &          $0.3410$ & $\mathbf{0.3375}$ &          $0.3401$ &          $0.3388$ &          $0.3390$ &          $0.3398$ &          $0.3399$ \\
75193  &                               covertype &          $0.0537$ &          $0.0519$ &          $0.0496$ &          $0.0496$ & $\mathbf{0.0454}$ &          $0.0461$ &          $0.0458$ &          $0.0459$ \\
\bottomrule
\end{tabular}
    \caption{Results from Table~\ref{tab:portfolio:results} for 60 minutes using portfolios. We boldface the lowest error.}
    \label{tab:raw_portfolios_60m}
\end{table}

\begin{table}[]
    \centering
    \small
\begin{tabular}{rl|rrr}
\toprule
Task ID &                                    Name & Auto-sklearn (2.0) & PoSH-Auto-sklearn & Auto-sklearn (1.0) \\
\midrule
167104 &                              Australian &           $0.1617$ & $\mathbf{0.1569}$ &           $0.1628$ \\
167184 &        blood-transfusion &           $0.3694$ &          $0.3610$ &  $\mathbf{0.3534}$ \\
167168 &                                 vehicle &           $0.2030$ &          $0.2267$ &  $\mathbf{0.1654}$ \\
167161 &                                credit-g &           $0.2903$ & $\mathbf{0.2841}$ &           $0.2951$ \\
167185 &                                  cnae-9 &  $\mathbf{0.0635}$ &          $0.0680$ &           $0.0674$ \\
189905 &                                     car &  $\mathbf{0.0015}$ &          $0.0049$ &           $0.0057$ \\
167152 &                           mfeat-factors &  $\mathbf{0.0123}$ &          $0.0164$ &           $0.0185$ \\
167181 &                                     kc1 &           $0.2707$ &          $0.2688$ &  $\mathbf{0.2301}$ \\
189906 &                                 segment &           $0.0646$ &          $0.0687$ &  $\mathbf{0.0624}$ \\
189862 &                                 jasmine &           $0.2020$ &          $0.2051$ &  $\mathbf{0.1989}$ \\
167149 &                                kr-vs-kp &           $0.0086$ & $\mathbf{0.0077}$ &           $0.0089$ \\
189865 &                                 sylvine &           $0.0597$ &          $0.0594$ &  $\mathbf{0.0583}$ \\
167190 &                                 phoneme &  $\mathbf{0.1158}$ &          $0.1245$ &           $0.1257$ \\
189861 &                               christine &  $\mathbf{0.2562}$ &          $0.2621$ &           $0.2666$ \\
189872 &                                  fabert &  $\mathbf{0.3250}$ &          $0.3399$ &           $0.3323$ \\
189871 &                                 dilbert &           $0.0240$ &          $0.0248$ &  $\mathbf{0.0066}$ \\
168794 &                                  robert &  $\mathbf{0.5861}$ & $\mathbf{0.5861}$ &           $0.6545$ \\
168797 &                                riccardo &  $\mathbf{0.0052}$ & $\mathbf{0.0052}$ &           $0.5000$ \\
168796 &                               guillermo &  $\mathbf{0.2102}$ & $\mathbf{0.2102}$ &           $0.5000$ \\
75097  &                  Amazon &           $0.2435$ & $\mathbf{0.2431}$ &           $0.2610$ \\
126026 &                                   nomao &  $\mathbf{0.0336}$ &          $0.0381$ &           $0.0383$ \\
189909 &  jungle\_chess &  $\mathbf{0.1205}$ &          $0.1251$ &           $0.1231$ \\
126029 &                          bank-marketing &  $\mathbf{0.1402}$ &          $0.1436$ &           $0.1412$ \\
126025 &                                   adult &  $\mathbf{0.1547}$ &          $0.1575$ &           $0.1608$ \\
75105  &                      KDDCup09 &  $\mathbf{0.2460}$ &          $0.2495$ &           $0.2863$ \\
168795 &                                 shuttle &  $\mathbf{0.0084}$ &          $0.0086$ &           $0.0111$ \\
168793 &                                 volkert &  $\mathbf{0.3717}$ &          $0.3724$ &           $0.4233$ \\
189874 &                                  helena &  $\mathbf{0.7493}$ &          $0.7624$ &           $0.9157$ \\
167201 &                               connect-4 &  $\mathbf{0.2642}$ &          $0.2721$ &           $0.2809$ \\
189908 &                           Fashion-MNIST &  $\mathbf{0.1070}$ &          $0.1098$ &           $0.1383$ \\
189860 &                              APSFailure &           $0.0372$ &          $0.0402$ &  $\mathbf{0.0370}$ \\
168792 &                                  jannis &           $0.3654$ &          $0.3685$ &  $\mathbf{0.3637}$ \\
167083 &                             numerai28.6 &  $\mathbf{0.4753}$ &          $0.4765$ &           $0.4774$ \\
167200 &                                   higgs &  $\mathbf{0.2746}$ &          $0.2764$ &           $0.2777$ \\
168798 &                               MiniBooNE &           $0.0603$ & $\mathbf{0.0589}$ &           $0.0622$ \\
189873 &                                  dionis &  $\mathbf{0.1205}$ & $\mathbf{0.1205}$ &           $0.6731$ \\
189866 &                                  albert &  $\mathbf{0.3171}$ & $\mathbf{0.3171}$ &           $0.4407$ \\
75127  &                                airlines &  $\mathbf{0.3404}$ &          $0.3424$ &           $0.3536$ \\
75193  &                               covertype &  $\mathbf{0.0564}$ & $\mathbf{0.0564}$ &           $0.8571$ \\
\bottomrule
\end{tabular}

    \caption{Results from Table~\ref{tab:RQ1} for 10 minutes. We boldface the lowest error.}
    \label{tab:raw_askl2_10m}
\end{table}

\begin{table}[]
    \centering
    \small
\begin{tabular}{rl|rrr}
\toprule
Task ID &                                    Name & Auto-sklearn (2.0) & PoSH-Auto-sklearn & Auto-sklearn (1.0) \\
\midrule
167104 &                              Australian &  $\mathbf{0.1562}$ &          $0.1674$ &           $0.1658$ \\
167184 &        blood-transfusion &           $0.3669$ &          $0.3618$ &  $\mathbf{0.3572}$ \\
167168 &                                 vehicle &           $0.2187$ &          $0.2344$ &  $\mathbf{0.1822}$ \\
167161 &                                credit-g &           $0.2980$ & $\mathbf{0.2895}$ &           $0.3004$ \\
167185 &                                  cnae-9 &  $\mathbf{0.0566}$ &          $0.0761$ &           $0.0620$ \\
189905 &                                     car &           $0.0038$ & $\mathbf{0.0013}$ &           $0.0043$ \\
167152 &                           mfeat-factors &  $\mathbf{0.0126}$ &          $0.0186$ &           $0.0136$ \\
167181 &                                     kc1 &           $0.2600$ &          $0.2739$ &  $\mathbf{0.2250}$ \\
189906 &                                 segment &  $\mathbf{0.0609}$ &          $0.0692$ &           $0.0697$ \\
189862 &                                 jasmine &  $\mathbf{0.1971}$ &          $0.2049$ &           $0.1985$ \\
167149 &                                kr-vs-kp &  $\mathbf{0.0060}$ &          $0.0080$ &           $0.0085$ \\
189865 &                                 sylvine &           $0.0572$ &          $0.0591$ &  $\mathbf{0.0555}$ \\
167190 &                                 phoneme &  $\mathbf{0.1140}$ &          $0.1237$ &           $0.1235$ \\
189861 &                               christine &  $\mathbf{0.2592}$ &          $0.2666$ &           $0.2619$ \\
189872 &                                  fabert &  $\mathbf{0.3120}$ &          $0.3319$ &           $0.3185$ \\
189871 &                                 dilbert &           $0.0163$ &          $0.0200$ &  $\mathbf{0.0090}$ \\
168794 &                                  robert &  $\mathbf{0.5199}$ & $\mathbf{0.5199}$ &           $0.5327$ \\
168797 &                                riccardo &           $0.0016$ &          $0.0016$ &  $\mathbf{0.0016}$ \\
168796 &                               guillermo &           $0.2025$ &          $0.2025$ &  $\mathbf{0.1964}$ \\
75097  &                  Amazon &  $\mathbf{0.2371}$ &          $0.2394$ &           $0.2481$ \\
126026 &                                   nomao &  $\mathbf{0.0323}$ &          $0.0353$ &           $0.0361$ \\
189909 &  jungle\_chess &           $0.1145$ &          $0.1221$ &  $\mathbf{0.1136}$ \\
126029 &                          bank-marketing &  $\mathbf{0.1387}$ &          $0.1398$ &           $0.1428$ \\
126025 &                                   adult &           $0.1544$ & $\mathbf{0.1541}$ &           $0.1574$ \\
75105  &                      KDDCup09 &           $0.2504$ & $\mathbf{0.2461}$ &           $0.2549$ \\
168795 &                                 shuttle &  $\mathbf{0.0093}$ &          $0.0107$ &           $0.0109$ \\
168793 &                                 volkert &           $0.3563$ &          $0.3673$ &  $\mathbf{0.3440}$ \\
189874 &                                  helena &  $\mathbf{0.7399}$ &          $0.7494$ &           $0.7693$ \\
167201 &                               connect-4 &  $\mathbf{0.2408}$ &          $0.2556$ &           $0.2709$ \\
189908 &                           Fashion-MNIST &           $0.1023$ & $\mathbf{0.0971}$ &           $0.0984$ \\
189860 &                              APSFailure &  $\mathbf{0.0343}$ &          $0.0374$ &           $0.0375$ \\
168792 &                                  jannis &  $\mathbf{0.3591}$ &          $0.3638$ &           $0.3641$ \\
167083 &                             numerai28.6 &  $\mathbf{0.4759}$ &          $0.4763$ &           $0.4760$ \\
167200 &                                   higgs &  $\mathbf{0.2690}$ &          $0.2734$ &           $0.2738$ \\
168798 &                               MiniBooNE &  $\mathbf{0.0561}$ &          $0.0583$ &           $0.0620$ \\
189873 &                                  dionis &  $\mathbf{0.1068}$ & $\mathbf{0.1068}$ &           $0.6731$ \\
189866 &                                  albert &           $0.3168$ &          $0.3168$ &  $\mathbf{0.3143}$ \\
75127  &                                airlines &  $\mathbf{0.3394}$ &          $0.3410$ &           $0.3449$ \\
75193  &                               covertype &  $\mathbf{0.0519}$ & $\mathbf{0.0519}$ &           $0.8571$ \\
\bottomrule
\end{tabular}
    \caption{Results from Table~\ref{tab:RQ1} for 60 minutes. We boldface the lowest error.}
    \label{tab:raw_askl2_60m}
\end{table}

\section{Theoretical properties of the greedy algorithm}

\subsection{Definitions}\label{sec:definitions}

\begin{mydef}
(Discrete derivative, from \citealp{krause-trac14a}) For a set function $f : 2^\mathcal{V} \rightarrow \mathbb{R}, \mathcal{S} \subseteq \mathcal{V}$ and $e \in \mathcal{V}$ let $\Delta_f(e|\mathcal{S}) = f(\mathcal{S} \cup \{e\}) - f(\mathcal{S})$ be the \emph{discrete derivative} of $f$ at $S$ with respect to $e$.
\end{mydef}

\begin{mydef}\label{def:submodularity}
(Submodularity, from \citealp{krause-trac14a}): A function $f : 2^\mathcal{V} \rightarrow \mathbb{R}$ is \emph{submodular} if for every $\mathcal{A} \subseteq \mathcal{B} \subseteq \mathcal{V}$ and $e \in \mathcal{V} \setminus \mathcal{B}$ it holds that $\Delta_f(e|\mathcal{A}) \ge \Delta_f(e|\mathcal{B})$.
\end{mydef}

\begin{mydef}\label{def:monotone}
(Monotonicity, from \citealp{krause-trac14a}): A function $f : 2^\mathcal{V} \rightarrow \mathbb{R}$ is \emph{monotone} if for every $\mathcal{A} \subseteq \mathcal{B} \subseteq V, f(\mathcal{A}) \le f(\mathcal{B})$.
\end{mydef}

\subsection{Choosing on the test set}\label{app:proof}

In this section we give a proof of Proposition 1 from the main paper:

\begin{prop}
Minimizing the test loss of a portfolio $\Portfolio$ on a set of datasets $\Dataset_1, \dots, \Dataset_{|\MetaSet|}$, when choosing a ML pipeline from $\Portfolio$ for $\Dataset_\dataiter$ based on performance on $\Dataset_{\dataiter,\text{test}}$, is equivalent to the \emph{sensor placement problem} for minimizing detection time~\citep{krause_jwrpm2008a}.
\end{prop}

Following Krause et al.~\citep{krause_jwrpm2008a}, sensor set placement aims at maximizing a so-called \emph{penalty reduction} $R(\mathcal{A}) = \sum_{i \in \mathcal{I}} P(i) R(\mathcal{A}, i)$, where $\mathcal{I}$ are intrusion scenarios following a probability distribution $P$ with $i$ being a specific intrusion. $\mathcal{A} \subset \Candidates$ is a \emph{sensor placement}, a subset of all possible locations $\Candidates$ where sensors are actually placed. Penalty reduction $R$ is defined as the reduction of the penalty when choosing $\mathcal{A}$ compared to the maximum penalty possible on scenario $i$: $R(\mathcal{A}, i) = \penaltyfunction_i(\infty) - \penaltyfunction_i(T(\mathcal{A}, i))$. In the simplest case where action is taken upon intrusion detection, the penalty is equal to the detection time ($\penaltyfunction_i(t)=t$). The detection time of a sensor placement $T(\mathcal{A}, i)$ is simply defined as the minimum of the detection times of its individual members: $\min_{s \in \mathcal{A}} T(s,i)$.

In our setting, we need to do the following replacements to find that the problems are equivalent:
\begin{enumerate}
    \item Intrusion scenarios $\mathcal{I}$: datasets $\{\Dataset_1, \dots, \Dataset_{|\MetaSet|}\}$,
    \item Possible sensor locations $\Candidates$: set of candidate ML pipelines of our algorithm  $\Candidates$,
    Detection time $T(s \in \mathcal{A},i)$ on intrusion scenario $i$: test performance $\Lossfunction(\Model_C, \Dataset_{\dataiter,\text{test}})$ on dataset $\Dataset_\dataiter$,
    \item Detection time of a sensor placement $T(\mathcal{A},i)$: test loss of applying portfolio $\Portfolio$ on dataset $\Dataset_\dataiter$: $\min_{p \in \Portfolio} \Lossfunction(p, \Dataset_{\dataiter,\text{test}})$
    \item Penalty function $\penaltyfunction_i(t)$: loss function $\Lossfunction$, in our case, the penalty is equal to the loss.
    \item Penalty reduction for an intrusion scenario $R(\mathcal{A}, i)$: the penalty reduction for successfully applying a portfolio $\Portfolio$ to dataset $\dataiter$: $R(\Portfolio,\dataiter) = \penaltyfunction_\dataiter(\infty) - \min_{p \in \Portfolio} \Lossfunction(p, \Dataset_{\dataiter,\text{test}})$.\footnote{This would be the general case for a metric with no upper bound. In case of metrics such as the misclassification error, the maximal penalty would be $1$.}
\end{enumerate}
$\blacksquare$

\subsection{Choosing on the validation set}\label{sec:choose_on_valid}

We demonstrate that choosing an ML pipeline from the portfolio via holdout (i.e. a validation set) and reporting its test performance is neither submodular nor monotone 
by a simple example.
To simplify notation we argue in terms of performance instead of penalty reduction, which is equivalent.

Let $\mathcal{B} = \{(5, 5), (7, 7), (10, 10)\}$ and $\mathcal{A} = \{(5, 5), (7, 7)\}$, where each tuple represents the validation and test performance. For $e = (8, 6)$ we obtain the discrete derivatives $\Delta_f(e|\mathcal{A}) = -1$ and $\Delta_f(e|B) = 0$ which violates Definition \ref{def:submodularity}.
The fact that the discrete derivative is negative violates Definition \ref{def:monotone} because $f(\mathcal{A}) > f(\mathcal{A} \cup \{e\})$.

\subsection{Successive Halving}
\label{app:choosing-for-sh}

As in the previous subsection, we use a simple example to demonstrate that selecting an algorithm via the successive halving model selection strategy is neither submodular nor monotone. 
To simplify notation we argue in terms of performance instead of penalty reduction, which is equivalent.

Let $\mathcal{B} = \{((5, 5), (8, 8)), ((5, 5), (6, 6)), ((4, 4), (5, 5))\}$ and $\mathcal{A} = \{((5, 5), (7, 7))\}$, where each tuple is a learning curve of validation-, test performance tuples. For $e = ((6, 5), (6, 5))$, we eliminate entries 2 and 3 from $\mathcal{B}$ in the first iteration of successive halving (while we advance entries 1 and 4), and we eliminate entry 1 from $\mathcal{A}$. After the second stage, the performances are $f(\mathcal{B}) = 8$ and $f(\mathcal{A}) = 5$, and the discrete derivatives $\Delta_f(e|\mathcal{A}) = -1$ and $\Delta_f(e|B) = 0$ which violates Definition \ref{def:submodularity}.
The fact that the discrete derivative is negative violates Definition \ref{def:monotone} because $f(\mathcal{A}) > f(\mathcal{A} \cup \{e\})$.

\subsection{Further equalities}

In addition, our problem can also be phrased as a \emph{facility location} problem~\citep{krarup_ejor1983a} and statements about the facility location problem can be applied to our problem setup as well.

\section{Implementation Details}

\subsection{Software}

We implemented the \automl{} systems and experiments in the Python3 programming language, using \emph{numpy}~\citep{numpy}, \emph{scipy}~\citep{scipy}, \emph{scikit-learn}~\citep{scikit-learn}, \emph{pandas}~\citep{pandas,pandas_1_2_5}, and \emph{matplotlib}~\citep{matplotlib}. We used version 0.12.6 of the \autosklearn{} Python package for the experiments and added \askltwo{} functionality in version 0.12.7 which we then used for the \automl{} benchmark. We give the exact version numbers used for the \automl{} benchmark in Table~\ref{tab:automl:versions}.

\begin{table}[]
    \centering
    \begin{tabular}{l|r}
        \toprule
        Package & Version \\
        \midrule
        \askltwo{} & 0.12.7 \\
        \asklone{} & 0.12.6 \\
        \autoweka{} & 2.6.3 \\
        \tpot{} & 0.11.7 \\
        \hto{} \automl{} & 3.32.1.4 \\
        Tuned Random Forest & 0.24.2 \\
        AutoML benchmark & 973de79617e68a881dcc640842ea1d21dfd4b36c \\
        \bottomrule
    \end{tabular}
    \caption{Package versions used for the \automl{} benchmark.}
    \label{tab:automl:versions}
\end{table}

\subsection{Configuration Space}\label{sec:configspace}

We give the \configspace{} we use in \askltwo{} in Table \ref{tab:configspace}.

\begin{table*}[htbp]
    \scriptsize
    \centering
\begin{tabular}{lccc}
\toprule
Name &                                             Domain &             Default &  Log \\
\midrule
  Classifier                                                   &                   (Extra Trees, Gradient Boosting, MLP,  &       Random Forest &    - \\
  {} & Passive Aggressive, Random Forest, SGD) & {} & {} \\
Extra Trees: Bootstrap                                     &                                                                                   (True, False) &               False &    - \\
Extra Trees: Criterion                                     &                                                                                 (gini, entropy) &                gini &    - \\
Extra Trees: Max Features                                  &                                                                                    $[0.0, 1.0]$ &                 0.5 &   No \\
Extra Trees: Min Samples Leaf                              &                                                                                       $[1, 20]$ &                   1 &   No \\
Extra Trees: Min Samples Split                             &                                                                                       $[2, 20]$ &                   2 &   No \\
Gradient Boosting: Early Stopping                          &                                                                             (off, valid, train) &                 off &    - \\
Gradient Boosting: L2 Regularization                       &                                                                                  $[1e-10, 1.0]$ &                 0.0 &  Yes \\
Gradient Boosting: Learning Rate                           &                                                                                   $[0.01, 1.0]$ &                 0.1 &  Yes \\
Gradient Boosting: Max Leaf Nodes                          &                                                                                     $[3, 2047]$ &                  31 &  Yes \\
Gradient Boosting: Min Samples Leaf                        &                                                                                      $[1, 200]$ &                  20 &  Yes \\
Gradient Boosting: N Iter No Change                        &                                                                                       $[1, 20]$ &                  10 &   No \\
Gradient Boosting: Validation Fraction                     &                                                                                   $[0.01, 0.4]$ &                 0.1 &   No \\
MLP: Activation                                            &                                                                                    (tanh, relu) &                relu &    - \\
MLP: Alpha                                                 &                                                                                  $[1e-07, 0.1]$ &              0.0001 &  Yes \\
MLP: Early Stopping                                        &                                                                                  (valid, train) &               valid &    - \\
MLP: Hidden Layer Depth                                    &                                                                                        $[1, 3]$ &                   1 &   No \\
MLP: Learning Rate Init                                    &                                                                                 $[0.0001, 0.5]$ &               0.001 &  Yes \\
MLP: Num Nodes Per Layer                                   &                                                                                     $[16, 264]$ &                  32 &  Yes \\
Passive Aggressive: C                                      &                                                                                 $[1e-05, 10.0]$ &                 1.0 &  Yes \\
Passive Aggressive: Average                                &                                                                                   (False, True) &               False &    - \\
Passive Aggressive: Loss                                   &                                                                          (hinge, squared\_hinge) &               hinge &    - \\
Passive Aggressive: Tol                                    &                                                                                  $[1e-05, 0.1]$ &              0.0001 &  Yes \\
Random Forest: Bootstrap                                   &                                                                                   (True, False) &                True &    - \\
Random Forest: Criterion                                   &                                                                                 (gini, entropy) &                gini &    - \\
Random Forest: Max Features                                &                                                                                    $[0.0, 1.0]$ &                 0.5 &   No \\
Random Forest: Min Samples Leaf                            &                                                                                       $[1, 20]$ &                   1 &   No \\
Random Forest: Min Samples Split                           &                                                                                       $[2, 20]$ &                   2 &   No \\
Sgd: Alpha                                                 &                                                                                  $[1e-07, 0.1]$ &              0.0001 &  Yes \\
Sgd: Average                                               &                                                                                   (False, True) &               False &    - \\
Sgd: Epsilon                                               &                                                                                  $[1e-05, 0.1]$ &              0.0001 &  Yes \\
Sgd: Eta0                                                  &                                                                                  $[1e-07, 0.1]$ &                0.01 &  Yes \\
Sgd: L1 Ratio                                              &                                                                                  $[1e-09, 1.0]$ &                0.15 &  Yes \\
Sgd: Learning Rate                                         &                                                                 (optimal, invscaling, constant) &          invscaling &    - \\
Sgd: Loss                                                  &                                         (hinge, log, modified Huber,  &                 log &    - \\
{} & squared hinge, perceptron) & {} & {} \\
Sgd: Penalty                                               &                                                                            (l1, l2, elasticnet) &                  l2 &    - \\
Sgd: Power T                                               &                                                                                  $[1e-05, 1.0]$ &                 0.5 &   No \\
Sgd: Tol                                                   &                                                                                  $[1e-05, 0.1]$ &              0.0001 &  Yes \\
\midrule
Balancing: Strategy                                        &                                                                               (none, weighting) &                none &    - \\
Categorical Encoding:   Choice                             &                                                                 (no encoding, one hot encoding) &    one hot encoding &    - \\
Category Coalescence:   Choice                             &                                                            (minority coalescer, no coalescense) &  minority coalescer &    - \\
Category Coalescence: Minimum Fraction &                                                                                 $[0.0001, 0.5]$ &                0.01 &  Yes \\
Imputation of missing values                                       &                                                                   (mean, median, most frequent) &                mean &    - \\
Rescaling:   Choice                                        &  (Min/Max, none, normalize, Power, &         standardize &    - \\
{} & Quantile, Robust, standardize) & {} & {} \\
Quantile Transformer: N Quantiles               &                                                                                    $[10, 2000]$ &                1000 &   No \\
Quantile Transformer: Output Distribution       &                                                                               (uniform, normal) &             uniform &    - \\
Robust Scaler: Q Max                            &                                                                                  $[0.7, 0.999]$ &                0.75 &   No \\
Robust Scaler: Q Min                            &                                                                                  $[0.001, 0.3]$ &                0.25 &   No \\
\bottomrule
\end{tabular}
\caption{Configuration space for \askltwo{} using only iterative models and only preprocessing to transform data into a format that can be usefully employed by the different classification algorithms. The final column (log) states whether we actually search $log_{10}(\lambda)$.}
    \label{tab:configspace}
\end{table*}

\subsection{Successive Halving hyperparameters}
\label{app:successive-halving-hp}

We used the same hyperparameters for all experiments. First, we set to $eta=4$. Next, we had to choose the minimal and maximal budgets assigned to each algorithm. For the tree-based methods we chose to go from $32$ to $512$, while for the linear models (SGD and passive aggressive) we chose $64$ as the minimal budget and $1024$ as the maximal budget. Further tuning these hyperparameters would be an interesting, but an expensive way forward.

\section{Datasets}
\label{app:datasets}

We give the name, OpenML task ID and the size of all datasets we used in Table~\ref{tab:setcharactertrain} and~\ref{tab:setcharactertest}.

\begin{table*}[htbp]
    \centering
    \tiny
\begin{minipage}[t]{0.29\textwidth}
\vspace{0pt}
\begin{tabular}{@{\hskip0.0cm}l@{\hskip0.01cm}r@{\hskip0.1cm}r@{\hskip0.1cm}r@{\hskip0.1cm}r@{\hskip0.0cm}}
name & tid &   \#obs & \#feat & \#cls \\
\midrule
\href{https://www.openml.org/d/1166}{OVA\_O\dots} &    \href{https://www.openml.org/t/75126}{75126} &    1545 &  10937 &     2 \\
\href{https://www.openml.org/d/1161}{OVA\_C\dots} &    \href{https://www.openml.org/t/75125}{75125} &    1545 &  10937 &     2 \\
\href{https://www.openml.org/d/1146}{OVA\_P\dots} &    \href{https://www.openml.org/t/75121}{75121} &    1545 &  10937 &     2 \\
\href{https://www.openml.org/d/1142}{OVA\_E\dots} &    \href{https://www.openml.org/t/75120}{75120} &    1545 &  10937 &     2 \\
\href{https://www.openml.org/d/1134}{OVA\_K\dots} &    \href{https://www.openml.org/t/75116}{75116} &    1545 &  10937 &     2 \\
\href{https://www.openml.org/d/1130}{OVA\_L\dots} &    \href{https://www.openml.org/t/75115}{75115} &    1545 &  10937 &     2 \\
\href{https://www.openml.org/d/1128}{OVA\_B\dots} &    \href{https://www.openml.org/t/75114}{75114} &    1545 &  10937 &     2 \\
\href{https://www.openml.org/d/41084}{UMIST\dots} &  \href{https://www.openml.org/t/189859}{189859} &     575 &  10305 &    20 \\
\href{https://www.openml.org/d/1457}{amazo\dots} &  \href{https://www.openml.org/t/189878}{189878} &    1500 &  10001 &    50 \\
\href{https://www.openml.org/d/1233}{eatin\dots} &  \href{https://www.openml.org/t/189786}{189786} &     945 &   6374 &     7 \\
\href{https://www.openml.org/d/40927}{CIFAR\dots} &  \href{https://www.openml.org/t/167204}{167204} &   60000 &   3073 &    10 \\
\href{https://www.openml.org/d/41989}{GTSRB\dots} &  \href{https://www.openml.org/t/190156}{190156} &   51839 &   2917 &    43 \\
\href{https://www.openml.org/d/4134}{Biore\dots} &    \href{https://www.openml.org/t/75156}{75156} &    3751 &   1777 &     2 \\
\href{https://www.openml.org/d/1039}{hiva\_\dots} &  \href{https://www.openml.org/t/166996}{166996} &    4229 &   1618 &     2 \\
\href{https://www.openml.org/d/41988}{GTSRB\dots} &  \href{https://www.openml.org/t/190157}{190157} &   51839 &   1569 &    43 \\
\href{https://www.openml.org/d/41986}{GTSRB\dots} &  \href{https://www.openml.org/t/190158}{190158} &   51839 &   1569 &    43 \\
\href{https://www.openml.org/d/40978}{Inter\dots} &  \href{https://www.openml.org/t/168791}{168791} &    3279 &   1559 &     2 \\
\href{https://www.openml.org/d/1515}{micro\dots} &  \href{https://www.openml.org/t/146597}{146597} &     571 &   1301 &    20 \\
\href{https://www.openml.org/d/40923}{Devna\dots} &  \href{https://www.openml.org/t/167203}{167203} &   92000 &   1025 &    46 \\
\href{https://www.openml.org/d/40645}{GAMET\dots} &  \href{https://www.openml.org/t/167085}{167085} &    1600 &   1001 &     2 \\
\href{https://www.openml.org/d/41991}{Kuzus\dots} &  \href{https://www.openml.org/t/190154}{190154} &  270912 &    785 &    49 \\
\href{https://www.openml.org/d/554}{mnist\dots} &    \href{https://www.openml.org/t/75098}{75098} &   70000 &    785 &    10 \\
\href{https://www.openml.org/d/41982}{Kuzus\dots} &  \href{https://www.openml.org/t/190159}{190159} &   70000 &    785 &    10 \\
\href{https://www.openml.org/d/300}{isole\dots} &    \href{https://www.openml.org/t/75169}{75169} &    7797 &    618 &    26 \\
\href{https://www.openml.org/d/1478}{har} &  \href{https://www.openml.org/t/126030}{126030} &   10299 &    562 &     6 \\
\href{https://www.openml.org/d/1485}{madel\dots} &  \href{https://www.openml.org/t/146594}{146594} &    2600 &    501 &     2 \\
\href{https://www.openml.org/d/42343}{KDD98\dots} &  \href{https://www.openml.org/t/211723}{211723} &   82318 &    478 &     2 \\
\href{https://www.openml.org/d/41145}{phili\dots} &  \href{https://www.openml.org/t/189864}{189864} &    5832 &    309 &     2 \\
\href{https://www.openml.org/d/41144}{madel\dots} &  \href{https://www.openml.org/t/189863}{189863} &    3140 &    260 &     2 \\
\href{https://www.openml.org/d/41082}{USPS} &  \href{https://www.openml.org/t/189858}{189858} &    9298 &    257 &    10 \\
\href{https://www.openml.org/d/1501}{semei\dots} &    \href{https://www.openml.org/t/75236}{75236} &    1593 &    257 &    10 \\
\href{https://www.openml.org/d/41990}{GTSRB\dots} &  \href{https://www.openml.org/t/190155}{190155} &   51839 &    257 &    43 \\
\href{https://www.openml.org/d/41972}{India\dots} &  \href{https://www.openml.org/t/211720}{211720} &    9144 &    221 &     8 \\
\href{https://www.openml.org/d/40670}{dna} &  \href{https://www.openml.org/t/167202}{167202} &    3186 &    181 &     3 \\
\href{https://www.openml.org/d/1116}{musk} &    \href{https://www.openml.org/t/75108}{75108} &    6598 &    170 &     2 \\
\href{https://www.openml.org/d/40536}{Speed\dots} &  \href{https://www.openml.org/t/146679}{146679} &    8378 &    123 &     2 \\
\href{https://www.openml.org/d/1479}{hill-\dots} &  \href{https://www.openml.org/t/146592}{146592} &    1212 &    101 &     2 \\
\href{https://www.openml.org/d/742}{fri\_c\dots} &  \href{https://www.openml.org/t/166866}{166866} &     500 &    101 &     2 \\
\href{https://www.openml.org/d/40966}{MiceP\dots} &  \href{https://www.openml.org/t/167205}{167205} &    1080 &     82 &     8 \\
\href{https://www.openml.org/d/279}{meta\_\dots} &      \href{https://www.openml.org/t/2356}{2356} &   45164 &     75 &    11 \\
\href{https://www.openml.org/d/1487}{ozone\dots} &    \href{https://www.openml.org/t/75225}{75225} &    2534 &     73 &     2 \\
\href{https://www.openml.org/d/458}{analc\dots} &  \href{https://www.openml.org/t/146576}{146576} &     841 &     71 &     4 \\
\href{https://www.openml.org/d/981}{kdd\_i\dots} &  \href{https://www.openml.org/t/166970}{166970} &   10108 &     69 &     2 \\
\href{https://www.openml.org/d/28}{optdi\dots} &        \href{https://www.openml.org/t/258}{258} &    5620 &     65 &    10 \\
\href{https://www.openml.org/d/1491}{one-h\dots} &    \href{https://www.openml.org/t/75154}{75154} &    1600 &     65 &   100 \\
\href{https://www.openml.org/d/377}{synth\dots} &  \href{https://www.openml.org/t/146574}{146574} &     600 &     62 &     6 \\
\href{https://www.openml.org/d/46}{splic\dots} &        \href{https://www.openml.org/t/275}{275} &    3190 &     61 &     3 \\
\href{https://www.openml.org/d/44}{spamb\dots} &        \href{https://www.openml.org/t/273}{273} &    4601 &     58 &     2 \\
\href{https://www.openml.org/d/1475}{first\dots} &    \href{https://www.openml.org/t/75221}{75221} &    6118 &     52 &     6 \\
\href{https://www.openml.org/d/837}{fri\_c\dots} &    \href{https://www.openml.org/t/75180}{75180} &    1000 &     51 &     2 \\
\href{https://www.openml.org/d/920}{fri\_c\dots} &  \href{https://www.openml.org/t/166944}{166944} &     500 &     51 &     2 \\
\href{https://www.openml.org/d/937}{fri\_c\dots} &  \href{https://www.openml.org/t/166951}{166951} &     500 &     51 &     2 \\
\href{https://www.openml.org/d/4541}{Diabe\dots} &  \href{https://www.openml.org/t/189828}{189828} &  101766 &     50 &     3 \\
\href{https://www.openml.org/d/311}{oil\_s\dots} &      \href{https://www.openml.org/t/3049}{3049} &     937 &     50 &     2 \\
\href{https://www.openml.org/d/722}{pol} &    \href{https://www.openml.org/t/75139}{75139} &   15000 &     49 &     2 \\
\href{https://www.openml.org/d/40705}{tokyo\dots} &  \href{https://www.openml.org/t/167100}{167100} &     959 &     45 &     2 \\
\href{https://www.openml.org/d/1494}{qsar-\dots} &    \href{https://www.openml.org/t/75232}{75232} &    1055 &     42 &     2 \\
\href{https://www.openml.org/d/40499}{textu\dots} &  \href{https://www.openml.org/t/126031}{126031} &    5500 &     41 &    11 \\
\href{https://www.openml.org/d/1549}{autoU\dots} &  \href{https://www.openml.org/t/189899}{189899} &     750 &     41 &     8 \\
\href{https://www.openml.org/d/734}{ailer\dots} &    \href{https://www.openml.org/t/75146}{75146} &   13750 &     41 &     2 \\
\href{https://www.openml.org/d/60}{wavef\dots} &        \href{https://www.openml.org/t/288}{288} &    5000 &     41 &     3 \\
\href{https://www.openml.org/d/6332}{cylin\dots} &  \href{https://www.openml.org/t/146600}{146600} &     540 &     40 &     2 \\
\href{https://www.openml.org/d/940}{water\dots} &  \href{https://www.openml.org/t/166953}{166953} &     527 &     39 &     2 \\
\href{https://www.openml.org/d/2}{annea\dots} &        \href{https://www.openml.org/t/232}{232} &     898 &     39 &     5 \\
\href{https://www.openml.org/d/1056}{mc1} &    \href{https://www.openml.org/t/75133}{75133} &    9466 &     39 &     2 \\
\href{https://www.openml.org/d/1049}{pc4} &    \href{https://www.openml.org/t/75092}{75092} &    1458 &     38 &     2 \\
\href{https://www.openml.org/d/1050}{pc3} &    \href{https://www.openml.org/t/75129}{75129} &    1563 &     38 &     2 \\
\href{https://www.openml.org/d/42206}{porto\dots} &  \href{https://www.openml.org/t/211722}{211722} &  595212 &     38 &     2 \\
\href{https://www.openml.org/d/1069}{pc2} &    \href{https://www.openml.org/t/75100}{75100} &    5589 &     37 &     2 \\
\bottomrule
\end{tabular}
\end{minipage}%
\hfill
\begin{minipage}[t]{0.29\textwidth}
\vspace{0pt}
\begin{tabular}{@{\hskip0.0cm}l@{\hskip0.01cm}r@{\hskip0.1cm}r@{\hskip0.1cm}r@{\hskip0.1cm}r@{\hskip0.0cm}}
name & tid &  \# obs & \# feat & \# class \\
\midrule

\href{https://www.openml.org/d/182}{satim\dots} &      \href{https://www.openml.org/t/2120}{2120} &    6430 &     37 &     6 \\
\href{https://www.openml.org/d/40900}{Satel\dots} &  \href{https://www.openml.org/t/189844}{189844} &    5100 &     37 &     2 \\
\href{https://www.openml.org/d/42}{soybe\dots} &        \href{https://www.openml.org/t/271}{271} &     683 &     36 &    19 \\
\href{https://www.openml.org/d/1466}{cardi\dots} &    \href{https://www.openml.org/t/75217}{75217} &    2126 &     36 &    10 \\
\href{https://www.openml.org/d/23380}{cjs} &  \href{https://www.openml.org/t/146601}{146601} &    2796 &     35 &     6 \\
\href{https://www.openml.org/d/930}{colle\dots} &    \href{https://www.openml.org/t/75212}{75212} &    1302 &     35 &     2 \\
\href{https://www.openml.org/d/752}{puma3\dots} &    \href{https://www.openml.org/t/75153}{75153} &    8192 &     33 &     2 \\
\href{https://www.openml.org/d/4538}{Gestu\dots} &    \href{https://www.openml.org/t/75109}{75109} &    9873 &     33 &     5 \\
\href{https://www.openml.org/d/41162}{kick} &  \href{https://www.openml.org/t/189870}{189870} &   72983 &     33 &     2 \\
\href{https://www.openml.org/d/833}{bank3\dots} &    \href{https://www.openml.org/t/75179}{75179} &    8192 &     33 &     2 \\
\href{https://www.openml.org/d/1510}{wdbc} &  \href{https://www.openml.org/t/146596}{146596} &     569 &     31 &     2 \\
\href{https://www.openml.org/d/4534}{Phish\dots} &    \href{https://www.openml.org/t/75215}{75215} &   11055 &     31 &     2 \\
\href{https://www.openml.org/d/40672}{fars} &  \href{https://www.openml.org/t/189840}{189840} &  100968 &     30 &     8 \\
\href{https://www.openml.org/d/57}{hypot\dots} &      \href{https://www.openml.org/t/3044}{3044} &    3772 &     30 &     4 \\
\href{https://www.openml.org/d/40982}{steel\dots} &  \href{https://www.openml.org/t/168785}{168785} &    1941 &     28 &     7 \\
\href{https://www.openml.org/d/1044}{eye\_m\dots} &  \href{https://www.openml.org/t/189779}{189779} &   10936 &     28 &     3 \\
\href{https://www.openml.org/d/715}{fri\_c\dots} &    \href{https://www.openml.org/t/75136}{75136} &    1000 &     26 &     2 \\
\href{https://www.openml.org/d/903}{fri\_c\dots} &    \href{https://www.openml.org/t/75199}{75199} &    1000 &     26 &     2 \\
\href{https://www.openml.org/d/1497}{wall-\dots} &    \href{https://www.openml.org/t/75235}{75235} &    5456 &     25 &     4 \\
\href{https://www.openml.org/d/40677}{led24\dots} &  \href{https://www.openml.org/t/189841}{189841} &    3200 &     25 &    10 \\
\href{https://www.openml.org/d/40971}{colli\dots} &  \href{https://www.openml.org/t/189845}{189845} &    1000 &     24 &    30 \\
\href{https://www.openml.org/d/41160}{rl} &  \href{https://www.openml.org/t/189869}{189869} &   31406 &     23 &     2 \\
\href{https://www.openml.org/d/24}{mushr\dots} &        \href{https://www.openml.org/t/254}{254} &    8124 &     23 &     2 \\
\href{https://www.openml.org/d/757}{meta} &  \href{https://www.openml.org/t/166875}{166875} &     528 &     22 &     2 \\
\href{https://www.openml.org/d/1053}{jm1} &    \href{https://www.openml.org/t/75093}{75093} &   10885 &     22 &     2 \\
\href{https://www.openml.org/d/1068}{pc1} &    \href{https://www.openml.org/t/75159}{75159} &    1109 &     22 &     2 \\
\href{https://www.openml.org/d/1063}{kc2} &  \href{https://www.openml.org/t/146583}{146583} &     522 &     22 &     2 \\
\href{https://www.openml.org/d/761}{cpu\_a\dots} &    \href{https://www.openml.org/t/75233}{75233} &    8192 &     22 &     2 \\
\href{https://www.openml.org/d/1547}{autoU\dots} &    \href{https://www.openml.org/t/75089}{75089} &    1000 &     21 &     2 \\
\href{https://www.openml.org/d/40646}{GAMET\dots} &  \href{https://www.openml.org/t/167086}{167086} &    1600 &     21 &     2 \\
\href{https://www.openml.org/d/40647}{GAMET\dots} &  \href{https://www.openml.org/t/167087}{167087} &    1600 &     21 &     2 \\
\href{https://www.openml.org/d/825}{bosto\dots} &  \href{https://www.openml.org/t/166905}{166905} &     506 &     21 &     2 \\
\href{https://www.openml.org/d/40648}{GAMET\dots} &  \href{https://www.openml.org/t/167088}{167088} &    1600 &     21 &     2 \\
\href{https://www.openml.org/d/40649}{GAMET\dots} &  \href{https://www.openml.org/t/167089}{167089} &    1600 &     21 &     2 \\
\href{https://www.openml.org/d/40701}{churn\dots} &  \href{https://www.openml.org/t/167097}{167097} &    5000 &     21 &     2 \\
\href{https://www.openml.org/d/40994}{clima\dots} &  \href{https://www.openml.org/t/167106}{167106} &     540 &     21 &     2 \\
\href{https://www.openml.org/d/41671}{micro\dots} &  \href{https://www.openml.org/t/189875}{189875} &   20000 &     21 &     5 \\
\href{https://www.openml.org/d/40650}{GAMET\dots} &  \href{https://www.openml.org/t/167090}{167090} &    1600 &     21 &     2 \\
\href{https://www.openml.org/d/42345}{Traff\dots} &  \href{https://www.openml.org/t/211724}{211724} &   70340 &     21 &     3 \\
\href{https://www.openml.org/d/1496}{ringn\dots} &    \href{https://www.openml.org/t/75234}{75234} &    7400 &     21 &     2 \\
\href{https://www.openml.org/d/1507}{twono\dots} &    \href{https://www.openml.org/t/75187}{75187} &    7400 &     21 &     2 \\
\href{https://www.openml.org/d/188}{eucal\dots} &      \href{https://www.openml.org/t/2125}{2125} &     736 &     20 &     5 \\
\href{https://www.openml.org/d/846}{eleva\dots} &    \href{https://www.openml.org/t/75184}{75184} &   16599 &     19 &     2 \\
\href{https://www.openml.org/d/802}{pbcse\dots} &  \href{https://www.openml.org/t/166897}{166897} &    1945 &     19 &     2 \\
\href{https://www.openml.org/d/185}{baseb\dots} &      \href{https://www.openml.org/t/2123}{2123} &    1340 &     18 &     3 \\
\href{https://www.openml.org/d/821}{house\dots} &    \href{https://www.openml.org/t/75174}{75174} &   22784 &     17 &     2 \\
\href{https://www.openml.org/d/897}{colle\dots} &    \href{https://www.openml.org/t/75196}{75196} &    1161 &     17 &     2 \\
\href{https://www.openml.org/d/4552}{BachC\dots} &  \href{https://www.openml.org/t/189829}{189829} &    5665 &     17 &   102 \\
\href{https://www.openml.org/d/32}{pendi\dots} &        \href{https://www.openml.org/t/262}{262} &   10992 &     17 &    10 \\
\href{https://www.openml.org/d/6}{lette\dots} &        \href{https://www.openml.org/t/236}{236} &   20000 &     17 &    26 \\
\href{https://www.openml.org/d/1503}{spoke\dots} &    \href{https://www.openml.org/t/75178}{75178} &  263256 &     15 &    10 \\
\href{https://www.openml.org/d/1471}{eeg-e\dots} &    \href{https://www.openml.org/t/75219}{75219} &   14980 &     15 &     2 \\
\href{https://www.openml.org/d/847}{wind} &    \href{https://www.openml.org/t/75185}{75185} &    6574 &     15 &     2 \\
\href{https://www.openml.org/d/375}{Japan\dots} &  \href{https://www.openml.org/t/126021}{126021} &    9961 &     15 &     9 \\
\href{https://www.openml.org/d/42193}{compa\dots} &  \href{https://www.openml.org/t/211721}{211721} &    5278 &     14 &     2 \\
\href{https://www.openml.org/d/307}{vowel\dots} &      \href{https://www.openml.org/t/3047}{3047} &     990 &     13 &    11 \\
\href{https://www.openml.org/d/735}{cpu\_s\dots} &    \href{https://www.openml.org/t/75147}{75147} &    8192 &     13 &     2 \\
\href{https://www.openml.org/d/1553}{autoU\dots} &  \href{https://www.openml.org/t/189900}{189900} &     700 &     13 &     3 \\
\href{https://www.openml.org/d/1552}{autoU\dots} &    \href{https://www.openml.org/t/75118}{75118} &    1100 &     13 &     5 \\
\href{https://www.openml.org/d/23381}{dress\dots} &  \href{https://www.openml.org/t/146602}{146602} &     500 &     13 &     2 \\
\href{https://www.openml.org/d/826}{senso\dots} &  \href{https://www.openml.org/t/166906}{166906} &     576 &     12 &     2 \\
\href{https://www.openml.org/d/40498}{wine-\dots} &  \href{https://www.openml.org/t/189836}{189836} &    4898 &     12 &     7 \\
\href{https://www.openml.org/d/40691}{wine-\dots} &  \href{https://www.openml.org/t/189843}{189843} &    1599 &     12 &     6 \\
\href{https://www.openml.org/d/1120}{Magic\dots} &    \href{https://www.openml.org/t/75112}{75112} &   19020 &     12 &     2 \\
\href{https://www.openml.org/d/881}{mv} &    \href{https://www.openml.org/t/75195}{75195} &   40768 &     11 &     2 \\
\href{https://www.openml.org/d/40706}{parit\dots} &  \href{https://www.openml.org/t/167101}{167101} &    1124 &     11 &     2 \\
\href{https://www.openml.org/d/40680}{mofn-\dots} &  \href{https://www.openml.org/t/167094}{167094} &    1324 &     11 &     2 \\
\href{https://www.openml.org/d/740}{fri\_c\dots} &    \href{https://www.openml.org/t/75149}{75149} &    1000 &     11 &     2 \\
\href{https://www.openml.org/d/155}{poker\dots} &        \href{https://www.openml.org/t/340}{340} &  829201 &     11 &    10 \\
\bottomrule
\end{tabular}
\end{minipage}%
\hfill
\begin{minipage}[t]{0.29\textwidth}
\vspace{0pt}
\begin{tabular}{@{\hskip0.0cm}l@{\hskip0.01cm}r@{\hskip0.1cm}r@{\hskip0.1cm}r@{\hskip0.1cm}r@{\hskip0.0cm}}
name & tid &   \#obs & \#feat & \#cls \\
\midrule
\href{https://www.openml.org/d/936}{fri\_c\dots} &  \href{https://www.openml.org/t/166950}{166950} &     500 &     11 &     2 \\
\href{https://www.openml.org/d/30}{page-\dots} &        \href{https://www.openml.org/t/260}{260} &    5473 &     11 &     5 \\
\href{https://www.openml.org/d/1480}{ilpd} &  \href{https://www.openml.org/t/146593}{146593} &     583 &     11 &     2 \\
\href{https://www.openml.org/d/727}{2dpla\dots} &    \href{https://www.openml.org/t/75142}{75142} &   40768 &     11 &     2 \\
\href{https://www.openml.org/d/901}{fried\dots} &    \href{https://www.openml.org/t/75161}{75161} &   40768 &     11 &     2 \\
\href{https://www.openml.org/d/717}{rmfts\dots} &  \href{https://www.openml.org/t/166859}{166859} &     508 &     11 &     2 \\
\href{https://www.openml.org/d/841}{stock\dots} &  \href{https://www.openml.org/t/166915}{166915} &     950 &     10 &     2 \\
\href{https://www.openml.org/d/50}{tic-t\dots} &        \href{https://www.openml.org/t/279}{279} &     958 &     10 &     2 \\
\href{https://www.openml.org/d/15}{breas\dots} &        \href{https://www.openml.org/t/245}{245} &     699 &     10 &     2 \\
\href{https://www.openml.org/d/40693}{xd6} &  \href{https://www.openml.org/t/167096}{167096} &     973 &     10 &     2 \\
\href{https://www.openml.org/d/23}{cmc} &        \href{https://www.openml.org/t/253}{253} &    1473 &     10 &     3 \\
\href{https://www.openml.org/d/470}{profb\dots} &  \href{https://www.openml.org/t/146578}{146578} &     672 &     10 &     2 \\
\href{https://www.openml.org/d/37}{diabe\dots} &        \href{https://www.openml.org/t/267}{267} &     768 &      9 &     2 \\
\href{https://www.openml.org/d/183}{abalo\dots} &      \href{https://www.openml.org/t/2121}{2121} &    4177 &      9 &    28 \\
\href{https://www.openml.org/d/725}{bank8\dots} &    \href{https://www.openml.org/t/75141}{75141} &    8192 &      9 &     2 \\
\href{https://www.openml.org/d/151}{elect\dots} &        \href{https://www.openml.org/t/336}{336} &   45312 &      9 &     2 \\
\href{https://www.openml.org/d/839}{kdd\_e\dots} &  \href{https://www.openml.org/t/166913}{166913} &     782 &      9 &     2 \\
\href{https://www.openml.org/d/823}{house\dots} &    \href{https://www.openml.org/t/75176}{75176} &   20640 &      9 &     2 \\
\href{https://www.openml.org/d/26}{nurse\dots} &        \href{https://www.openml.org/t/256}{256} &   12960 &      9 &     5 \\
\href{https://www.openml.org/d/807}{kin8n\dots} &    \href{https://www.openml.org/t/75166}{75166} &    8192 &      9 &     2 \\
\href{https://www.openml.org/d/181}{yeast\dots} &      \href{https://www.openml.org/t/2119}{2119} &    1484 &      9 &    10 \\
\href{https://www.openml.org/d/816}{puma8\dots} &    \href{https://www.openml.org/t/75171}{75171} &    8192 &      9 &     2 \\
\href{https://www.openml.org/d/728}{analc\dots} &    \href{https://www.openml.org/t/75143}{75143} &    4052 &      8 &     2 \\
\href{https://www.openml.org/d/1483}{ldpa} &    \href{https://www.openml.org/t/75134}{75134} &  164860 &      8 &    11 \\
\href{https://www.openml.org/d/750}{pm10} &  \href{https://www.openml.org/t/166872}{166872} &     500 &      8 &     2 \\
\href{https://www.openml.org/d/886}{no2} &  \href{https://www.openml.org/t/166932}{166932} &     500 &      8 &     2 \\
\href{https://www.openml.org/d/40496}{LED-d\dots} &  \href{https://www.openml.org/t/146603}{146603} &     500 &      8 &    10 \\
\href{https://www.openml.org/d/1459}{artif\dots} &  \href{https://www.openml.org/t/126028}{126028} &   10218 &      8 &    10 \\
\href{https://www.openml.org/d/335}{monks\dots} &      \href{https://www.openml.org/t/3055}{3055} &     554 &      7 &     2 \\
\href{https://www.openml.org/d/737}{space\dots} &    \href{https://www.openml.org/t/75148}{75148} &    3107 &      7 &     2 \\
\href{https://www.openml.org/d/1481}{kr-vs\dots} &    \href{https://www.openml.org/t/75223}{75223} &   28056 &      7 &    18 \\
\href{https://www.openml.org/d/334}{monks\dots} &      \href{https://www.openml.org/t/3054}{3054} &     601 &      7 &     2 \\
\href{https://www.openml.org/d/40922}{Run\_o\dots} &  \href{https://www.openml.org/t/167103}{167103} &   88588 &      7 &     2 \\
\href{https://www.openml.org/d/819}{delta\dots} &    \href{https://www.openml.org/t/75173}{75173} &    9517 &      7 &     2 \\
\href{https://www.openml.org/d/770}{strik\dots} &  \href{https://www.openml.org/t/166882}{166882} &     625 &      7 &     2 \\
\href{https://www.openml.org/d/310}{mammo\dots} &      \href{https://www.openml.org/t/3048}{3048} &   11183 &      7 &     2 \\
\href{https://www.openml.org/d/333}{monks\dots} &      \href{https://www.openml.org/t/3053}{3053} &     556 &      7 &     2 \\
\href{https://www.openml.org/d/184}{kropt\dots} &      \href{https://www.openml.org/t/2122}{2122} &   28056 &      7 &    18 \\
\href{https://www.openml.org/d/803}{delta\dots} &    \href{https://www.openml.org/t/75163}{75163} &    7129 &      6 &     2 \\
\href{https://www.openml.org/d/40983}{wilt} &  \href{https://www.openml.org/t/167105}{167105} &    4839 &      6 &     2 \\
\href{https://www.openml.org/d/799}{fri\_c\dots} &    \href{https://www.openml.org/t/75131}{75131} &    1000 &      6 &     2 \\
\href{https://www.openml.org/d/1046}{mozil\dots} &  \href{https://www.openml.org/t/126024}{126024} &   15545 &      6 &     2 \\
\href{https://www.openml.org/d/871}{polle\dots} &    \href{https://www.openml.org/t/75192}{75192} &    3848 &      6 &     2 \\
\href{https://www.openml.org/d/934}{socmo\dots} &    \href{https://www.openml.org/t/75213}{75213} &    1156 &      6 &     2 \\
\href{https://www.openml.org/d/451}{irish\dots} &  \href{https://www.openml.org/t/146575}{146575} &     500 &      6 &     2 \\
\href{https://www.openml.org/d/884}{fri\_c\dots} &  \href{https://www.openml.org/t/166931}{166931} &     500 &      6 &     2 \\
\href{https://www.openml.org/d/949}{arsen\dots} &  \href{https://www.openml.org/t/166957}{166957} &     559 &      5 &     2 \\
\href{https://www.openml.org/d/947}{arsen\dots} &  \href{https://www.openml.org/t/166956}{166956} &     559 &      5 &     2 \\
\href{https://www.openml.org/d/1509}{walki\dots} &    \href{https://www.openml.org/t/75250}{75250} &  149332 &      5 &    22 \\
\href{https://www.openml.org/d/469}{analc\dots} &  \href{https://www.openml.org/t/146577}{146577} &     797 &      5 &     6 \\
\href{https://www.openml.org/d/1462}{bankn\dots} &  \href{https://www.openml.org/t/146586}{146586} &    1372 &      5 &     2 \\
\href{https://www.openml.org/d/951}{arsen\dots} &  \href{https://www.openml.org/t/166959}{166959} &     559 &      5 &     2 \\
\href{https://www.openml.org/d/923}{visua\dots} &    \href{https://www.openml.org/t/75210}{75210} &    8641 &      5 &     2 \\
\href{https://www.openml.org/d/11}{balan\dots} &        \href{https://www.openml.org/t/241}{241} &     625 &      5 &     3 \\
\href{https://www.openml.org/d/950}{arsen\dots} &  \href{https://www.openml.org/t/166958}{166958} &     559 &      5 &     2 \\
\href{https://www.openml.org/d/1536}{volca\dots} &  \href{https://www.openml.org/t/189902}{189902} &   10130 &      4 &     5 \\
\href{https://www.openml.org/d/1502}{skin-\dots} &    \href{https://www.openml.org/t/75237}{75237} &  245057 &      4 &     2 \\
\href{https://www.openml.org/d/40985}{tamil\dots} &  \href{https://www.openml.org/t/189846}{189846} &   45781 &      4 &    20 \\
\href{https://www.openml.org/d/772}{quake\dots} &    \href{https://www.openml.org/t/75157}{75157} &    2178 &      4 &     2 \\
\href{https://www.openml.org/d/1541}{volca\dots} &  \href{https://www.openml.org/t/189893}{189893} &    8654 &      4 &     5 \\
\href{https://www.openml.org/d/1538}{volca\dots} &  \href{https://www.openml.org/t/189890}{189890} &    8753 &      4 &     5 \\
\href{https://www.openml.org/d/1535}{volca\dots} &  \href{https://www.openml.org/t/189887}{189887} &    9989 &      4 &     5 \\
\href{https://www.openml.org/d/1532}{volca\dots} &  \href{https://www.openml.org/t/189884}{189884} &   10668 &      4 &     5 \\
\href{https://www.openml.org/d/1531}{volca\dots} &  \href{https://www.openml.org/t/189883}{189883} &   10176 &      4 &     5 \\
\href{https://www.openml.org/d/1530}{volca\dots} &  \href{https://www.openml.org/t/189882}{189882} &    1515 &      4 &     5 \\
\href{https://www.openml.org/d/1529}{volca\dots} &  \href{https://www.openml.org/t/189881}{189881} &    1521 &      4 &     5 \\
\href{https://www.openml.org/d/1528}{volca\dots} &  \href{https://www.openml.org/t/189880}{189880} &    1623 &      4 &     5 \\
\href{https://www.openml.org/d/40704}{Titan\dots} &  \href{https://www.openml.org/t/167099}{167099} &    2201 &      4 &     2 \\
\href{https://www.openml.org/d/1542}{volca\dots} &  \href{https://www.openml.org/t/189894}{189894} &    1183 &      4 &     5 \\
\bottomrule
\end{tabular}
\end{minipage}
\caption{Characteristics of the $\ndatasets$ datasets in $\MetaSet$ (first part) sorted by number of features. We report for each dataset the name and the task id (as a link) as used on \href{https.openml.org}{OpenML.org}, and furthermore the number of observations, the number of features and the number of classes.}
\label{tab:setcharactertrain}
\end{table*}

\begin{table*}
\small
\begin{minipage}[t]{0.49\textwidth}
\vspace{0pt}
\begin{tabular}{lrrrr}
name & tid &   \#obs & \#feat & \#cls \\
\midrule
\href{https://www.openml.org/d/41165}{rober\dots} &  \href{https://www.openml.org/t/168794}{168794} &   10000 &   7201 &    10 \\
\href{https://www.openml.org/d/41161}{ricca\dots} &  \href{https://www.openml.org/t/168797}{168797} &   20000 &   4297 &     2 \\
\href{https://www.openml.org/d/41159}{guill\dots} &  \href{https://www.openml.org/t/168796}{168796} &   20000 &   4297 &     2 \\
\href{https://www.openml.org/d/41163}{dilbe\dots} &  \href{https://www.openml.org/t/189871}{189871} &   10000 &   2001 &     5 \\
\href{https://www.openml.org/d/41142}{chris\dots} &  \href{https://www.openml.org/t/189861}{189861} &    5418 &   1637 &     2 \\
\href{https://www.openml.org/d/1468}{cnae-\dots} &  \href{https://www.openml.org/t/167185}{167185} &    1080 &    857 &     9 \\
\href{https://www.openml.org/d/41164}{faber\dots} &  \href{https://www.openml.org/t/189872}{189872} &    8237 &    801 &     7 \\
\href{https://www.openml.org/d/40996}{Fashi\dots} &  \href{https://www.openml.org/t/189908}{189908} &   70000 &    785 &    10 \\
\href{https://www.openml.org/d/1111}{KDDCu\dots} &    \href{https://www.openml.org/t/75105}{75105} &   50000 &    231 &     2 \\
\href{https://www.openml.org/d/12}{mfeat\dots} &  \href{https://www.openml.org/t/167152}{167152} &    2000 &    217 &    10 \\
\href{https://www.openml.org/d/41166}{volke\dots} &  \href{https://www.openml.org/t/168793}{168793} &   58310 &    181 &    10 \\
\href{https://www.openml.org/d/41138}{APSFa\dots} &  \href{https://www.openml.org/t/189860}{189860} &   76000 &    171 &     2 \\
\href{https://www.openml.org/d/41143}{jasmi\dots} &  \href{https://www.openml.org/t/189862}{189862} &    2984 &    145 &     2 \\
\href{https://www.openml.org/d/1486}{nomao\dots} &  \href{https://www.openml.org/t/126026}{126026} &   34465 &    119 &     2 \\
\href{https://www.openml.org/d/41147}{alber\dots} &  \href{https://www.openml.org/t/189866}{189866} &  425240 &     79 &     2 \\
\href{https://www.openml.org/d/41167}{dioni\dots} &  \href{https://www.openml.org/t/189873}{189873} &  416188 &     61 &   355 \\
\href{https://www.openml.org/d/41168}{janni\dots} &  \href{https://www.openml.org/t/168792}{168792} &   83733 &     55 &     4 \\
\href{https://www.openml.org/d/1596}{cover\dots} &    \href{https://www.openml.org/t/75193}{75193} &  581012 &     55 &     7 \\
\href{https://www.openml.org/d/41150}{MiniB\dots} &  \href{https://www.openml.org/t/168798}{168798} &  130064 &     51 &     2 \\
\href{https://www.openml.org/d/40668}{conne\dots} &  \href{https://www.openml.org/t/167201}{167201} &   67557 &     43 &     3 \\
\bottomrule
\end{tabular}
\end{minipage}%
\hfill
\begin{minipage}[t]{0.49\textwidth}
\vspace{0pt}
\begin{tabular}{lrrrr}
name & tid &   \#obs & \#feat & \#cls \\
\midrule
\href{https://www.openml.org/d/3}{kr-vs\dots} & \href{https://www.openml.org/t/167149}{167149} &    3196 &     37 &     2 \\
\href{https://www.openml.org/d/23512}{higgs\dots} &  \href{https://www.openml.org/t/167200}{167200} &   98050 &     29 &     2 \\
\href{https://www.openml.org/d/41169}{helen\dots} &  \href{https://www.openml.org/t/189874}{189874} &   65196 &     28 &   100 \\
\href{https://www.openml.org/d/1067}{kc1} &  \href{https://www.openml.org/t/167181}{167181} &    2109 &     22 &     2 \\
\href{https://www.openml.org/d/23517}{numer\dots} &  \href{https://www.openml.org/t/167083}{167083} &   96320 &     22 &     2 \\
\href{https://www.openml.org/d/31}{credi\dots} &  \href{https://www.openml.org/t/167161}{167161} &    1000 &     21 &     2 \\
\href{https://www.openml.org/d/41146}{sylvi\dots} &  \href{https://www.openml.org/t/189865}{189865} &    5124 &     21 &     2 \\
\href{https://www.openml.org/d/40984}{segme\dots} &  \href{https://www.openml.org/t/189906}{189906} &    2310 &     20 &     7 \\
\href{https://www.openml.org/d/54}{vehic\dots} &  \href{https://www.openml.org/t/167168}{167168} &     846 &     19 &     4 \\
\href{https://www.openml.org/d/1461}{bank-\dots} &  \href{https://www.openml.org/t/126029}{126029} &   45211 &     17 &     2 \\
\href{https://www.openml.org/d/40981}{Austr\dots} &  \href{https://www.openml.org/t/167104}{167104} &     690 &     15 &     2 \\
\href{https://www.openml.org/d/1590}{adult\dots} &  \href{https://www.openml.org/t/126025}{126025} &   48842 &     15 &     2 \\
\href{https://www.openml.org/d/4135}{Amazo\dots} &    \href{https://www.openml.org/t/75097}{75097} &   32769 &     10 &     2 \\
\href{https://www.openml.org/d/40685}{shutt\dots} &  \href{https://www.openml.org/t/168795}{168795} &   58000 &     10 &     7 \\
\href{https://www.openml.org/d/1169}{airli\dots} &    \href{https://www.openml.org/t/75127}{75127} &  539383 &      8 &     2 \\
     \href{https://www.openml.org/d/40975}{car} &  \href{https://www.openml.org/t/189905}{189905} &    1728 &      7 &     4 \\
\href{https://www.openml.org/d/41027}{jungl\dots} &  \href{https://www.openml.org/t/189909}{189909} &   44819 &      7 &     3 \\
\href{https://www.openml.org/d/1489}{phone\dots} &  \href{https://www.openml.org/t/167190}{167190} &    5404 &      6 &     2 \\
\href{https://www.openml.org/d/1464}{blood\dots} &  \href{https://www.openml.org/t/167184}{167184} &     748 &      5 &     2 \\
\bottomrule
\end{tabular}
\end{minipage}
\caption{Characteristics of the $\nautodata$ datasets in $\AutoMLSet$ sorted by number of features. We report for each dataset the name and the task id (as a link) as used on \href{https.openml.org}{OpenML.org}, and furthermore the number of observations, the number of features and the number of classes.}
\label{tab:setcharactertest}
\end{table*}

\newpage{}
\bibliography{stringslocal,strings,local,lib,proc,proclocal}

\end{document}